\documentclass[lettersize,journal]{IEEEtran}
\usepackage{amsmath,amsfonts}
\usepackage{algorithmic}
\usepackage{algorithm}
\usepackage{array}
\usepackage{bibunits}
\usepackage{pifont}

\usepackage{textcomp}
\usepackage{stfloats}
\usepackage{url}
\usepackage{verbatim}
\usepackage{graphicx}
\usepackage{cite}
\usepackage{xcolor}
\usepackage[table]{xcolor}
\definecolor{lightgray}{RGB}{235,235,235}
\definecolor{lightblue}{RGB}{220,235,255}

\definecolor{iccvblue}{rgb}{0.21,0.49,0.74}
\usepackage{amssymb}
\usepackage{booktabs}
\usepackage{multirow}
\usepackage{colortbl}
\usepackage{adjustbox}
\usepackage{dsfont}

\usepackage{wrapfig}
\newcommand{\tsb}{\textsubscript}
\usepackage{hyperref}
\usepackage{float}
\usepackage{adjustbox}
\usepackage{graphicx}
\usepackage{subcaption}

\title{CKAA: Cross-subspace Knowledge Alignment and Aggregation for Robust Continual Learning}

\author{
Lingfeng He,~\IEEEmembership{}
De Cheng,~\IEEEmembership{} 
Zhiheng Ma,~\IEEEmembership{} 
Huaijie Wang,~\IEEEmembership{} 
Dingwen Zhang,~\IEEEmembership{} 
Nannan Wang,~\IEEEmembership{Senior Member,~IEEE,}
Xinbo Gao,~\IEEEmembership{Fellow,~IEEE}
\thanks{
This work was supported in part by the National Key Research and
Development Program of China under Grant 2023YFA1008600; in part by the
National Natural Science Foundation of China under Grant 62576262, Grant
U22A2096;  in part by the Natural Science Basic Research Program of Shaanxi (Program No. 2026JC-YXQN-202), in part by the Scientific and Technological
Innovation Teams in Shaanxi Province under Grant 2025RS-CXTD-011; in
part by Shaanxi Province Core Technology Research and Development Project
under Grant 2024QY2-GJHX-11; in part by the Fundamental Research Funds
for the Central Universities under Grant QTZX25083 and Grant QTZX23042; 
and in part by CCF-Kuaishou Large Model Explorer Fund under Grant CCF KuaiShou 2025006.

Lingfeng He, De Cheng and Nannan Wang are with School of Telecommunications Engineering, Xidian University, Xi'an, China (email: lfhe@stu.xidian.edu.cn, dcheng@xidian.edu.cn, 
nnwang@xidian.edu.cn).
Huaijie Wang, Xinbo Gao are with School of Electronic Engineering, Xidian University, Xi'an, China (email: huaijie\_wang@stu.xidian.edu.cn, xbgao@mail.xidian.edu.cn).
Zhiheng Ma is with SUAT-Faculty of Computational Microelectronics, Shenzhen University of Advanced Technology, Shenzhen, China (email: mazhiheng@suat-sz.edu.cn)
Dingwen Zhang is with the Brain and Artificial Intelligence Laboratory, Northwestern Polytechnical University, Xi'an, China (email: zdw2006yyy@nwpu.edu.cn)

Corresponding Author: De Cheng (dcheng@xidian.edu.cn)
}
}

\begin{document}
\maketitle
\begin{abstract}
Continual Learning (CL) empowers AI models to continuously learn from sequential task streams.
Recently, parameter-efficient fine-tuning (PEFT)-based CL methods have garnered increasing attention due to their superior performance.
They typically allocate a unique sub-module for learning each task, with a task recognizer to select the appropriate sub-modules for testing images.
However, due to the feature subspace misalignment from independently trained sub-modules, these methods tend to produce ambiguous decisions under misleading task-ids.
To address this, we propose Cross-subspace Knowledge Alignment and Aggregation (CKAA), a novel framework that enhances model robustness against misleading task-ids through two key innovations:
(1) Dual-level Knowledge Alignment (DKA):
By aligning intra-class feature distributions across different subspaces and learning a robust global classifier through a feature simulation process, DKA enables the model to distinguish features from both correct and incorrect subspaces during training.
(2) Task-Confidence-guided Mixture of Adapters (TC-MoA): A robust inference scheme that adaptively aggregates task-specific knowledge from relevant sub-modules based on task-confidence scores, avoiding overconfidence in misleading task-id predictions.
Extensive experiments demonstrate that CKAA outperforms existing PEFT-based CL methods. 
Code is available at \href{https://github.com/HHHLF/CKAA_for_CIL}{\textcolor{blue}{https://github.com/HHHLF/CKAA\_for\_CIL}}.
\end{abstract}




\section{Introduction}

Continual Learning (CL) requires the model to continuously learn from sequentially arriving data streams, which is essential for AI models to adapt to the ever-evolving real-world scenarios \cite{iCaRL, ssaitCVPR24, vptnspNIPS-24}.
CL presents a major challenge: as the model is updated to accommodate new categories, they often suffer from catastrophic forgetting \cite{EwC, regularization4, forgetting2}, wherein knowledge of old classes is diminished.
There are two common scenarios in CL: Task-Incremental Learning (TIL) \cite{TIL0} and Class-Incremental Learning (CIL) \cite{CIL0}.
We investigate the more challenging CIL setting with unknown task-ids during inference \cite{over-simplified-1}.

With the advancement of pre-trained models, Parameter Efficient Fine-Tuning (PEFT)-based methods \cite{lora, adapter, PromptTuning} exhibit superior performance in vision tasks.
These methods keep the pre-trained backbone frozen while inserting lightweight modules for training~\cite{xuPADG, xuRD-MLDG, cllora-cvpr25, he2026task} rather than learning a network from scratch~\cite{replay1, replay2, he2025exploring}.
PEFT-based methods have been investigated in the field of CL \cite{l2pCVPR22, infloraCVPR24, c-adaECCV24} and have proven highly effective in mitigating catastrophic forgetting by maintaining the strong generalizability of pre-trained models throughout sequential training.

\begin{figure}[!t]
\centering
\includegraphics[width=0.50\textwidth]{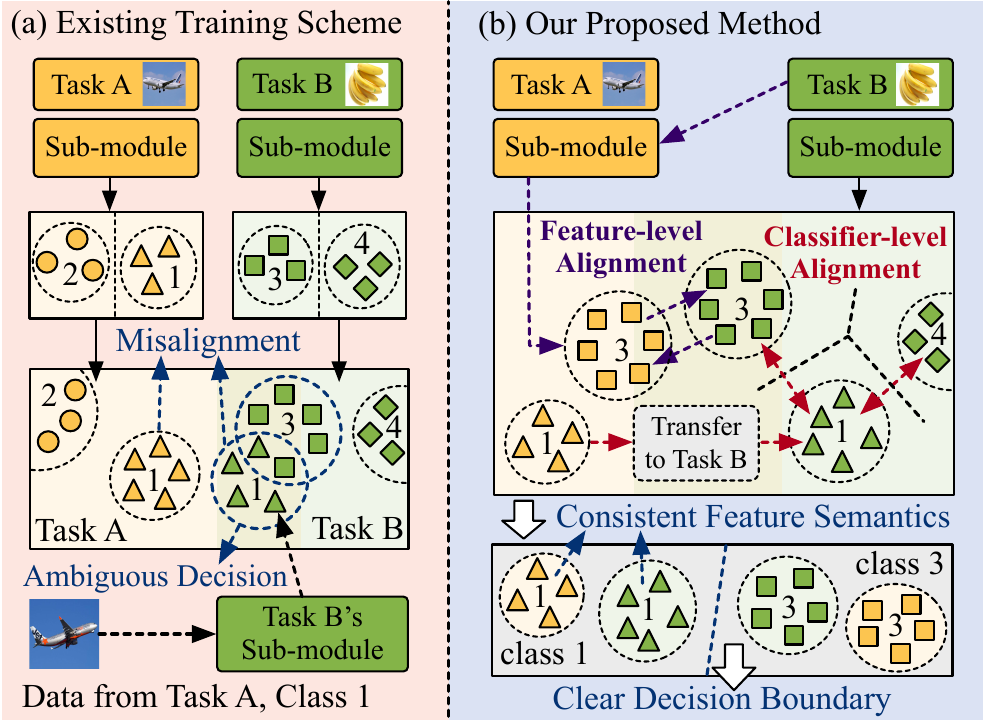}
\vspace{-6mm}
\caption{
Illustration of our motivation.
Existing methods (a) encounter ambiguous decisions for features projected into the incorrect subspace due to the misalignment between subspaces.
Our CKAA (b) addresses this by semantically aligning feature distributions across subspaces and learning robust decision boundaries, effectively distinguishing wrongly projected features.
}\label{fig:introduction}
\vspace{-5mm}
\end{figure}




Several methods \cite{l2pCVPR22, dualpromptECCV22, codaCVPR23, easeCVPR24} are proposed to harness the potential of PEFT methods in the field of CIL.
They typically allocate a dedicated sub-module for each task, effectively uncovering the unique knowledge for each task and ensuring model plasticity for new tasks.
Since task-ids are unavailable during CIL inference, a task recognizer is designed to identify task-ids and select the appropriate sub-module for testing images.
Unfortunately, task-id misidentification is inevitable, leading to features being erroneously projected into incorrect subspaces, which in turn results in ambiguous and unreliable decisions.
As illustrated in Fig.\ref{fig:introduction}(a), when an image from [Task A, Class 1] selects an improper [sub-module B], the model delivers an ambiguous decision between Class 1 and Class 3
for the feature within task B's subspace, resulting in prediction errors. 
Such errors stem from the misaligned feature subspaces caused by the independent training and diverse optimization objectives of the sub-modules, where the classifier fails to distinguish wrongly projected features within incorrect subspaces, despite their meaningful semantic information.

\begin{figure}[!t]
\centering
\includegraphics[width=0.48\textwidth]{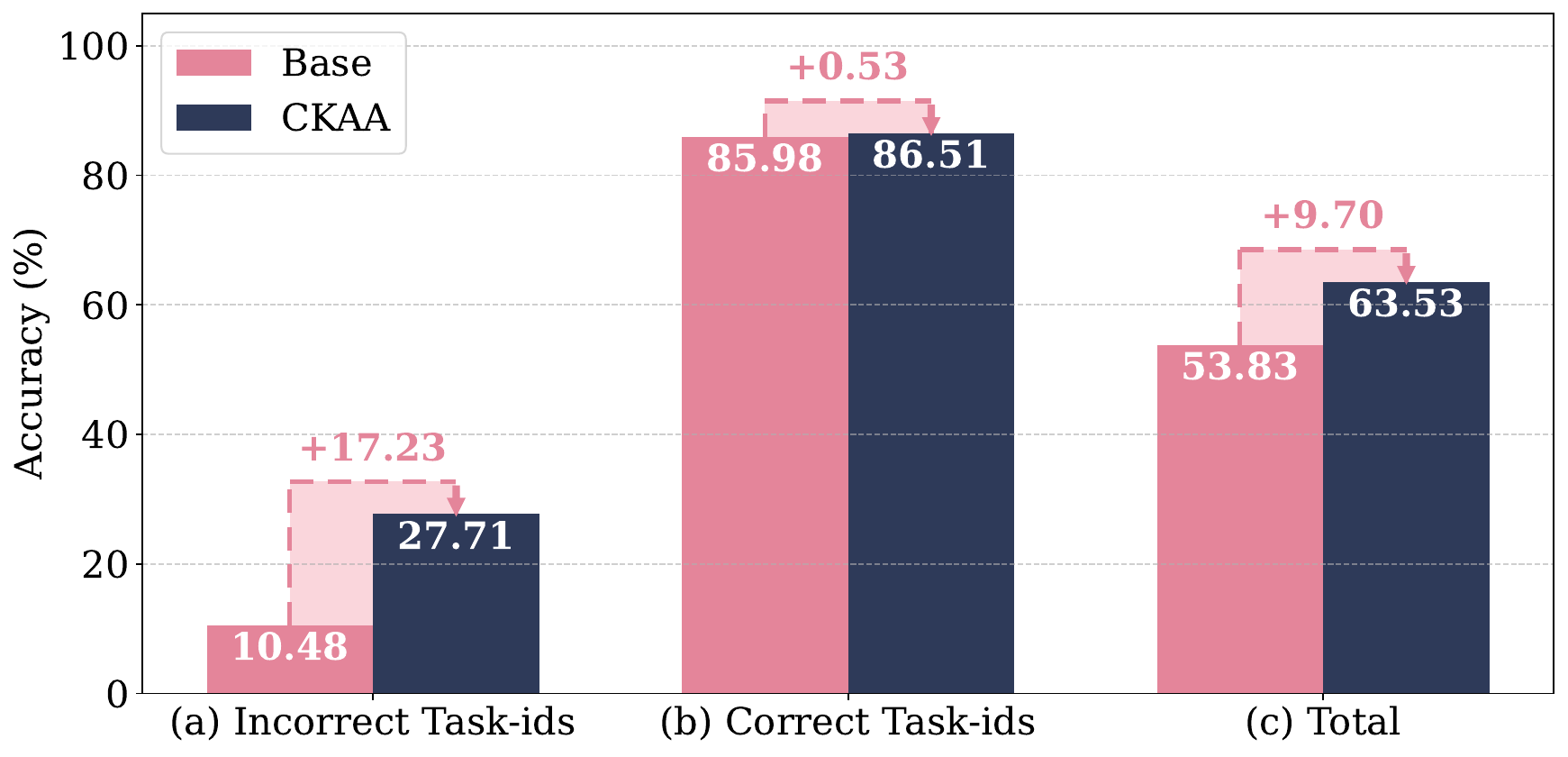}
\vspace{-3mm}
\caption{Accuracy of samples with correct \& incorrect task-ids.
}\label{fig:introduction2}
\vspace{-3mm}
\end{figure}

As shown in Fig.\ref{fig:introduction2},
we visualize the classification accuracy of samples predicted with (a) correct and (b) incorrect task-ids on 10S-ImageNetA.
The term \textbf{Base} refers to performance obtained using task-specific modules trained independently. 
It suffers from the misalignment problem, resulting in poor accuracy (10.48\%).
These observations highlight the vulnerability of the model's robustness when provided with incorrect task-ids, leading to performance degradation.
We further consider that the misalignment issue is mainly reflected in two aspects:
(1) The feature semantics of the same image from the correct and incorrect sub-modules remain inconsistent;
(2) The decision boundaries trained solely with correct task-ids cannot classify features within incorrect subspaces.

Motivated by these analysis, we propose a novel Cross-subspace Knowledge Alignment and Aggregation (CKAA) framework.
First, we build a visual prompt-based task recognizer, alongside dedicated task-specific sub-modules to maintain model plasticity.
Building on this architecture, we propose a Dual-level Knowledge Alignment (DKA, Fig.\ref{fig:introduction}(b)) approach, including feature and classifier-level alignment. Our Feature-level Alignment (FA) groups intra-class features across subspaces together via contrastive learning. 
Unlike standard contrastive learning~\cite{contrastive1, contrastive2} in CL, the positive pairs in FA are defined across task-specific subspaces to model the feature variation caused by incorrect task-IDs. Moreover, rather than directly minimizing cross-subspace feature discrepancies through $L_2$- or KL-based regularization~\cite{EwC, regularization4}, FA is a relative objective that encourages cross-subspace within-class features to be more compact than negative samples, preserving plasticity of task-specific modules.
Building upon these semantically unified feature subspaces, we further train a robust global classifier capable of distinguishing features projected into both correct and incorrect subspaces. 
To achieve this, we proposes an affinity-guided dense feature simulation to simulate wrongly projected features, and train a task-adaptive classifier for each task to distinguish features of all seen classes projected into its subspace. Such dense simulation better preserves complex intra-class structures and generates diverse features for classifier training. 
The task-adaptive classifiers are aggregated into a global classifier for robust inference. Consequently, DKA mitigates the misalignment in both two aforementioned aspects. 

To further enhance model robustness against misleading task-ids during inference, we propose a Task-Confidence-guided Mixture of Adapters (TC-MoA) scheme, which dynamically aggregates task-specific knowledge from semantically unified sub-modules. Specifically, the task recognizer computes task-confidence scores for each test image, reflecting its semantic relevance to each task. These scores serve as routing weights to adaptively combine intermediate features from task-specific sub-modules, forming a Mixture of Adapters (MoA) architecture. Different from existing module selection strategies \cite{l2pCVPR22, codaCVPR23}, our soft aggregation mechanism avoids overconfidence in misleading task-ids by adaptively integrating knowledge from potentially relevant sub-modules, resulting in a more effective and robust inference framework for CIL frameworks with task-specific design.

Unlike existing CIL methods with task-specific designs that mainly focus on improving module selection~\cite{l2pCVPR22, dualpromptECCV22} or aggregation~\cite{vqprompt-nips24, mos-AAAI2025} for inference, CKAA identifies the underlying cause of prediction failure under misleading task-IDs as cross-subspace semantic and decision-boundary misalignment, and improves robustness from both training and inference perspectives. Through the proposed DKA and TC-MoA approaches, CKAA preserves the plasticity of independently adapted task-specific modules while establishing compatible predictions across different subspaces.

The main contributions can be summarized as follows:

\begin{itemize}
\item To enhance the model's robustness against misleading task-ids, we propose a novel CKAA framework, which introduces Dual-level Knowledge Alignment (DKA) to align feature semantics and learn robust global decision boundaries across task-specific subspaces, effectively mitigating the misalignment issue.
\item A Task-Confidence-guided Mixture of Adapters (TC-MoA) scheme is designed to adaptively aggregate knowledge from potentially relevant sub-modules during inference, avoiding overconfidence in misleading task-ids.
\item Extensive experiments on four mainstream CL benchmarks demonstrate that the proposed CKAA outperforms existing state-of-the-art methods.
\end{itemize}

\section{Related Works}

\subsection{Continual Learning}

Continual Learning aims to update the model with new data while mitigating catastrophic forgetting \cite{forgetting1, forgetting2, wei2024class} of previously acquired knowledge.
Traditional methods can be categorized into replay-based \cite{replay1, replay2, replay3, zhou2024balanced},
regularization-based \cite{regularization1, regularization2, lu2024pamk, qiu2025geodesic}, and
expansion-based methods \cite{expansion1, expansion2, expansion3, wang2024model}.
Replay-based methods maintain a memory buffer to store the instances of previous tasks.
Regularization-based methods introduce penalty terms to prevent substantial changes in the parameters important for old tasks.
Expansion-based methods freeze specific parameters for old tasks while allocating new parameters for subsequent tasks.

Recently, continual learning with pre-trained models \cite{l2pCVPR22, codaCVPR23, dualpromptECCV22, adaICCV23, wang2025ekpc, wang2025stpr, cheng2024mamba, li2026few} have gained growing attention, owing to the strong representation capacity of pre-trained models.
Some methods fully fine-tune the pre-trained models \cite{transferECCV22, slcaICCV23}, which is highly time-consuming.
Other methods incorporate PEFT methods into continual learning.
Prompt pool-based methods \cite{l2pCVPR22, dualpromptECCV22, codaCVPR23, cpromptCVPR24} maintain a prompt pool for all tasks and select appropriate prompts for each specific task.
Some adapter-based methods \cite{easeCVPR24, c-adaECCV24, he2026harnessing} build expandable adapter-based architecture while reducing catastrophic forgetting through regularization terms or semantic adjustment.
PGP \cite{pgpICLR24}, VPT-NSP \cite{vptnspNIPS-24}, InfLoRA~\cite{infloraCVPR24} and LoDA~\cite{he2026task} theoretically  integrate gradient projection into the prompt-based and LoRA-based CL, 
achieving impressive performance.

\begin{figure*}[htbp]
\centering
\includegraphics[width=1.0\textwidth]{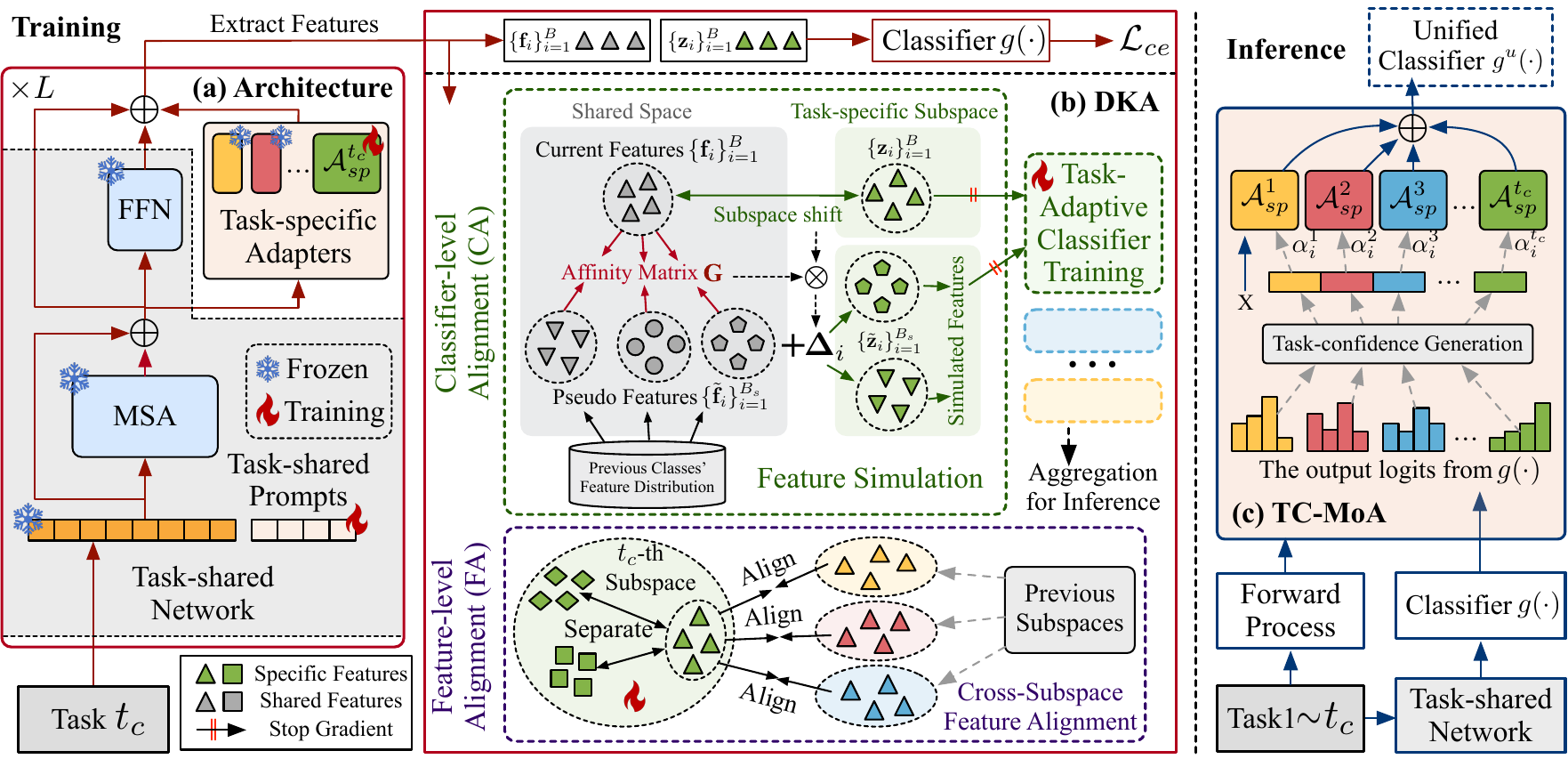}
\vspace{-2mm}
\caption{The framework of our proposed method. Our network architecture (a) is composed of a visual prompt-based task-shared network and adapter-based task-specific sub-modules. 
To enhance the model's robustness against incorrect task-ids, we propose Dual-level Knowledge Alignment approach (b, DKA), which aligns task-specific distributions at the feature and classifier levels. 
During inference, the Task-Confidence-guided Mixture of Adapters (c, TC-MoA) effectively aggregates task-specific knowledge into the task-shared backbone.
}
\vspace{-5mm}
\label{fig:framework}
\end{figure*}

\subsection{Parameter-Efficient Fine-Tuning}

Parameter-Efficient Fine-Tuning (PEFT) is a paradigm to fine-tune pre-trained models by inserting sub-modules with a small number of parameters while keeping the pre-trained parameters frozen.
PEFT methods have shown superior performance in NLP \cite{PEFTinNLP1, lora, PrefixTuning}.
Recently, PEFT methods have been proposed for Vision Transformer (ViT) \cite{vptECCV22, cheng2024disentangled, xu2025adversarial}.
Prompt Tuning \cite{PromptTuning, zhao2024learning, xuan2024adapting} and Prefix-Tuning \cite{PrefixTuning} prepend a set of learnable tokens to the inputs before the transformer blocks.
Adapter \cite{adapter} inserts lightweight modules into the Feed Forward Network (FFN) and Low-Rank Adaptation (LoRA) \cite{lora} injects learnable rank decomposition matrices to re-parameterize the pre-trained weights.
Mixture-of-Experts (MoE) \cite{Mixture-of-Experts} aggregates multiple sub-networks as experts via a routing network.
The above PEFT methods have demonstrated comparable performance to full fine-tuning.

\section{Methodology}


\noindent \textbf{Problem Definition.} In Continual Learning, there is a sequence of tasks with non-overlapping classes. We define the sequential dataset as $\mathcal{D} = \{\mathcal{D}_1, \cdots, \mathcal{D}_{T} \}$. $\mathcal{D}_t = \{\mathbf{x}_i^t, y(\mathbf{x}_i^t)\}_{i=1}^{|\mathcal{D}_t|}$ is the 
dataset associated with the $t$-th task of size $|\mathcal{D}_t|$, and $y(\mathbf{x}_i^t) \in \mathcal{Y}^t$ denotes the corresponding class of $\mathbf{x}_i^t \in \mathbb{R}^{H \times W \times C}$. $\mathcal{Y}^t$ is the label space of task $t$, and $\mathcal{Y}^t \cap \mathcal{Y}^{t'} = \varnothing$ for $t \neq t'$. The objective of continual learning is to train a model sequentially on $T$ tasks and perform well on testing images from all seen classes $\mathcal{Y}^1 \cup  \cdots \cup \mathcal{Y}^T$.

We follow existing PEFT-based methods \cite{l2pCVPR22, dualpromptECCV22, infloraCVPR24} and assume that a pre-trained Vision Transformer (ViT) \cite{ViT} is available for model initialization. The model is decoupled into a feature extractor $\mathcal{F}: \mathbb{R}^{H \times W \times C} \rightarrow \mathbb{R}^d$ and a classifier $g: \mathbb{R}^{d} \rightarrow \mathbb{R}^{\sum_{t=1}^T |\mathcal{Y}^t|}$. Similar to existing work \cite{l2pCVPR22}, we focus on the Class-Incremental Learning (CIL) setting, where task-ids are unknown during inference.




\noindent \textbf{Overall Framework.} The overall framework of our CKAA is illustrated in Fig.\ref{fig:framework}.
Our network architecture (Fig.\ref{fig:framework}(a)) consists of a task-shared network to function as a task recognizer, and task-specific sub-modules to capture unique knowledge for individual tasks.
To enhance the model's robustness against incorrect task-ids, we propose a Dual-level Knowledge Alignment (DKA, Fig.\ref{fig:framework}(b)) approach for training these sub-modules.
Our DKA consists of two components: 
(1) Feature-level Alignment (FA), which groups the same class features from different feature subspaces together;
(2) Classifier-level Alignment (CA), which learns a task-adaptive classifier during each session to address cases where previous images are incorrectly assigned with the current task-id, and aggregates them to build a robust global classifier for inference.
During inference, we introduce a Task-Confidence-guided Mixture of Adapters (TC-MoA, Fig.\ref{fig:framework}(c)) approach, which generates task-confidence scores as routing weights to adaptively aggregate task-specific knowledge without task-ids.


\subsection{Network Architecture}


Our network architecture is composed of: 
(1) A trainable task-shared network, which serves as a task recognizer and captures the common knowledge from the downstream task;
(2) Task-specific sub-modules, which explore the unique knowledge for individual tasks and ensure model plasticity.

\noindent\textbf{Task-shared Network.} We develop our task-shared component on top of
Visual Prompt Tuning (VPT) \cite{vptECCV22}.
Since the visual prompts can inject knowledge before each transformer block as a global operation, they are well-suited to capturing the common knowledge of downstream tasks with strong generalizability.
Specifically, a set of task-shared learnable vectors denoted as $\mathcal{P}_{sh} = \{\mathbf{P}_l\}_{l=1}^L$, are inserted into the original input at each transformer block,
where $\mathbf{P}_l \in \mathbb{R}^{N_p \times d}$ is the prompt corresponding to the $l$-th block. 
At $l$-th block, the prompt $\mathbf{P}_l$ is concatenated with the patch representations $\mathbf{X}_{l}$ to serve as the input to the MSA layer: 

\vspace{-3mm}
\begin{equation}\label{eq: prompt_forward}
    [\hat{\mathbf{X}}_{l},\hat{\mathbf{P}}_l] = \text{MSA}([\mathbf{X}_{l}; \mathbf{P}_{l}]);
    \quad 
    \mathbf{X}_{l+1} = \text{FFN}(\hat{\mathbf{X}}_{l}),
\end{equation}

\noindent 
where $\text{MSA}(\cdot)$ and $\text{FFN}(\cdot)$ denotes the Multi-head Self-Attention (MSA) layer and the Feed Forward Network (FFN) layer in ViT.
The pre-trained feature extractor $\mathcal{F}$ and the prompts $\mathcal{P}_{sh}$ work together to serve as the task-shared network, the shared feature representation of input $\mathbf{x}$ is denoted as $\mathbf{f} = \mathcal{F}(\mathbf{x}; \mathcal{P}_{sh})$. 
Since the task-shared network should adapt to all downstream tasks, it is essential to mitigate its forgetting when training new tasks.
To achieve this objective,
we update $\mathcal{P}_{sh}$ orthogonally to the previous feature space following \cite{vptnspNIPS-24, 2021nullspace, pgpICLR24}.
These prompts $\mathcal{P}_{sh}$ are updated throughout all sequential tasks, encapsulating the common knowledge among them. 
Such orthogonally optimized prompts establish a shared backbone and a powerful task recognizer, tailored for downstream tasks.

\noindent\textbf{Task-specific Sub-modules.}
Due to the stringent orthogonal constraints, the task-shared network is insufficient to capture fine-grained knowledge specific to each individual task, thereby limiting the model's plasticity.
Therefore, we allocate a task-specific sub-module for each task to capture its unique knowledge and generate more discriminative features.
We design task-specific sub-modules based on an adapter architecture \cite{adapter}, denoted as $\mathcal{A}_{sp}^t = \{ \mathcal{A}^t_l \}_{l=1}^{L}$.
Since the residual connection nature of adapter-tuning, it can effectively enable the model to aggregate unique knowledge from different tasks independently.
For the adapter $\mathcal{A}^t_l \in \mathcal{A}_{sp}^t$ in $l$-th block , it is composed of a downsampling matrix $\mathbf{W}_{down} \in \mathbb{R}^{d \times \hat{d}}$ and an upsampled matrix $\mathbf{W}_{up} \in \mathbb{R}^{\hat{d} \times d}$. 
Denote the input as $\mathbf{X}$, the forward process can be formulated as $\mathcal{A}(\mathbf{X}) = \text{ReLU}(\mathbf{X} \cdot \mathbf{W}_{down}) \cdot \mathbf{W}_{up}$.
At $l$-th block, $\mathcal{A}^t_l$ incorporate task-specific knowledge in the FFN layer through a residual connection:

\vspace{-2mm}
\begin{equation}\label{eq: adapter_forward}
    \mathbf{X}_{l+1} = \text{FFN}(\hat{\mathbf{X}}_l) + \mathcal{A}^t_l (\hat{\mathbf{X}}_l).
\end{equation}

\noindent The final task-specific feature of input $\mathbf{x}$ is denoted as $\mathbf{z} = \mathcal{F}(\mathbf{x}; \mathcal{P}_{sh}, \mathcal{A}^t_{sp})$.

\noindent \textbf{Basic loss.} During $t_c$-th session, given a batch $\{\mathbf{x}_i\}_{i=1}^B$ of size $B$ (with the superscript $t_c$ omitted for simplicity), the features $\{\mathbf{f}_i\}_{i=1}^B$ and $\{\mathbf{z}_i\}_{i=1}^B$ from the shared network and the $t_c$-th sub-module are extracted, where $\mathbf{f}_i = \mathcal{F}(\mathbf{x}_i; \mathcal{P}_{sh})$ and $\mathbf{z}_i = \mathcal{F}(\mathbf{x}_i; \mathcal{P}_{sh}, \mathcal{A}^t_{sp})$. These features are trained with the classifier $g$ and a commonly-used cross-entropy loss: 

\vspace{-4.0mm}
\begin{equation}\label{eq: loss_ce}
\mathcal{L}_{ce} = \frac{1}{B} \sum_{i=1}^{B} \mathcal{L}(\sigma(g(\mathbf{f}_i)), y(\mathbf{f}_i)) + \mathcal{L}(\sigma(g(\mathbf{z}_i)), y(\mathbf{z}_i)),
\end{equation}
\vspace{-2.0mm}

\noindent where $y(\mathbf{f})$ represents the ground-truth class label of feature $\mathbf{f}$. $\mathcal{L}$ denotes the standard cross-entropy loss, and $\sigma(\cdot)$ denotes the softmax operation.
Following existing methods \cite{l2pCVPR22, codaCVPR23, vptnspNIPS-24}, $g$ is trained separately on the classes of each task, and the weights are concatenated for inference.

\subsection{Dual-level Knowledge Alignment}

Given our task-specific designs, it is essential to select appropriate sub-modules by estimating task-ids through the task recognizer during inference.
However, there are always inevitable incorrect task-ids predictions.
To enhance the model's robustness against these incorrect task-ids, we propose a Dual-level Knowledge Alignment (DKA) training approach.

\noindent \textbf{Feature-level Alignment (FA).}
For a robust feature extractor capable of generating meaningful features despite incorrect task-ids,
the same image should achieve consistent semantics
when projected into the correct and the incorrect feature subspaces.
Therefore, we semantically unify features across different feature subspaces through FA.
Specifically, we propose a Cross-Subspace Feature Alignment (CSFA) loss, which encourages features of the same class in the current subspace to align with those in the previous subspace.
We first extract the features of the current dataset $\mathcal{D}_{t_c}$ in the $t_p$-th previous subspace for $t_p < t_c$, denoted as $\tilde{\mathbf{z}}^{t_c \rightarrow t_p}_i = \mathcal{F}(\mathbf{x}_i^{t_c}; \mathcal{P}_{sh}, \mathcal{A}_{sp}^{t_p})$ and $\mathbf{x}^{t_c}_i \in \mathcal{D}_{t_c}$.
During training, we randomly sample a feature set $\widetilde{\mathcal{C}}_{sp} = \{\tilde{\mathbf{z}}^{t_c \rightarrow t_p}_j\}_{j=1}^{B}$ in previous subspaces. 
Given an input batch $\mathcal{C}_{sp} = \{\mathbf{z}_i\}_{i=1}^B$, for a feature $\mathbf{z}_k \in \mathcal{C}_{sp}$ with label $y(\mathbf{z}_k)$ in the current subspace, we build its positive set by features from the sampled set with the same ground-truth, where $\mathcal{P}(\mathbf{z}_k) = \{\tilde{\mathbf{z}}^{t_c \rightarrow t_p}_j | y(\tilde{\mathbf{z}}^{t_c \rightarrow t_p}_j) = y(\mathbf{z}_k), \tilde{\mathbf{z}}^{t_c \rightarrow t_p}_j \in \widetilde{\mathcal{C}}_{sp}\}$.
Meanwhile, the features from the current subspace with different labels are regarded as the negative samples, where $\mathcal{N}(\mathbf{z}_k) = \{\mathbf{z}_j |y(\mathbf{z}_j) \neq y(\mathbf{z}_k), \mathbf{z}_j \in \mathcal{C}_{sp} \}$. Then CSFA loss for an input batch $\{\mathbf{z}_i\}_{i=1}^B$ is formulated as follows:


\vspace{-3mm}
\begin{equation}\label{eq: fu-cscl-loss}
    \mathcal{L}_{csfa} = -\frac{1}{B} \sum_{i=1}^B 
    \log
    \frac{\sum_{\mathbf{z}^{+} \in \mathcal{P}(\mathbf{z}_i)} \exp(\text{sim}(\mathbf{z}_i, \mathbf{z}^{+}) / \tau_f)}{
    \sum_{\mathbf{z}_j \in \mathcal{P}(\mathbf{z}_i) \cup \mathcal{N}(\mathbf{z}_i)} \exp(\text{sim}(\mathbf{z}_i, \mathbf{z}_j) / \tau_f)
    },
\end{equation}

\noindent where $\tau_f$ is a temperature factor for contrastive learning. 
The CSFA loss groups the same class features together in different subspaces, alleviating feature-level misalignment.

\noindent \textbf{Classifier-level Alignment (CA).}
Building upon the semantically unified subspaces enhanced by FA, CA aims to train a robust global classifier capable of distinguishing features across all subspaces even under incorrect task-ids.
Specifically, in $t_c$-th session, we train a task-adaptive classifier $g^{t_c}$, which seeks to avoid ambiguous predictions when previous data $\mathbf{x}_i^{t_c} \in \mathcal{D}_{1} \cup \cdots \cup \mathcal{D}_{t_c-1}$ are projected into the current feature subspace.
To achieve this objective, we first simulate the features of the previous data in the current subspace, 
then the simulated features are used to jointly train $g^{t_c}$ alongside the current features.
We further decompose the simulation process into \emph{\textbf{three steps}}: 
\emph{\textbf{(1) Modeling the affinities}} between current and previous data within the shared feature space through the task-shared network;
\emph{\textbf{(2) Modeling the subspace shift}} within current data between the current subspace and the shared ones;
\emph{\textbf{(3) Transferring the subspace shift}} of the current shared features to the previous ones through the affinity matrix.

Since modeling the affinities requires the previous data distributions but there are no exemplars saved for training, we address this issue through Gaussian sampling.
After the $t$-th session, for the $k$-th class in the $t$-th task, we 
approximate its feature distribution in the shared feature space as a Gaussian distribution $\mathcal{G}_{k,t}$.
During the $t_c$-th session, for each arriving batch, we sample pseudo representations $\mathcal{S}_{sh} = \{\tilde{\mathbf{f}}_i, y(\tilde{\mathbf{f}}_i)\}_{i=1}^{B_s}$ with size $B_s$ from previous shared feature distributions $\cup_{t \leq t_c-1} \mathcal{G}_{k,t}$.
In conjunction with the current features $\mathcal{C}_{sh} = \{\mathbf{f}_i, y(\mathbf{f}_i)\}_{i=1}^{B}$ and $\mathcal{C}_{sp} = \{\mathbf{z}_i, y(\mathbf{z}_i)\}_{i=1}^{B}$, the affinity matrix $\mathbf{G} \in \mathbb{R}^{B_s \times B}$ between the previous data and current data in the shared feature space are formulated as follows:

\vspace{-4mm}
\begin{equation}\label{eq: cu-build-graph}
\mathbf{G}_{ij} = \left\{
\begin{array}{ll}
\exp(\text{sim}(\tilde{\mathbf{f}_i}, \mathbf{f}_j) / \tau_g), \quad \text{if } \mathbf{f}_j \in \mathcal{K}(\tilde{\mathbf{f}}_i, K_g), 
\vspace{1mm} \\
0, \quad \text{otherwise},
\end{array}
\right.
\end{equation}

\noindent where $\tau_g$ is a temperature factor.
$\text{sim}(\mathbf{a},\mathbf{b})$ denotes the cosine similarity between vector $\mathbf{a}$ and $\mathbf{b}$, where $\text{sim}(\mathbf{a},\mathbf{b}) = {\mathbf{a}}^{\top} \mathbf{b} / (\Vert \mathbf{a} \Vert \cdot \Vert \mathbf{b} \Vert$). $\mathcal{K}(\tilde{\mathbf{f}}_i, K_g)$ denotes the $K_g$ nearest neighbors in $\mathcal{C}_{sh}$ of $\tilde{\mathbf{f}}_i$.
$\mathbf{G}_{ij}$ represents the affinity between sampled feature $\tilde{\mathbf{f}}_i$ and current feature $\mathbf{f}_j$.

The subspace shift of current data between the current subspace and the shared feature space can be simply modeled by the residual between the specific feature $\{\mathbf{z}_i\}_{i=1}^B$ ad the shared feature $\{\mathbf{f}_i\}_{i=1}^B$ of current data. Subsequently, we transfer the subspace shift from the current features $\{\mathbf{f}_i\}_{i=1}^B$ to the sampled pseudo features $\{\tilde{\mathbf{f}}_i\}_{i=1}^{B_s}$. The simulated features $\{\tilde{\mathbf{z}}_i\}_{i=1}^{B_s}$ are derived as follows:

\vspace{-5mm}
\begin{equation}\label{eq: cu-simulation}
\tilde{\mathbf{z}}_i = \tilde{\mathbf{f}}_i + \Delta_i,
\quad
\Delta_i = \frac{\sum_{j=1}^{B} \mathbf{G}_{ij} (\mathbf{z}_j - \mathbf{f}_j)}
{\sum_{j=1}^{B} \mathbf{G}_{ij}}.
\vspace{-0.7mm}
\end{equation}

Through the above process, we derive the simulated feature set
$\mathcal{S}_{sp} = \{\tilde{\mathbf{z}}_i, y(\tilde{\mathbf{z}}_i)\}_{i=1}^{B_s}$.
Then we obtain the unified feature set $\mathcal{U} = \mathcal{S}_{sh} \cup \mathcal{S}_{sp} \cup \mathcal{C}_{sh} \cup \mathcal{C}_{sp}$, which includes features of images from all seen tasks in the current subspace.
We utilize it to train the task-adaptive classifier $g^{t_c}$:

\vspace{-2mm}
\begin{equation}\label{eq: loss_cu}
    \mathcal{L}_{ca} = \frac{1}{|\mathcal{U}|}\sum_{\{\mathbf{f}, y(\mathbf{f})\} \in \mathcal{U}} \mathcal{L}(\sigma(g^{t_c}(\mathbf{f})), y(\mathbf{f})).
    \vspace{-0.3mm}
\end{equation}

After training $t_c$-th session, to reduce computational overhead for inference, we aggregate the parameters from all the task-adaptive classifiers, $i.e.$, $g^1, \cdots, g^{t_c}$, to build a global classifier $g^u$ that fits all the task-specific feature subspaces.
For the parameter corresponding to the $k$-th class in the $t_p$-th task, $i.e.$, $k \in \mathcal{Y}^{t_p}$, the parameter aggregation can be formulated as follows:

\vspace{-4mm}
\begin{equation}\label{eq: cu-aggregation}
    \mathbf{W}^u[k] = \frac{\sum_{t=t_p}^{t_c} \mathbf{W}^{t}[k]}{t_c-t_p+1}, \quad
    \mathbf{b}^u[k] = \frac{\sum_{t=t_p}^{t_c} \mathbf{b}^{t}[k]}{t_c-t_p+1},
\end{equation}

\noindent where $\mathbf{W}^u[k]$ ($\mathbf{b}^u[k]$) is the $k$-th class's weight (bias) in $g^u$, and $\mathbf{W}^t[k]$ ($\mathbf{b}^{t}[k]$) is the $k$-th class's weight (bias) in $g^t$.
Such an aggregation strategy enables the global classifier to differentiate features projected into all subspaces, even when confronted with misleading task-ids.

\begin{algorithm}[!t] 
\caption{Training pipeline for task $t_c$}  
\label{alg::training}
\begin{algorithmic}[1]
\REQUIRE Dataset $\mathcal{D}_{t_c}$, backbone $\mathcal{F}$, shared prompts $\mathcal{P}_{sh}$, adapter $\mathcal{A}^{t_c}_{sp}$.
\ENSURE Updated $\mathcal{P}_{sh}$, $\mathcal{A}^{t_c}_{sp}$, classifier $g^{t_c}$.
\STATE Initialize $\mathcal{A}^{t_c}_{sp}$ and $g^{t_c}$; \textbf{if} $t_c=1$, initialize $\mathcal{P}_{sh}$;
\FOR{$\{\mathbf{x}_i, y_i\} \in \mathcal{D}_{t_c}$}
    \STATE $\mathbf{f}_i = \mathcal{F}(\mathbf{x}_i; \mathcal{P}_{sh})$; \hfill \textcolor{iccvblue}{// Shared features}
    \STATE $\mathbf{z}_i = \mathcal{F}(\mathbf{x}_i; \mathcal{P}_{sh}, \mathcal{A}^{t_c}_{sp})$; \hfill \textcolor{iccvblue}{// Specific features}
    \STATE Compute $\mathcal{L}_{ce}$ (Eq.\textcolor{iccvblue}{3});
    \IF{$t_c > 1$}
        \STATE Extract $\mathbf{\tilde{z}}_i^{t_c \rightarrow t_p} = \mathcal{F}(\mathbf{x}_i; \mathcal{P}_{sh}, \mathcal{A}^{t_p}_{sp})$;
        \STATE Compute $\mathcal{L}_{csfa}$ (Eq.\textcolor{iccvblue}{4});
        \hfill \textcolor{iccvblue}{// Feature-level alignment}
        \STATE Sample $\mathcal{S}_{sh}$ from $\cup_{t<t_c} \mathcal{G}_{k,t}$;
        \STATE Compute affinities $\mathbf{G}$ (Eq.\textcolor{iccvblue}{5});
        \STATE Simulate $\mathcal{S}_{sp}$ (Eq.\textcolor{iccvblue}{6});
        \STATE Compute $\mathcal{L}_{cu}$ (Eq.\textcolor{iccvblue}{7});
        \hfill \textcolor{iccvblue}{// Classifier-level alignment}
    \ENDIF

    \STATE Compute total loss $\mathcal{L}$ (Eq.\textcolor{iccvblue}{9});
    \STATE Update $\mathcal{P}_{sh}$, $\mathcal{A}^{t_c}_{sp}$, $g^{t_c}$;
\ENDFOR

\FOR{class $y \in \mathcal{Y}^{t_c}$}
    \STATE Compute $\mathcal{G}_{y,t_c}$ via feature statistics;
\ENDFOR
\STATE Build unified classifier $g^u$ (Eq.\textcolor{iccvblue}{8});
\end{algorithmic}
\end{algorithm}

\subsection{Optimization}

The overall framework is trained by minimizing:

\vspace{-2mm}
\begin{equation}\label{eq: total_loss}
    \mathcal{L} = \mathcal{L}_{ce} + \lambda_{fa} \mathcal{L}_{csfa} +\lambda_{ca} \mathcal{L}_{ca}.
    \vspace{-0.7mm}
\end{equation}

\noindent For simplicity, the trade-off parameters $\lambda_{fa}$ and $\lambda_{ca}$ for loss terms are set to 1.0. During each training session, our framework can be optimized in an end-to-end fashion. The overall training pipeline is outlined in Alg.\ref{alg::training}.

\begin{algorithm}[!t]
\caption{Inference pipeline after task $t_c$}  
\label{alg::inference}
\begin{algorithmic}[1]
\REQUIRE Test dataset $\{\mathbf{x}_i\}_{i=1}^{N}$, backbone $\mathcal{F}$, prompts $\mathcal{P}_{sh}$, adapters $\{\mathcal{A}^t_{sp}\}_{t=1}^{t_c}$, unified classifier $g^u$.
\ENSURE Predictions $\{\hat{k}_i\}_{i=1}^N$.
\FOR{$\mathbf{x}_i$ in $\{\mathbf{x}_i\}_{i=1}^N$}
    \STATE $\mathbf{f}_i = \mathcal{F}(\mathbf{x}_i; \mathcal{P}_{sh})$; \hfill \textcolor{iccvblue}{// Shared features}
    \STATE Compute task-confidence $\{P_t(t|\mathbf{f}_i)\}_{t=1}^{t_c}$ (Eq.\textcolor{iccvblue}{10}, \textcolor{iccvblue}{11});
    \STATE $\hat{\mathbf{z}}_i = \mathcal{F}(\mathbf{x}_i; \mathcal{P}_{sh}, \{\mathcal{A}^t_{sp}\}, \{\alpha^t_i\})$; \hfill \textcolor{iccvblue}{// TC-MoA aggregation}
    \STATE Predict label $\hat{k}_i$ via $g^u$ and $g$ (Eq.\textcolor{iccvblue}{13});
\ENDFOR
\end{algorithmic}
\end{algorithm}
\vspace{-4mm}

\subsection{Task-Confidence-guided Mixture of Adapters} 

Given the task-specific nature of our architecture,
it is essential to aggregate the task-specific knowledge from relevant sub-modules during inference without task-ids.
To achieve this objective, we leverage the task-shared knowledge from the task recognizer to perform an instance-adaptive aggregation for each testing image.
Given an image $\mathbf{x}_i$ and its shared feature $\mathbf{f}_i$, we denote the output of the classifier $g$ before softmax as $g(\mathbf{f}_i) \in \mathbb{R}^{K}$, and its $k$-th element as $g(\mathbf{f}_i)\left[k\right]$.
$K=\sum_{t=1}^{t_c} |\mathcal{Y}^t|$ is the total number of seen classes and $t_c$ is the current task identity.
To extract relevant information from each task, we only retain the classes that are semantically related to the testing image, while discarding others.
Therefore, we only keep the largest $K_c$ elements in $g(\mathbf{f}_i)$, which corresponds to the classes with the most similar semantics to the testing image:

\vspace{-2mm}
\begin{equation}\label{eq: tc-moa-topk}
\tilde{g}(\mathbf{f}_i)\left[k\right] = \left\{
\begin{array}{ll}
g(\mathbf{f}_i)\left[k\right], \quad \text{if } g(\mathbf{f}_i)\left[k\right] \in \text{Top-}K_c (g(\mathbf{f}_i)), 
\vspace{1mm} \\
-\infty, \quad \text{otherwise},
\end{array}\right.
\vspace{-0.2mm}
\end{equation}

\noindent where $\text{Top-}K_c(g(\mathbf{f}_i))$ denotes the top-$K_c$ highest elements in $g(\mathbf{f}_i)$ and $K_c \ll K$. 
Subsequently, we derive the confidence score $\alpha^t_i$ of $\mathbf{f}_i$ in belonging to task $t$ for $t \leq t_c$, representing the amount of semantic information relevant to the input contained within each task:

\vspace{0mm}
\begin{equation}\label{eq: tc-moa-soft-score}
\alpha^t_i = P_t(t | \mathbf{f}_i) = \frac{\sum_{k \in \mathcal{Y}^t}\exp(\tilde{g}(\mathbf{f}_i)[k] / \tau)}{\sum_{j=1}^{j=K}\exp(\tilde{g}(\mathbf{f}_i)[j] / \tau)},
\end{equation}
\vspace{-2mm}

\noindent 
where $\tau$ is a temperature to control the smoothness of the task-confidence scores.
Afterwards, the scores $\{\alpha_i^t\}_{t=1}^{t_c}$
serves as routing weights to aggregate task-specific intermediate features, forming a Mixture-of-Adapter (MoA) architecture.
At the $l$-th block, given the input $\mathbf{X}_{i,l}$ of $\mathbf{x}_i$, the forward process is formulated as follows:

\vspace{-2mm}
\begin{equation}\label{eq: tc-moa_forward}
\left\{
\begin{array}{ll}
[\hat{\mathbf{X}}_{i,l},\hat{\mathbf{P}}_l] =\text{MSA}([\mathbf{X}_{i,l}; \mathbf{P}_{l}]), \vspace{1.5mm}
\\
\mathbf{X}_{i,l+1} = \text{FFN}(\hat{\mathbf{X}}_{i,l}) + \sum_{t=1}^{T}\alpha^t_i \mathcal{A}^t_l(\hat{\mathbf{X}}_{i,l}).
\end{array}\right.
\end{equation}

Compared to assigning a single task to each test image, such a Mixture-of-Adapters (MoA) structure avoids overconfident aggregation, further enhancing the model's robustness against misleading task-ids.
The aggregated feature after the $L$-th block is utilized for final classification, denoted as $\hat{\mathbf{z}}_i=\mathcal{F}(\mathbf{x}_i; \mathcal{P}_{sh}, \{\mathcal{A}^t_{sp}\}_{t=1}^{t_c}, \{\alpha^t_i\}_{t=1}^{t_c})$.
The final prediction $\hat{k}_i$ is jointly determined by the global classifier $g^u$ and the training classifier $g$: 

\vspace{-3mm}
\begin{equation}\label{eq: cu-inference}
     \hat{k}_i = \arg\mathop{\max} \limits_{k} \left[\frac{t_c}{t_c+1}g^u(\hat{\mathbf{z}}_i)[k] + \frac{1}{t_c+1} g(\hat{\mathbf{z}}_i)[k]\right].
\end{equation} 

The inference of our CKAA includes deriving the task-confidence scores through the task-shared network and aggregating task-specific knowledge through the proposed TC-MoA.
The overall inference pipeline is illustrated in Alg.\ref{alg::inference}.

\begin{figure}[h]
\centering
\includegraphics[width=0.48\textwidth]{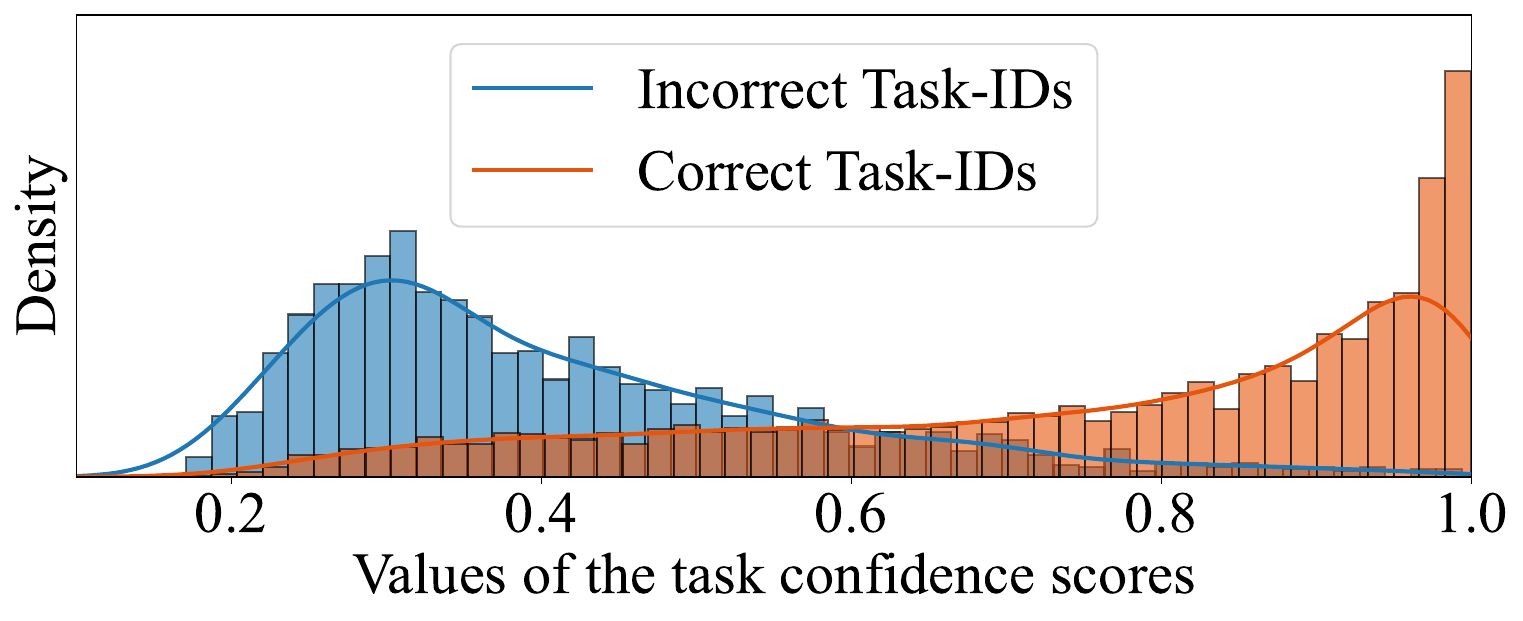}
\caption{Visualization of the distributions of the top-1 task-confidence scores.
}\label{fig:visualization_tcmoa}
\vspace{-3mm}
\end{figure}

\textbf{Analysis of the working mechanism of TC-MoA.} We visualize the top-1 confidence score distribution of samples with correct and incorrect task-ids in Fig.\ref{fig:visualization_tcmoa} on 10S-ImageNetA.
For a test sample $\mathbf{x}_i^{t}$ from task $t$, its top-1 confidence score is denoted as $\alpha^{\hat{t}}_i$, where $\hat{t} = \arg\mathop{\max} \limits_{i} \alpha_i^t$.
For $\mathbf{x}_i^{t}$, if the predicted top-1 task matches its ground-truth task id, where $\hat{t} == t$, the sample is considered to have a correct prediction of task id.

As shown in Fig.\ref{fig:visualization_tcmoa}, samples with correct task-ids generally have higher top-1 scores.
Therefore, for samples predicted with correct task-ids, TC-MoA assigns a single task-specific module with high confidence while neglecting others, acting as a \emph{\textbf{deterministic module selector}}.
For samples predicted with incorrect task-ids, TC-MoA adopts \emph{\textbf{a soft-label-like approach}}, softly compensating for the incorrect modules through intermediate features from potential correct modules.
Such a mechanism mitigates overconfidence in incorrect task-ids while preserving performance for correct ones, highlighting TC-MoA as a novel approach in handling incorrect task-ids in inference.

\section{Theoretical Analysis of the Proposed Method}
\label{sec:theory}

We provide a theoretical analysis to explain why CKAA improves robustness against misleading task-ID predictions.
The detailed derivations are provided in Appx.\textcolor{red}{A}. Consider the learning of task $r$, where $(\mathbf{x},y)\sim\mathcal{D}_r$ denotes an input-label pair from the task distribution $\mathcal{D}_r$. Let $\mathcal{F}(\cdot)$ be the pre-trained backbone, $\mathcal{P}_{sh}$ be the task-shared prompt, and $\mathcal{A}_{sp}^{t}$ be the task-specific adapter of task $t$. The feature obtained by routing $\mathbf{x}$ to task $t$ is denoted as $\mathbf{z}_t(\mathbf{x})=\mathcal{F}(\mathbf{x};\mathcal{P}_{sh},\mathcal{A}_{sp}^{t})$. Let $g(\cdot)$ be the classifier, $g_k(\cdot)$ be its logit for class $k$, and $h_g(\mathbf{z})=\arg\max_k g_k(\mathbf{z})$ be the prediction function. The risk of classifying samples from task $r$ through the subspace of task $s$ is defined as $R_{r\rightarrow s}(g)=\Pr_{(\mathbf{x},y)\sim\mathcal{D}_r}[h_g(\mathbf{z}_s(\mathbf{x}))\neq y]$. Therefore, the robustness objective of CKAA can be formulated as follows: 
\begin{equation}
\begin{aligned}
\mathcal{L}_{r}^{\rm robust}
& =
R_{r\rightarrow r}(g) +
\frac{1}{r-1}\sum_{s=1}^{r-1}
\left[
\lambda_1 R_{r\rightarrow s}(g)
+
\lambda_2 R_{s\rightarrow r}(g)
\right],
\label{eq:robust_objective}
\end{aligned}
\end{equation}
where $\lambda_1$ and $\lambda_2$ are trade-off hyper-parameters. The first term is optimized by standard classification training, while the latter two terms correspond to the new-to-old and old-to-new misrouting risks. CKAA reduces these two risks through feature-level alignment and classifier-level alignment, respectively.

\textbf{Feature-level alignment.}
For $(\mathbf{x},y)$, define the classification margins in the correct subspace $r$ and an incorrect historical subspace $s$ as $m_r(\mathbf{x},y)=g_y(\mathbf{z}_r(\mathbf{x}))-\max_{k\neq y}g_k(\mathbf{z}_r(\mathbf{x}))$ and $m_s(\mathbf{x},y)=g_y(\mathbf{z}_s(\mathbf{x}))-\max_{k\neq y}g_k(\mathbf{z}_s(\mathbf{x}))$, respectively. Assume that the logit difference $g_y(\mathbf{z})-g_k(\mathbf{z})$ is $L_g$-Lipschitz with respect to $\mathbf{z}$, where $L_g$ is the Lipschitz constant. Then, for any margin threshold $\gamma>0$, we derive an upper bound:
\begin{equation}\label{eq:fa_bound}
\begin{aligned}
R_{r\rightarrow s}(g)
& \leq
\Pr[m_r(\mathbf{x},y)<\gamma]
\\&+
\Pr\!\left[
\|\mathbf{z}_r(\mathbf{x})-\mathbf{z}_s(\mathbf{x})\|_2
\geq
\frac{\gamma}{L_g}
\right].
\end{aligned}
\end{equation}
Eq.~\eqref{eq:fa_bound} shows that reducing the new-to-old misrouting risk requires both enlarging the correct-subspace margin and reducing the cross-subspace feature discrepancy. Specifically, our CSFA loss can be rewritten as follows: 
\begin{equation} 
\begin{aligned} 
\mathcal{L}_{\rm csfa} &= \frac{1}{B} \sum_{i=1}^{B} \log \left[ \sum_{\mathbf{z}_j \in \mathcal{P}(\mathbf{z}_i)\cup \mathcal{N}(\mathbf{z}_i)} \exp \left( \frac{\operatorname{sim}(\mathbf{z}_i,\mathbf{z}_j)} {\tau_f} \right) \right] 
\\&- \frac{1}{B} \sum_{i=1}^{B} \log \left[ \sum_{\mathbf{z}^{+} \in \mathcal{P}(\mathbf{z}_i)} \exp \left( \frac{\operatorname{sim}(\mathbf{z}_i,\mathbf{z}^{+})} {\tau_f} \right) \right].
\end{aligned} 
\end{equation}
\noindent Minimizing $\mathcal{L}_{\rm csfa}$ suppresses the similarity between $\mathbf{z}_i$ and inter-class features $\mathbf{z}_j\in\mathcal{N}(\mathbf{z}_i)$, which encourages larger margins $m_r(\mathbf{x},y)$ in the correct subspace. Meanwhile, it maximizes the similarity between $\mathbf{z}_i$ and intra-class positives $\mathbf{z}^{+}\in\mathcal{P}(\mathbf{z}_i)$ from incorrect subspaces, reducing the cross-subspace discrepancy $\|\mathbf{z}_r(\mathbf{x})-\mathbf{z}_s(\mathbf{x})\|_2$. 
Therefore, $\mathcal{L}_{\rm csfa}$ jointly reduces the two terms in Eq.~\ref{eq:fa_bound}.

\textbf{Classifier-level alignment.}
For an old task $s<r$, historical samples are unavailable when learning task $r$. Let $\mathbf{f}=\mathcal{F}(\mathbf{x};\mathcal{P}_{sh})$ denote the task-shared feature of a historical sample, $\tilde{\mathbf{f}}_i$ denote a sampled historical feature from the class-wise Gaussian distribution, and $\tilde{\mathbf{z}}_i$ denote the simulated feature in the current subspace. Let $g^r$ be the task-adaptive classifier after learning task $r$, and let $L_{g^r}$ be the Lipschitz constant of its logit differences. Define the simulated margin as $\tilde{m}_{s\rightarrow r}(\tilde{\mathbf{z}},y)=g_y^r(\tilde{\mathbf{z}})-\max_{k\neq y}g_k^r(\tilde{\mathbf{z}})$. Then, for any $\gamma>0$, the old-to-new misrouting risk is bounded by:
\begin{equation}\label{eq:ca_bound}
\begin{aligned} 
R_{s\rightarrow r}(g^r) & \leq \Pr \left[ \tilde{m}_{s\rightarrow r} \bigl(\tilde{\mathbf{z}},y\bigr) < \gamma \right] \\ & + \frac{L_{g^r}}{\gamma} \Bigg\{ \left( 1+L_{\Delta} \right) \mathbb{E} \left[ \left\| \mathbf{f} - \tilde{\mathbf{f}}_i \right\|_2 \right]
\\ & + L_{\Delta} \mathbb{E} \left[ \frac{ \sum_{j=1}^{B} G_{ij} \left\| \tilde{\mathbf{f}}_i - \mathbf{f}_j \right\|_2 }{ \sum_{j=1}^{B}G_{ij} } \right] \Bigg\}. 
\end{aligned} 
\end{equation} 
The first term in Eq.~\ref{eq:ca_bound} is explicitly reduced by training $g^r$ with simulated features. The second term depends on the quality of feature simulation, which can be tightened via our Gaussian sampling and affinity-guided subspace transfer (the detailed analysis is provided in Appx.\textcolor{red}{A}).

\textbf{TC-MoA.} We further analyze why TC-MoA is more robust than deterministic maximum task selection. For a test sample $\mathbf{x}_i^t$ from task $t$, let $\alpha_i^{t'}$ be the task confidence assigned to task $t'$, and let $\hat{t}=\arg\max_{t'}\alpha_i^{t'}$ be the predicted task-ID. Define $p=\Pr(\hat{t}=t)$ as the task-ID accuracy, $\tau_1=\mathbb{E}[\alpha_i^t\mid\hat{t}=t]$ as the expected confidence of the ground-truth task when the task-ID is correct, and $\tau_2=\mathbb{E}[\alpha_i^t\mid\hat{t}\neq t]$ as the expected confidence of the ground-truth task when the top-1 task-ID is incorrect. Let $A_l^t(\cdot)$ be the adapter output of task $t$ at Transformer block $l$, $L$ be the number of blocks, $\mathbf{X}_{i,l}$ be the input tokens of $\mathbf{x}_i^t$ at block $l$, and $\rho_l$ be the Lipschitz constant that measures how perturbations at block $l$ affect the final prediction. The expected routing-induced error of TC-MoA is bounded by follows:
\begin{equation}
\begin{aligned}
\mathbb{E}[\mathcal{E}_{\rm TC\text{-}MoA}]
&\leq
\left[p(1-\tau_1)+(1-p)(1-\tau_2)\right]
\\&
\cdot \max_{\mathbf{x}_i^t,t'\neq t}
\sum_{l=1}^{L}
\rho_l
\|A_l^{t'}(\mathbf{X}_{i,l})-A_l^{t}(\mathbf{X}_{i,l})\|_2, \\
\mathbb{E}[\mathcal{E}_{\rm TC\text{-}MoA}]
&\leq
\mathbb{E}[\mathcal{E}_{\rm Max}]
\quad \text{if} \quad
p < \frac{\tau_2}{1-\tau_1+\tau_2}.
\end{aligned}
\label{eq:tcmoa_bound}
\end{equation}
Eq.~\eqref{eq:tcmoa_bound} explains why TC-MoA tightens the routing-error bound under ambiguous task boundaries. 

\begin{table}[H]
\centering
\caption{{Empirical verification of the condition for TC-MoA.}}
\label{tab: empirical_moa}
\begin{adjustbox}{max width=0.49\textwidth}
\begin{tabular}{l|c|ccccc}
\hline
Benchmark & Filtered & $p$ & $\tau_1$ & $\tau_2$ &
$\frac{\tau_2}{1-\tau_1+\tau_2}$ &  Satisfied \\
\hline
10S-ImageNetA & \ding{55} & 0.541 & 0.843 & 0.247 & 0.611 &
\textcolor{green!50!black}{\ding{51}~(0.611 $>$ 0.541)} \\
10S-ImageNetA & \ding{51} & 0.096 & 0.546 & 0.253 & 0.358 &
\textcolor{green!50!black}{\ding{51}~(0.358 $>$ 0.096)} \\
20S-ImageNetA & \ding{55} & 0.484 & 0.815 & 0.224 & 0.548 &
\textcolor{green!50!black}{\ding{51}~(0.548 $>$ 0.484)} \\
20S-ImageNetA & \ding{51} & 0.042 & 0.476 & 0.208 & 0.284 &
\textcolor{green!50!black}{\ding{51}~(0.284 $>$ 0.042)} \\
10S-ImageNetR & \ding{55} & 0.847 & 0.953 & 0.313 & 0.869 &
\textcolor{green!50!black}{\ding{51}~(0.869 $>$ 0.847)} \\
10S-ImageNetR & \ding{51} & 0.030 & 0.541 & 0.310 & 0.403 &
\textcolor{green!50!black}{\ding{51}~(0.403 $>$ 0.030)} \\
20S-ImageNetR & \ding{55} & 0.830 & 0.916 & 0.292 & 0.777 &
\textcolor{red}{\ding{55}~(0.777 $<$ 0.830)} \\
20S-ImageNetR & \ding{51} & 0.067 & 0.680 & 0.275 & 0.462 &
\textcolor{green!50!black}{\ding{51}~(0.462 $>$ 0.067)} \\
\hline
\end{tabular}
\end{adjustbox}
\end{table}

{We empirically verify whether the theoretical condition $p<\tau_2/(1-\tau_1+\tau_2)$ holds across five CIL benchmarks. We report the results in Tab.\ref{tab: empirical_moa}. ``Filtered'' denotes \emph{\textbf{excluding simple samples that can already be correctly classified using only the task-shared network}}. Such samples do not require additional task-specific information and are therefore not the primary cases affected by adapter-routing errors. The results demonstrate that TC-MoA tightens the routing-error for ambiguous samples whose predictions genuinely depend on task-specific information. Further details are in Appx.A.C.}


\setlength{\arrayrulewidth}{0.6pt}
\begin{table*}[t]
\centering
\caption{Experimental results on four CIL benchmarks. We report the averaged results over 3 trials. VPS-NSP$^{\dag}$ denotes our reproduction of the results under the same settings and seeds in our method. The highest results are in \textbf{bold}, and the second best results are \underline{underlined}.}\label{tab: compare-with-sota}
\vspace{-2mm}
\begin{adjustbox}{max width=1.0\textwidth}
\begin{tabular}{l|l|cc|cc|cc|cc}
\hline
\multirow{2}*{Method} &
\multirow{2}*{Venue} &
\multicolumn{2}{c|}{10s-ImageNetR} &
\multicolumn{2}{c|}{10s-ImageNetA} &
\multicolumn{2}{c|}{10s-CIFAR100} &
\multicolumn{2}{c}{10s-DomainNet}\\
\cline{3-10}
&{}
& Last-acc $\uparrow$ &Avg-acc $\uparrow$
& Last-acc $\uparrow$ &Avg-acc $\uparrow$
& Last-acc $\uparrow$ &Avg-acc $\uparrow$
& Last-acc $\uparrow$ &Avg-acc $\uparrow$\\
\hline
L2P \cite{l2pCVPR22}  & CVPR-22 &72.34\tsb{\(\pm\)0.17} & 77.36\tsb{\(\pm\)0.64} &44.04\tsb{\(\pm\)0.93} &51.24\tsb{\(\pm\)2.26} &84.06\tsb{\(\pm\)0.88} &88.26\tsb{\(\pm\)1.34}
&81.17\tsb{\(\pm\)0.83} &87.43\tsb{\(\pm\)0.95}\\
DualPrompt \cite{dualpromptECCV22}    & ECCV-22 & 69.10\tsb{\(\pm\)0.62} &  74.28\tsb{\(\pm\)0.66} &  53.19\tsb{\(\pm\)0.74} &64.59\tsb{\(\pm\)0.08} & 86.93\tsb{\(\pm\)0.24} &91.13\tsb{\(\pm\)0.32} &81.70\tsb{\(\pm\)0.78} &87.80\tsb{\(\pm\)0.99} \\
ADA \cite{adaICCV23} & NIPS-22 &73.76\tsb{\(\pm\)0.27} & 79.57\tsb{\(\pm\)0.84} &50.16\tsb{\(\pm\)0.20} &59.43\tsb{\(\pm\)2.20} &88.25\tsb{\(\pm\)0.26} &91.85\tsb{\(\pm\)1.32} & -- & --\\
CODAPrompt \cite{codaCVPR23} & CVPR-23 & 73.31\tsb{\(\pm\)0.50} & 78.47\tsb{\(\pm\)0.53} &52.08\tsb{\(\pm\)0.12} &  63.92\tsb{\(\pm\)0.12}  & 83.21\tsb{\(\pm\)3.39} &87.71\tsb{\(\pm\)3.17}  
&80.04\tsb{\(\pm\)0.79} &86.27\tsb{\(\pm\)0.82}\\
LAE \cite{laeICCV23}
  &ICCV-23 &  
72.29\tsb{\(\pm\)0.14}	&77.99\tsb{\(\pm\)0.46} &47.18\tsb{\(\pm\)1.17}	&58.15\tsb{\(\pm\)0.73} &85.25\tsb{\(\pm\)0.43} &89.80\tsb{\(\pm\)1.20}
& -- & --\\
SLCA \cite{slcaICCV23} & ICCV-23 & 79.35\tsb{\(\pm\)0.28}	&83.29\tsb{\(\pm\)0.46} & 61.05\tsb{\(\pm\)0.63}&	68.88\tsb{\(\pm\)2.31} & 91.26\tsb{\(\pm\)0.37}  & 94.29\tsb{\(\pm\)0.92} & -- & --\\
Adam-adapter \cite{adamIJCV2024} 
&IJCV-24 & 65.79\tsb{\(\pm\)0.98}	&72.42\tsb{\(\pm\)1.41}& 48.81\tsb{\(\pm\)0.08}  &58.84\tsb{\(\pm\)1.37}  &87.29\tsb{\(\pm\)0.27}&91.21\tsb{\(\pm\)1.33} & -- & --\\
RAPF \cite{rapfECCV24} & ECCV-24 & \multicolumn{1}{l}{\underline{79.62}} & \multicolumn{1}{l|}{\underline{86.28}} & -- & -- & \multicolumn{1}{l}{79.04} & \multicolumn{1}{l|}{86.19} & -- & --\\
InfLoRA \cite{infloraCVPR24} & CVPR-24 & 75.65\tsb{\(\pm\)0.14} & 80.82\tsb{\(\pm\)0.24} & 47.04\tsb{\(\pm\)0.90} & 56.91\tsb{\(\pm\)1.27} & 86.51\tsb{\(\pm\)0.73} & 91.70\tsb{\(\pm\)0.32} & 81.45\tsb{\(\pm\)0.68} & --\\
EASE~\cite{easeCVPR24} & CVPR-24 & 71.43\tsb{\(\pm\)0.18} & 78.04\tsb{\(\pm\)0.67} & 59.25\tsb{\(\pm\)0.88} &  68.92\tsb{\(\pm\)2.06} & 85.71\tsb{\(\pm\)0.76} & 90.96\tsb{\(\pm\)0.83} & -- & -- \\
CPrompt \cite{cpromptCVPR24} & CVPR-24 & 77.14\tsb{\(\pm\)0.11} & 82.92\tsb{\(\pm\)0.70} & 55.15\tsb{\(\pm\)0.85} & 65.49\tsb{\(\pm\)0.52} & 87.82\tsb{\(\pm\)0.21} & 92.53\tsb{\(\pm\)0.23} & 82.97\tsb{\(\pm\)0.34} & 88.54\tsb{\(\pm\)0.41}\\
SSIAT \cite{ssaitCVPR24}  & CVPR-24 & 79.38\tsb{\(\pm\)0.59} & 83.63\tsb{\(\pm\)0.43} & 
\underline{62.43}\tsb{\(\pm\)1.63} &  70.83\tsb{\(\pm\)1.63} & \underline{91.35}\tsb{\(\pm\)0.26} & \underline{94.35}\tsb{\(\pm\)0.60} & \underline{85.11}\tsb{\(\pm\)0.56} & \underline{89.80}\tsb{\(\pm\)0.34}\\
VQ-Prompt~\cite{vqprompt-nips24} & NeurIPS-24 & 78.71\tsb{\(\pm\)0.22} & 83.24\tsb{\(\pm\)0.68} & -- & -- & 88.73\tsb{\(\pm\)0.27} &  92.84\tsb{\(\pm\)0.73} & -- & -- \\
VPT-NSP$^{\dag}$ \cite{vptnspNIPS-24} & NeurIPS-24 & 77.95\tsb{\(\pm\)0.22} & 83.44\tsb{\(\pm\)0.40} & 53.83\tsb{\(\pm\)0.37} & 63.93\tsb{\(\pm\)1.08} & 90.19\tsb{\(\pm\)0.43} & 94.12\tsb{\(\pm\)1.05} &83.05\tsb{\(\pm\)1.06} & 87.83\tsb{\(\pm\)0.92}  \\
DIA~\cite{DIA-cvpr25} & CVPR-25 
& \multicolumn{1}{l}{79.03} & \multicolumn{1}{l|}{85.61}  & \multicolumn{1}{l}{61.69} & \multicolumn{1}{l|}{\underline{71.58}} & \multicolumn{1}{l}{90.80} & \multicolumn{1}{l|}{94.29} & -- & -- \\
ACMap~\cite{ACMap-cvpr25} & CVPR-25 & 
\multicolumn{1}{l}{70.49} & \multicolumn{1}{l|}{77.31} & \multicolumn{1}{l}{56.19} & \multicolumn{1}{l|}{65.19} & \multicolumn{1}{l}{87.81} & \multicolumn{1}{l|}{92.04} & -- & -- \\
SEMA~\cite{sema-cvpr25} & CVPR-25 & 
\multicolumn{1}{l}{78.00} & 
\multicolumn{1}{l|}{83.56} & 
\multicolumn{1}{l}{53.32} & 
\multicolumn{1}{l|}{64.53} & 
\multicolumn{1}{l}{86.98} & 
\multicolumn{1}{l|}{91.37} & 
-- & -- \\
LoRA-DRS~\cite{lora-drs-cvpr25} & CVPR-25 & 74.74\tsb{\(\pm\)0.78} & 81.16\tsb{\(\pm\)0.59} & -- & -- & 89.14\tsb{\(\pm\)0.23} & 92.55\tsb{\(\pm\)0.25} & -- & -- \\
CL-LoRA~\cite{cllora-cvpr25} & CVPR-25 & -- & -- & 60.54\tsb{\(\pm\)0.63} & 70.15\tsb{\(\pm\)2.23} & 85.32\tsb{\(\pm\)0.08} & 91.02\tsb{\(\pm\)0.12} & -- & -- \\
\hline
\textbf{CKAA (Ours)}   & -- & \textbf{82.41}\tsb{\(\pm\)0.13} & \textbf{87.20}\tsb{\(\pm\)0.35}
& \textbf{63.53}\tsb{\(\pm\)0.66} &  \textbf{71.11}\tsb{\(\pm\)1.03} &\textbf{92.26}\tsb{\(\pm\)0.39} & \textbf{95.18}\tsb{\(\pm\)0.86} & \textbf{86.72}\tsb{\(\pm\)0.55} & \textbf{90.68}\tsb{\(\pm\)0.78} \\ 
\hline
\end{tabular}
\end{adjustbox}
\vspace{-2.0mm}
\end{table*}

\begin{table}[t]
\centering
\caption{Experimental results under the long-term setting (20 incremental tasks) on ImageNetR and ImageNetA. }\label{tab: compare-with-sota-1}
\begin{adjustbox}{max width=0.48\textwidth}
\begin{tabular}{l|cc|cc}
\hline
\multirow{2}*{Method} &
\multicolumn{2}{c|}{20s-ImageNetR} &
\multicolumn{2}{c}{20s-ImageNetA}\\
\cline{2-5}
& Last-acc $\uparrow$ &Avg-acc $\uparrow$
& Last-acc $\uparrow$ &Avg-acc $\uparrow$ \\
\hline
L2P \cite{l2pCVPR22} & 69.64\tsb{\(\pm\)0.42}	& 75.28\tsb{\(\pm\)0.57} & 40.48\tsb{\(\pm\)1.78} & 49.62\tsb{\(\pm\)1.46} \\
DualPrompt \cite{dualpromptECCV22} & 66.61\tsb{\(\pm\)0.58} & 72.45\tsb{\(\pm\)0.37} & 42.28\tsb{\(\pm\)1.94} & 53.39\tsb{\(\pm\)1.64} \\
CODAPrompt \cite{codaCVPR23} & 69.96\tsb{\(\pm\)0.50} & 75.34\tsb{\(\pm\)0.85} & 44.62\tsb{\(\pm\)1.92} & 54.86\tsb{\(\pm\)0.50} \\
LAE \cite{laeICCV23} & 69.86\tsb{\(\pm\)0.43} & 77.38\tsb{\(\pm\)0.61} & 39.52\tsb{\(\pm\)0.78} & 51.75\tsb{\(\pm\)2.15}\\
SLCA \cite{slcaICCV23} & 74.63\tsb{\(\pm\)1.55}& 79.92\tsb{\(\pm\)1.29} &  36.69\tsb{\(\pm\)21.31} & 56.35\tsb{\(\pm\)7.09} \\
Adam-adapter \cite{adamIJCV2024} &57.42\tsb{\(\pm\)0.84}  & 64.75\tsb{\(\pm\)0.79}  &  48.65\tsb{\(\pm\)0.12} & 59.55\tsb{\(\pm\)1.07} \\
InfLoRA \cite{infloraCVPR24}  & 71.01\tsb{\(\pm\)0.45} & 77.28\tsb{\(\pm\)0.45} & -- & -- \\
CPrompt \cite{cpromptCVPR24} & 74.79\tsb{\(\pm\)0.28} & 81.46\tsb{\(\pm\)0.93} & -- & -- \\
SSIAT \cite{ssaitCVPR24} & 75.67\tsb{\(\pm\)0.14} & \underline{82.30}\tsb{\(\pm\)0.36} & \underline{59.16}\tsb{\(\pm\)1.03} & \underline{68.45}\tsb{\(\pm\)1.92} 
\\
VPT-NSP$^{\dag}$ \cite{vptnspNIPS-24} & {\underline{75.69}\tsb{\(\pm\)0.61}} & 81.87\tsb{\(\pm\)0.59} & 49.81\tsb{\(\pm\)1.29} & 61.41\tsb{\(\pm\)1.84} \\
DIA~\cite{DIA-cvpr25} & \multicolumn{1}{l}{76.32} & \multicolumn{1}{l|}{83.51} & -- & -- \\
\hline
\textbf{CKAA (Ours)} & \textbf{80.98}\tsb{\(\pm\)0.54} & \textbf{86.28}\tsb{\(\pm\)0.18} & \textbf{60.54}\tsb{\(\pm\)0.50} & \textbf{70.55}\tsb{\(\pm\)0.64} \\ 
\hline
\end{tabular}
\end{adjustbox}
\vspace{-2mm}
\end{table}

\begin{table}[t]
\centering
\caption{Results under the long-term setting (50 \& 100 tasks) on ImageNetR and DomainNet. Last-acc is reported.}\label{tab: long-term}
\begin{adjustbox}{max width=0.48\textwidth}
\begin{tabular}{l|l|l|l|l}
\hline
\multirow{2}*{Method} & 
\multicolumn{2}{c|}{S-ImageNetR} &
\multicolumn{2}{c}{S-DomainNet}\\
\cline{2-5}
& \multicolumn{1}{c|}{50s} & \multicolumn{1}{c|}{100s} & \multicolumn{1}{c|}{50s} & \multicolumn{1}{c}{100s}\\
\cline{2-5}
\hline
L2P\cite{l2pCVPR22} & 51.38 & 41.51 & 63.13 & 54.83 \\
OVOR-Deep\cite{ovorICLR24} & 63.25 & 43.02 & 68.29 & 52.09\\
ConvPrompt\cite{convpromptCVPR24} & 64.61 & 44.32 & 71.76 & 56.21\\
InfLoRA\cite{infloraCVPR24} & 62.81 & 42.23 & \underline{71.87} &48.06 \\
Cprompt\cite{cpromptCVPR24} & \underline{70.75} & 59.90 & 70.74 & \underline{57.60}\\
EASE \cite{easeCVPR24} & 70.27 & 51.56 & 65.34 & 37.56\\
VPT-NSP$^{\dag}$ \cite{vptnspNIPS-24} & 69.48 & \underline{62.23} & 71.28 & 57.35 \\
\hline
\textbf{CKAA (Ours)} & \textbf{78.20}\tsb{\(\pm\)0.65} & \textbf{73.41}\tsb{\(\pm\)0.51} & \textbf{81.53}\tsb{\(\pm\)2.60} & \textbf{77.28}\tsb{\(\pm\)3.09}\\
\hline
\end{tabular}
\end{adjustbox}
\vspace{-3mm}
\end{table}

\section{Experiments}

\noindent \textbf{Dataset and Evaluation Metrics.} We evaluate our proposed method on four public CIL benchmarks, namely ImagenetR \cite{imagenet-r}, ImagenetA \cite{imagenet-a}, CIFAR100 \cite{cifar100} and DomainNet \cite{domainet}.
ImagenetR comprises 30000 images of 200 classes, which share the same class names with ImegeNet-21K \cite{imagenet-21k} but belong to different domains.
ImagenetA is a challenging dataset with 7500 images, which contains hard samples and exhibits significant class imbalance.
Cifar100 is a commonly-used dataset in continual learning, which consists of 60000 32$\times$32 images of 100 classes.
DomainNet is a cross-domain dataset including images from 345 classes and 6 diverse domains.
Following \cite{cpromptCVPR24, vptnspNIPS-24}, the 200 categories with the maximum number of images are selected for training and evaluation.
We randomly split the datasets into 10 or 20 non-overlapping tasks.
Note that for DomainNet, the 200 categories with the maximum number of images are selected for training and evaluation following \cite{cpromptCVPR24, vptnspNIPS-24,esnAAAI23}.

Following existing PEFT-based CIL methods \cite{ssaitCVPR24, cpromptCVPR24}, we evaluate the model performance through two commonly-used metrics: the accuracy of all classes after learning the last session (Last-acc), and the averaged accuracy of all incremental sessions (Avg-acc). 
We conduct all experiments under three different random seeds.
The average performance and the maximum deviation are reported.

\noindent \textbf{Implementation Details.}
Experiments are conducted on a single NVIDIA RTX 3090 GPU.
We use ViT-B/16 \cite{ViT} as our backbone, which is pre-trained on ImageNet-21K \cite{imagenet-21k}. 
The Adam optimizer with an initial learning rate of 0.01 is employed for training. 
We train each task for 100 epochs and the mini-batch is 110.
Following \cite{vptnspNIPS-24, ssaitCVPR24, infloraCVPR24}, the PEFT modules are inserted into all transformer blocks to avoid searching.
The prompt length is 4 and the width of adapters are 64. 
In DKA, $\tau_f$ in Eq.\ref{eq: fu-cscl-loss} is set to 0.05, $\tau_g$ and $K_g$ in Eq.\ref{eq: cu-build-graph} are 0.2 and 20.
In TC-MoA, $K_c$ in Eq.\ref{eq: tc-moa-topk} is set to 20 for ImageNetA and 10 for others,
and $\tau$ in Eq.\ref{eq: tc-moa-soft-score} is set to 3.0.

\subsection{Comparison with the State-of-the-Arts}

We compare our CKAA with state-of-the-art PEFT-based CIL methods, as shown in Tab.\ref{tab: compare-with-sota}, Tab.\ref{tab: compare-with-sota-1} and Tab.\ref{tab: long-term}.
We compare our method with prompt-based methods \cite{l2pCVPR22, codaCVPR23, dualpromptECCV22, vptnspNIPS-24}, adapter-based methods \cite{ssaitCVPR24, adamIJCV2024, adaICCV23} and other PEFT-based methods \cite{rapfECCV24, infloraCVPR24}. 
Tab.\ref{tab: compare-with-sota} shows the results of the standard 10-split incremental setting on four commonly-used benchmarks.
Notably, our proposed method outperforms the state-of-the-art method SSAIT by 3.03\%, 1.10\%, 0.91\%, and 1.61\% in Last-acc on four datasets, respectively.

To further demonstrate the scalability of our method under long-term settings, we conduct experiments under the more challenging settings of 20, 50, and 100 tasks on ImagenetR, ImageNetA, and DomainNet.
The results are shown in Tab.\ref{tab: compare-with-sota-1} and Tab.\ref{tab: long-term}.
Under the 20-split setting, our method outperforms SSIAT by 5.31\% and 1.38\% in Last-acc on two datasets.
While under the 50-split and 100-split settings, our proposed CKAA exhibits superior performances, which surpasses VPT-NSP by 8.71\% and 11.18\% in Last-Acc on the ImageNetR dataset.
For the 50-split and 100-split settings, the intermediate dimensions of adapters are set to 16 and 8 to avoid additional inference overhead.
The results highlight its scalability and ability to address forgetting in long-term CIL problems.






\begin{table}[t]
\centering
\caption{Experimental results on 10S-CUB-200. }\label{tab: cub200}
\begin{adjustbox}{max width=0.50\textwidth}
\begin{tabular}{l|l|cc}
\hline
\multirow{1}*{Method} & \multirow{1}*{Venue}
& Last-acc $\uparrow$ &Avg-acc $\uparrow$\\
\hline
L2P \cite{l2pCVPR22} & CVPR-22 & 67.02\tsb{\(\pm\) 1.90} &79.62\tsb{\(\pm\) 1.60}\\
DualPrompt \cite{dualpromptECCV22} & ECCV-22 & 68.48\tsb{\(\pm\) 0.47} &80.59\tsb{\(\pm\) 1.50}\\
CODAPrompt \cite{codaCVPR23} & CVPR-23 & 77.23\tsb{\(\pm\) 1.12} &81.90\tsb{\(\pm\) 0.85} \\
LAE \cite{laeICCV23} & ICCV-23 & 80.97\tsb{\(\pm\) 0.51}	& 87.22\tsb{\(\pm\) 1.21} \\
SLCA \cite{slcaICCV23} & ICCV-23 & 84.68\tsb{\(\pm\) 0.09}&	90.77\tsb{\(\pm\) 0.79} \\
Adam-adapter \cite{adamIJCV2024} & IJCV-24 & 85.84\tsb{\(\pm\) 0.08} &	91.33\tsb{\(\pm\)0.49} \\
EASE \cite{easeCVPR24} & CVPR-24 & \multicolumn{1}{l}{86.81} & \multicolumn{1}{l}{92.23} \\
SSIAT \cite{ssaitCVPR24} & CVPR-24 & \underline{88.75}\tsb{\(\pm\) 0.38} &\underline{93.00}\tsb{\(\pm\) 0.90} \\
VPT-NSP$^{\dag}$ \cite{vptnspNIPS-24} & NIPS-24 & 83.08\tsb{\(\pm\) 0.75} & 90.32\tsb{\(\pm\) 0.62}\\
CoFiMA \cite{CoFiMAECCV24} & ECCV-24 & 87.11\tsb{\(\pm\)0.56} & 91.87\tsb{\(\pm\)0.69}\\
DIA~\cite{DIA-cvpr25} & CVPR-25 & \multicolumn{1}{l}{86.73} & \multicolumn{1}{l}{93.21} \\
ACMap~\cite{ACMap-cvpr25} & CVPR-25 & \multicolumn{1}{l}{86.66} & \multicolumn{1}{l}{91.56} \\
\hline
\textbf{CKAA (Ours)} & -- & \textbf{89.48}\tsb{\(\pm\)0.46} &  \textbf{93.34}\tsb{\(\pm\)0.26} \\ 
\hline
\end{tabular}
\end{adjustbox}
\vspace{-2mm}
\end{table}


\begin{table}[t]
\centering
\caption{Experimental results on 10S-StanfordCars. }\label{tab: stanfordcars}
\begin{adjustbox}{max width=0.48\textwidth}
\begin{tabular}{l|l|cc}
\hline
\multirow{1}*{Method} & \multirow{1}*{Venue}
& Last-acc $\uparrow$ &Avg-acc $\uparrow$\\
\hline
L2P \cite{l2pCVPR22} & CVPR-22 & 60.39\tsb{\(\pm\)1.99} & 71.92\tsb{\(\pm\)1.12}\\
DualPrompt \cite{dualpromptECCV22} & ECCV-22 & 57.27\tsb{\(\pm\)0.34} & 70.36\tsb{\(\pm\)2.33}\\
ESN \cite{esnAAAI23} & AAAI-23 & 56.91\tsb{\(\pm\)0.56} & 72.82\tsb{\(\pm\)0.79} \\
CODAPrompt \cite{codaCVPR23} & CVPR-23 & 62.24\tsb{\(\pm\)0.14} & 73.28\tsb{\(\pm\)0.93} \\
SLCA \cite{slcaICCV23} & ICCV-23 & 67.73\tsb{\(\pm\)0.85} & 76.93\tsb{\(\pm\)1.21} \\
CPrompt \cite{cpromptCVPR24} & CVPR-24 & 66.77\tsb{\(\pm\)0.37} & 76.81\tsb{\(\pm\)0.27} \\
SSIAT \cite{ssaitCVPR24} & CVPR-24 & 75.32\tsb{\(\pm\)0.45} & 82.87\tsb{\(\pm\)0.68} \\
CoMA \cite{CoFiMAECCV24} & ECCV-24 & 73.35\tsb{\(\pm\)0.59} & 78.55\tsb{\(\pm\)0.42} \\ 
CoFiMA \cite{CoFiMAECCV24} & ECCV-24 & 76.96\tsb{\(\pm\)0.64} & 82.65\tsb{\(\pm\)0.96} \\
HiDe-PreT \cite{wang2025hide} & TPAMI-25 
& 68.75\tsb{\(\pm\)1.01} & 69.62\tsb{\(\pm\)0.38} \\
$\alpha_{2}$-SLDC \cite{SLDC-AAAI26} & AAAI-26 & 74.07\tsb{\(\pm\)0.78} & 83.32\tsb{\(\pm\)0.55} \\ 
$\beta_{1}$-SLDC+ADE \cite{SLDC-AAAI26} & AAAI-26 & \underline{80.61}\tsb{\(\pm\)0.36} & \underline{85.51}\tsb{\(\pm\)0.41} \\
\hline
\textbf{CKAA (Ours)} & -- & \textbf{80.92}\tsb{\(\pm\)0.08} &  \textbf{87.32}\tsb{\(\pm\)0.90} \\ 
\hline
\end{tabular}
\end{adjustbox}
\end{table}

\begin{table*}[t]
\centering
\caption{Ablation Studies of each component in our proposed method on ImageNetR and ImageNetA. }\label{tab: ablation-study}
\begin{adjustbox}{max width=1.0\textwidth}
\begin{tabular}{c|cccc|cc|cc|cc}
\hline
\multirow{2}*{Idx} & 
\multirow{2}*{$\mathcal{P}_{sh}$} &
\multirow{2}*{TC-MoA} & 
\multirow{2}*{FA} & 
\multirow{2}*{CA} &
\multicolumn{2}{c|}{10S-ImageNetR} &
\multicolumn{2}{c|}{10S-ImageNetA} & 
\multicolumn{2}{c}{20S-ImageNetR}\\
\cline{6-11}
& {} & {} & {} & {}
& Last-acc $\uparrow$ &Avg-acc $\uparrow$
& Last-acc $\uparrow$ &Avg-acc $\uparrow$
& Last-acc $\uparrow$ &Avg-acc $\uparrow$\\
\hline
\multicolumn{11}{l}{\textit{CKAA with frozen pre-trained task-shared network.}}\\
\hline
1& -- & -- & -- & -- & 65.65\tsb{\(\pm\)0.58} & 72.82\tsb{\(\pm\)0.67} & 46.01\tsb{\(\pm\)0.92} & 56.51\tsb{\(\pm\)0.70} & 63.67\tsb{\(\pm\)0.68} & 71.56\tsb{\(\pm\)1.14}\\
2& -- & \checkmark & -- & -- & 76.72\tsb{\(\pm\)0.61} & 83.14\tsb{\(\pm\)0.32} & 53.92\tsb{\(\pm\)0.85} & 64.16\tsb{\(\pm\)0.18} & 73.91\tsb{\(\pm\)0.35} & 81.15\tsb{\(\pm\)0.32}\\
3& -- & \checkmark & \checkmark & -- & 78.54\tsb{\(\pm\)0.39} & 84.36\tsb{\(\pm\)0.28} &  57.12\tsb{\(\pm\)1.75} & 65.68\tsb{\(\pm\)0.25} & 75.58\tsb{\(\pm\)0.19} & 82.43\tsb{\(\pm\)0.76} \\
4& -- & \checkmark & -- & \checkmark & 79.77\tsb{\(\pm\)0.12} & 85.09\tsb{\(\pm\)0.24} & 59.71\tsb{\(\pm\)0.20} & 68.55\tsb{\(\pm\)0.32}  & 78.65\tsb{\(\pm\)0.27} & 84.52\tsb{\(\pm\)0.41}\\
5& -- & \checkmark & \checkmark & \checkmark & \textbf{81.56}\tsb{\(\pm\)0.13} & \textbf{86.63}\tsb{\(\pm\)0.13} & \textbf{62.49}\tsb{\(\pm\)1.53} & \textbf{70.59}\tsb{\(\pm\)0.56} & \textbf{79.86}\tsb{\(\pm\)0.32} & \textbf{85.33}\tsb{\(\pm\)0.52}\\
\hline
\multicolumn{11}{l}{\textit{CKAA with learnable visual prompt-based task-shared network.}}\\
\hline
6& \checkmark & -- & -- & -- & 77.95\tsb{\(\pm\)0.22} & 83.44\tsb{\(\pm\)0.40} & 53.83\tsb{\(\pm\)0.37} & 63.93\tsb{\(\pm\)1.08} & 75.69\tsb{\(\pm\)0.62} & 81.87\tsb{\(\pm\)0.56}\\
7 & \checkmark & \checkmark & -- & -- & 79.14\tsb{\(\pm\)0.21} & 85.04\tsb{\(\pm\)0.33} & 54.33\tsb{\(\pm\)0.87} & 64.43\tsb{\(\pm\)1.22} & 76.88\tsb{\(\pm\)0.57} & 82.91\tsb{\(\pm\)0.52}\\
8 & \checkmark & \checkmark & \checkmark & -- & 79.74\tsb{\(\pm\)0.34} & 85.29\tsb{\(\pm\)0.19} & 58.13\tsb{\(\pm\)0.20} & 66.59\tsb{\(\pm\)0.84} & 77.82\tsb{\(\pm\)0.55} & 83.71\tsb{\(\pm\)0.39}\\
9 & \checkmark & \checkmark & -- & \checkmark & 81.80\tsb{\(\pm\)0.38} & 86.59\tsb{\(\pm\)0.18} & 61.29\tsb{\(\pm\)0.46} & 69.31\tsb{\(\pm\)1.16} & 80.18\tsb{\(\pm\)0.42} & 85.75\tsb{\(\pm\)0.24}\\
10& \checkmark & \checkmark & \checkmark & \checkmark & 
\textbf{82.41}\tsb{\(\pm\)0.13} & \textbf{87.20}\tsb{\(\pm\)0.35} & \textbf{63.53}\tsb{\(\pm\)0.66} &  \textbf{71.11}\tsb{\(\pm\)1.03} &
\textbf{80.98}\tsb{\(\pm\)0.35} & \textbf{86.28}\tsb{\(\pm\)0.18}\\
\hline
\end{tabular}
\end{adjustbox}
\vspace{-3mm}
\label{performance}
\end{table*}

\subsection{Comparison with Existing Methods on Fine-Grained Datasets}

We evaluate our method on the fine-grained datasets CUB-200 \cite{CUB}
and StanfordCars \cite{stanfordcars2013} to demonstrate its effectiveness in fine-grained CIL classification.
CUB-200 \cite{CUB} contains 11,788 images of 200 bird species with fine-grained class labels.
StanfordCars \cite{stanfordcars2013} is a fine-grained car dataset including 196 classes and 16185 car images. 
Following existing methods \cite{ssaitCVPR24, adamIJCV2024, cpromptCVPR24}, we evaluate our method under the 10-split setting.
The experiment results on CUB-200 and StanfordCars are outlined in Tab.\ref{tab: cub200} and Tab.\ref{tab: stanfordcars}, respectively.
The results demonstrate the robustness of our proposed CKAA against the catastrophic forgetting of fine-grained knowledge.

\setlength{\arrayrulewidth}{0.8pt}
\begin{table}[t]
\centering
\caption{Ablation studies of various task-specific knowledge aggregation strategies without and with DKA on two datasets.}\label{tab: ablation-tc-moa}
\vspace{-2mm}
\begin{adjustbox}{max width=0.50\textwidth}
\begin{tabular}{l|cc|cc}
\hline
\multirow{2}*{Method} & 
\multicolumn{2}{c|}{10S-ImageNetR} & \multicolumn{2}{c}{10S-ImageNetA} \\
\cline{2-5}
& Last-acc $\uparrow$ &Avg-acc $\uparrow$ & Last-acc $\uparrow$ &Avg-acc $\uparrow$
\\
\hline
\multicolumn{5}{l}{\textit{Without Dual-level Knowledge Alignment (DKA)}} \\
\hline
Query-key & 74.60\tsb{\(\pm\)0.58} & 80.75\tsb{\(\pm\)0.82} & 47.30\tsb{\(\pm\)0.88} & 58.21\tsb{\(\pm\)0.76} \\
Maximum & 77.03\tsb{\(\pm\)0.46} & 82.95\tsb{\(\pm\)0.98} & 49.18\tsb{\(\pm\)0.72} & 59.57\tsb{\(\pm\)1.43} \\
\textbf{TC-MoA} & 
\textbf{79.14}\tsb{\(\pm\)0.21}& \textbf{85.04}\tsb{\(\pm\)0.33}&
\textbf{54.33}\tsb{\(\pm\)0.87}& \textbf{64.43}\tsb{\(\pm\)1.22} \\
\hline
\multicolumn{5}{l}{\textit{With Dual-level Knowledge Alignment (DKA)}}
\\
\hline
Query-key & 76.02\tsb{\(\pm\)0.45} & 81.88\tsb{\(\pm\)0.67} & 56.39\tsb{\(\pm\)0.37} & 65.80\tsb{\(\pm\)1.28}\\
Maximum & 78.55\tsb{\(\pm\)0.25} & 84.20\tsb{\(\pm\)0.72} & 58.12\tsb{\(\pm\)0.57} & 67.53\tsb{\(\pm\)0.56}\\
\textbf{TC-MoA} 
& \textbf{82.41}\tsb{\(\pm\)0.13} 
& \textbf{87.20}\tsb{\(\pm\)0.35}
& \textbf{63.53}\tsb{\(\pm\)0.66} 
& \textbf{71.11}\tsb{\(\pm\)1.03}\\
\hline
\end{tabular}
\end{adjustbox}
\label{performance}
\end{table}

\begin{table}[t]
\centering
\caption{Ablation studies of feature simulation on 10S-ImageNetA.}\label{tab: ablation-cu}
\begin{adjustbox}{max width=0.50\textwidth}
\begin{tabular}{l|cc}
\hline
\multirow{1}*{Method} 
& Last-acc $\uparrow$ &Avg-acc $\uparrow$
\\
\hline
\textbf{CKAA (Ours)} & \textbf{63.53}\tsb{\(\pm\)0.66} & \textbf{71.11}\tsb{\(\pm\)1.03} \\
\hline
\multicolumn{3}{l}{\textit{The following methods are based on FA and TC-MoA.}}\\
\hline
CKAA w/o CA & 58.13\tsb{\(\pm\)0.20} & 66.59\tsb{\(\pm\)0.84}\\
+ CA w/o subspace transfer & 62.38\tsb{\(\pm\)0.70} & 70.35\tsb{\(\pm\)1.28}\\
+ CA w/o affinities $\mathbf{G}$ & 63.20\tsb{\(\pm\)0.63} & 70.83\tsb{\(\pm\)0.92} \\
\hline
\end{tabular}
\end{adjustbox}
\vspace{-3mm}
\end{table}

\subsection{Ablation Study}

\begin{figure}[!t]
\centering
\includegraphics[width=0.50\textwidth]{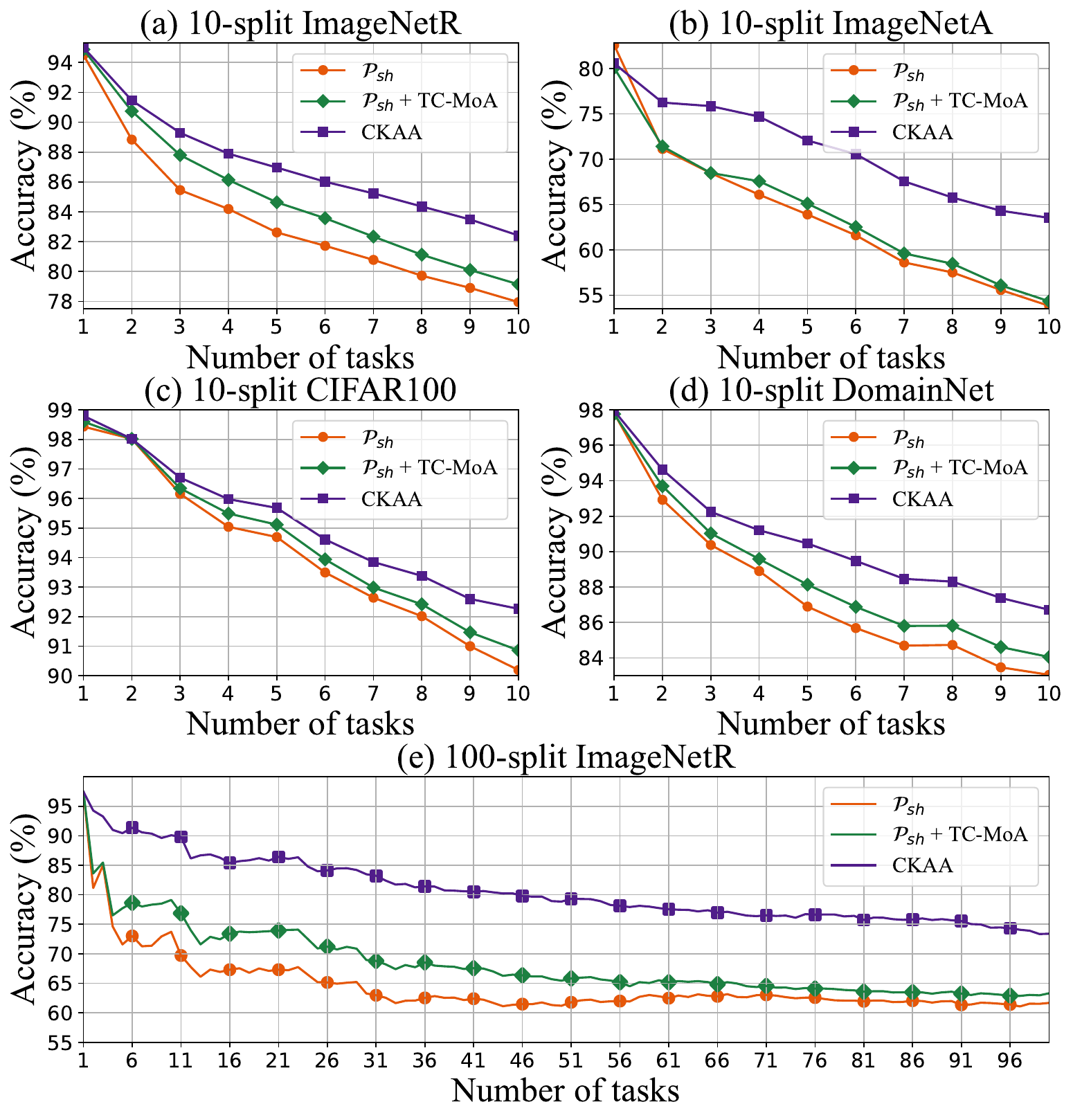}
\vspace{-4mm}
\caption{The task-by-task accuracy changing curves of our baseline ($\mathcal{P}_{sh}$), $\mathcal{P}_{sh}$+TC-MoA and CKAA on four benchmarks.
}\label{fig:ablation}
\vspace{-3mm}
\end{figure}

In this subsection, we perform ablation studies on the components of our framework on ImageNetR and ImageNetA datasets with 10- or 20-split tasks, as shown in Tab.\ref{tab: ablation-study}.
We also visualize the task-by-task accuracy changing curves of the ablation studies on four benchmarks with 10- or 100-split tasks, as shown in Fig.\ref{fig:ablation}.
Our baseline is the frozen ViT backbone with a trainable classifier (Idx 1) and Fig.\ref{fig:ablation}.


\textbf{The effectiveness of $\mathcal{P}_{sh}$.}
We conduct experiments on the learnable visual prompt-based task-shared network to demonstrate its effectiveness.
Specifically, we directly replace the learnable task-shared network with the frozen pre-trained backbone.
As indicated in Tab.\ref{tab: ablation-study},
the learnable visual prompts $\mathcal{P}_{sh}$ improves the model's Last-acc by 0.85\% and Avg-acc by 0.57\% on 10S-ImagenetR (Idx 5 $v.s.$ Idx 10).
This improvement is mainly due to the learnable task-shared network’s ability to capture common knowledge across downstream datasets, compared to a frozen pre-trained backbone.
It results in more precise task identification during inference and more appropriate routing weights for aggregating task-specific knowledge in TC-MoA.

\textbf{The effectiveness of DKA.}
The proposed DKA (FA \& CA) yields an overall improvement of 3.27\%$\sim$9.20\% on Last-acc compared to the independent training of sub-modules.
The Feature-level Alignment (FA) improves the performance by 0.60\%$\sim$3.80\% Last-acc (Idx 7 $v.s.$ Idx 8).
Our FA effectively aligns feature semantics within different subspaces, resulting in a performance gain.
The Classifier-level Alignment (CA) further substantially improves the model performance of 2.67\%$\sim$5.40\% on Last-acc (Idx 8 $v.s.$ Idx 10). 
Unlike existing methods that use features with the proper task-ids to train the classifier, our CA simulates the wrongly projected features to train a unified classifier, improving the classifier's robustness.

Tab.\ref{tab: ablation-tc-moa} further shows the performance \textbf{\emph{without and with DKA under various task-specific knowledge aggregation strategies}}.
In Tab.\ref{tab: ablation-tc-moa}, 
``Query-key'' matching designs a task-key for each task-specific sub-module and selects the adapter corresponding to the task-key with the highest similarity to the input for integrating task-specific knowledge, which is a widely-used technology in PEFT-based CIL methods \cite{l2pCVPR22, dualpromptECCV22, jung2023generating, cpromptCVPR24, codaCVPR23}.
``Maximum'' selection directly selects the sub-module with the highest task-confidence score, representing a one-hot variant of our TC-MoA. 
The results (upper $vs.$ lower in Tab.\ref{tab: ablation-tc-moa}) 
indicate that DKA gains model performance with various aggregation strategies, further demonstrating the effectiveness of our DKA.

To gain a more detailed understanding of our CA, we conduct ablation studies of \textbf{\emph{the feature simulation process}} in CA on 10S-ImageNetA to indicate its efficacy, as shown in Tab.\ref{tab: ablation-cu}.
``CA w/o subspace transfer'' denotes training the classifier using features obtained directly from Gaussian Sampling without subspace transfer.
``CA w/o affinities $\mathbf{G}$'' denotes modeling the shift of previous features without considering their affinities $\mathbf{G}$ with current features, instead directly utilizing the average shift of current features.
The results demonstrate the efficacy of each step in our affinity-guided feature simulation. 

Additionally, we examine the effectiveness of our DKA through the task-by-task accuracy curves in Fig.\ref{fig:ablation}, highlighting DKA’s superiority under long-term settings.

\begin{table}[H]
\centering
\caption{{Comparison of DKA with traditional alignment approaches.} }\label{tab: alignment_ablation}
\begin{adjustbox}{max width=0.49\textwidth}
\begin{tabular}{l|cc|cc|c}
\hline
\multirow{2}{*}{Methods} & \multicolumn{2}{c|}{10S-ImageNetA} & \multicolumn{2}{c|}{10S-ImageNetR} & \multirow{2}{*}{Gain}\\
\cline{2-5}
& Last-acc $\uparrow$ &Avg-acc $\uparrow$
& Last-acc $\uparrow$ &Avg-acc $\uparrow$ \\
\hline
\rowcolor{lightgray} w/o Alignment & 54.33\tsb{\(\pm\)0.87} & 64.43\tsb{\(\pm\)1.22} & 79.14\tsb{\(\pm\)0.21} & 85.04\tsb{\(\pm\)0.33} & -- \\ 
+ $L_2$-reg & 50.82\tsb{\(\pm\)0.69} & 61.26\tsb{\(\pm\)1.32} & 79.16\tsb{\(\pm\)0.40} & 84.68\tsb{\(\pm\)0.28} & \textcolor{red}{-1.76} \\
+ KL-reg & 45.12\tsb{\(\pm\)1.12} & 56.33\tsb{\(\pm\)1.65} & 74.91\tsb{\(\pm\)0.53} & 81.99\tsb{\(\pm\)0.38} &  \textcolor{red}{-6.15}\\
+ Proto-reg & 53.66\tsb{\(\pm\)0.83} & 63.22\tsb{\(\pm\)1.35} & 77.76\tsb{\(\pm\)0.15} & 84.05\tsb{\(\pm\)0.52} &  \textcolor{red}{-1.06} \\
\rowcolor{lightblue} + FA (Ours)  & 58.13\tsb{\(\pm\)0.20} & 66.59\tsb{\(\pm\)0.84} & 79.74\tsb{\(\pm\)0.34} & 85.29\tsb{\(\pm\)0.19} & \textcolor{green!50!black}{+1.70} \\
\rowcolor{lightblue} + CA (Ours) & 61.29\tsb{\(\pm\)0.46} & 69.31\tsb{\(\pm\)1.16} & 81.80\tsb{\(\pm\)0.38} & 86.59\tsb{\(\pm\)0.18} & \textcolor{green!50!black}{+4.01}\\
\rowcolor{lightblue} + DKA (Ours) & \textbf{63.53}\tsb{\(\pm\)0.66} &  \textbf{71.11}\tsb{\(\pm\)1.03} & \textbf{82.41}\tsb{\(\pm\)0.13} & \textbf{87.20}\tsb{\(\pm\)0.35} & \textcolor{green!50!black}{\textbf{+5.33}} \\
\hline
\end{tabular}
\end{adjustbox}
\end{table}

{
\textbf{Comparison of DKA (FA+CA) with traditional alignment approaches.} We compare DKA with three traditional alignment objectives: (1) ``$L_2$-reg" minimizes the sample-wise feature distance between the current and historical subspaces; (2) ``KL-reg" minimizes the KL-divergence between their prediction distributions; (3) ``Proto-reg" matches current class-wise mean features maintained by Exponential Moving Average (EMA) with historical prototypes. The results in Tab.\ref{tab:  alignment_ablation} show that these approaches even underperform the model without alignment, whereas DKA achieves the best results. Since each task-specific module primarily captures distinctive task knowledge, such absolute consistency regularization suppresses the task-specific variations essential for learning new tasks. In contrast, DKA promotes \emph{\textbf{compatible discriminative subspaces}} by structuring relative inter-class relationships across subspaces (FA) and adapting classifier boundaries to simulated cross-subspace variations (CA), without enforcing identical representations and preserving model plasticity.
}

\textbf{The effectiveness of TC-MoA.} TC-MoA achieves an improvement of 0.50\%$\sim$1.19\% Last-acc (Idx 6 $v.s.$ Idx 7) compared to directly utilizing the task-shared network for classification.
This observation indicates that our TC-MoA is a powerful tool for aggregating task-specific knowledge, while the task-shared network alone is insufficient for the final decision.
We further compare \textbf{\emph{different aggregation strategies}} in Tab.\ref{tab: ablation-tc-moa}, where existing ``Query-Key'' matching exhibits limited performance as a single task key is insufficient to represent the data distribution within an entire task.
In contrast, our TC-MoA softly aggregates task-specific knowledge based on task-confidence scores, avoiding overconfidence in misleading task-ids and fully leveraging the fine-grained instance-class relationships.

\begin{table}[h]
\centering
\caption{Ablation study of the dual classifier design.}\label{tab: r2_classifier_ablation}
\begin{adjustbox}{max width=0.48\textwidth}
\begin{tabular}{l|cc|cc}
\hline
\multirow{2}{*}{Methods} & \multicolumn{2}{c|}{10S-ImageNetA} & \multicolumn{2}{c}{10S-ImageNetR}\\
\cline{2-5}
& Last-acc $\uparrow$ &Avg-acc $\uparrow$
& Last-acc $\uparrow$ &Avg-acc $\uparrow$ \\
\hline
Only $g$ & 
58.13\tsb{\(\pm\)0.20} & 
66.59\tsb{\(\pm\)0.84} &
79.74\tsb{\(\pm\)0.34} & 
85.29\tsb{\(\pm\)0.19} \\
Only $g^u$ & 
63.00\tsb{\(\pm\)0.13} & 
70.75\tsb{\(\pm\)0.90} &
82.10\tsb{\(\pm\)0.16} &
87.03\tsb{\(\pm\)0.08}
\\
$g^u$ + $g$ ($w=0.50$) & 63.13\tsb{\(\pm\)0.78} & 70.94\tsb{\(\pm\)1.16} & 82.21\tsb{\(\pm\)0.24} & 87.15\tsb{\(\pm\)0.17} \\
$g^u$ + $g$ ($w=0.80$) & 63.25\tsb{\(\pm\)0.57} & 70.86\tsb{\(\pm\)1.23} & 82.18\tsb{\(\pm\)0.40} & 87.08\tsb{\(\pm\)0.12} \\
$g^u$ + $g$ ($w=\frac{t_c}{t_c+1}$) & \textbf{63.53}\tsb{\(\pm\)0.66} & \textbf{71.11}\tsb{\(\pm\)1.03} & \textbf{82.41}\tsb{\(\pm\)0.13} & \textbf{87.20}\tsb{\(\pm\)0.35} \\
\hline
\end{tabular}
\end{adjustbox}
\end{table}

\textbf{Ablation studies of the global classifier $g^u$ and the standard classifier $g$.}
The two classifiers provide complementary decision capabilities: (1) $g$ learns stable category discrimination from shared and correct task-specific features, and (2) $g^u$ aggregates task-adaptive classifiers trained with real and simulated incorrect-subspace features to improve robustness under misleading task-IDs. As shown in Tab.~\ref{tab: r2_classifier_ablation}, their combination achieves the best performance, confirming their complementary roles. We also report performances under different fusion weights $w$ in prediction, $i.e.$, $w\cdot g^u(\hat{\mathbf{z}}_i)[k]+(1-w)\cdot g(\hat{\mathbf{z}}_i)[k]$. The dynamic weight $w=\frac{t_c}{t_c+1}$ outperforms fixed weights by gradually increasing the contribution of $g^u$ as more cross-subspace knowledge is accumulated.

\begin{table*}[!t]
\centering
\caption{Task-by-task accuracy of the \emph{\textbf{challenging samples}} on 10S-ImageNetA.}\label{tab: incorrect-acc-imageneta}
\vspace{-2mm}
\begin{adjustbox}{max width=1.0\textwidth}
\begin{tabular}{l|ccccccccc}
\hline
Methods & Ses.2 & Ses.3 & Ses.4 & Ses.5 & Ses.6 & Ses.7 & Ses.8 & Ses.9 & Ses.10 \\
\hline
\multicolumn{9}{l}{\textit{CKAA with frozen pre-trained task-shared network.}}\\
\hline
pre-trained backbone & 0.00\tsb{\(\pm\)0.00}	& 0.00\tsb{\(\pm\)0.00}& 0.00\tsb{\(\pm\)0.00}&	0.00\tsb{\(\pm\)0.00}&	0.00\tsb{\(\pm\)0.00}	&0.00\tsb{\(\pm\)0.00}	&0.00\tsb{\(\pm\)0.00}	&0.00\tsb{\(\pm\)0.00}	&0.00\tsb{\(\pm\)0.00}\\
+ TC-MoA 
&43.26\tsb{\(\pm\)5.96}	
&44.84\tsb{\(\pm\)5.08} 
&43.67\tsb{\(\pm\)1.45} 
&39.47\tsb{\(\pm\)2.20}	
&36.09\tsb{\(\pm\)2.26}	
&35.01\tsb{\(\pm\)2.87}
&32.29\tsb{\(\pm\)1.01}
&31.00\tsb{\(\pm\)1.75}	
&30.20\tsb{\(\pm\)1.46}\\
+ TC-MoA + CA 
&52.20\tsb{\(\pm\)2.80}	
&51.74\tsb{\(\pm\)4.33}	
&48.49\tsb{\(\pm\)1.28}	
&46.61\tsb{\(\pm\)1.56}	
&43.99\tsb{\(\pm\)1.64}	
&41.93\tsb{\(\pm\)2.53}	
&40.51\tsb{\(\pm\)0.51}	
&39.19\tsb{\(\pm\)0.41}	
&37.88\tsb{\(\pm\)0.47}\\
+ TC-MoA + FA 
&43.72\tsb{\(\pm\)7.96}	
&48.07\tsb{\(\pm\)7.06}	
&47.08\tsb{\(\pm\)2.65}	
&43.17\tsb{\(\pm\)2.16}	
&41.81\tsb{\(\pm\)2.52}	
&37.80\tsb{\(\pm\)3.44}	
&36.68\tsb{\(\pm\)3.04}	
&34.73\tsb{\(\pm\)1.53}	
&33.76\tsb{\(\pm\)1.24}\\
+ TC-MoA + DKA 
&\textbf{53.99}\tsb{\(\pm\)6.41}
&\textbf{53.06}\tsb{\(\pm\)2.95}	
&\textbf{52.60}\tsb{\(\pm\)1.86}	
&\textbf{49.65}\tsb{\(\pm\)0.27}	
&\textbf{47.33}\tsb{\(\pm\)2.52}	
&\textbf{45.11}\tsb{\(\pm\)2.82}	
&\textbf{44.54}\tsb{\(\pm\)0.87}	
&\textbf{42.87}\tsb{\(\pm\)0.89}	
&\textbf{42.01}\tsb{\(\pm\)0.66} \\
\hline
\multicolumn{9}{l}{\textit{CKAA with learnable visual prompt-based task-shared network.}}\\
\hline
$\mathcal{P}_{sh}$ & 0.00\tsb{\(\pm\)0.00}	& 0.00\tsb{\(\pm\)0.00}& 0.00\tsb{\(\pm\)0.00}&	0.00\tsb{\(\pm\)0.00}&	0.00\tsb{\(\pm\)0.00}	&0.00\tsb{\(\pm\)0.00}	&0.00\tsb{\(\pm\)0.00}	&0.00\tsb{\(\pm\)0.00}	&0.00\tsb{\(\pm\)0.00}\\
+ TC-MoA & 23.90\tsb{\(\pm\)5.54}	& 19.97\tsb{\(\pm\)1.67}& 17.62\tsb{\(\pm\)1.23}&	14.82\tsb{\(\pm\)2.42}&	13.29\tsb{\(\pm\)1.13}	&12.31\tsb{\(\pm\)0.33}	&12.48\tsb{\(\pm\)1.05}	&11.41\tsb{\(\pm\)0.5}	&10.48\tsb{\(\pm\)0.22} \\
+ TC-MoA + CA & 30.48\tsb{\(\pm\)5.96} &	29.59\tsb{\(\pm\)0.88} &	30.80\tsb{\(\pm\)6.53} &	28.56\tsb{\(\pm\)5.17} &	27.54\tsb{\(\pm\)2.62} &	27.36\tsb{\(\pm\)3.25} & 25.97\tsb{\(\pm\)1.60} &	25.50\tsb{\(\pm\)1.08} &	25.68\tsb{\(\pm\)0.39} \\
+ TC-MoA + FA & 20.96\tsb{\(\pm\)4.42} &	21.41\tsb{\(\pm\)4.48} & 21.95\tsb{\(\pm\)3.96} &	19.60\tsb{\(\pm\)2.12} &	17.94\tsb{\(\pm\)1.73} &	17.59\tsb{\(\pm\)3.21} &	17.25\tsb{\(\pm\)2.05} & 16.81\tsb{\(\pm\)1.57} & 16.43\tsb{\(\pm\)1.52}\\
+ TC-MoA + DKA & \textbf{35.76}\tsb{\(\pm\)0.93} & \textbf{33.75}\tsb{\(\pm\)8.83} & \textbf{36.02}\tsb{\(\pm\)2.03} & \textbf{31.84}\tsb{\(\pm\)2.85} & \textbf{31.28}\tsb{\(\pm\)2.78} & \textbf{29.91}\tsb{\(\pm\)2.85} & \textbf{29.25}\tsb{\(\pm\)1.42} & \textbf{28.31}\tsb{\(\pm\)1.97} & \textbf{27.71}\tsb{\(\pm\)0.53} \\
\hline
\end{tabular}
\end{adjustbox}
\vspace{-3mm}
\end{table*}

\begin{table*}[!t]
\centering
\caption{Task-by-task accuracy of the \emph{\textbf{easy samples}} on 10S-ImageNetA.}\label{tab: correct-acc-imageneta}
\vspace{-2mm}
\begin{adjustbox}{max width=1.0\textwidth}
\begin{tabular}{l|ccccccccc}
\hline
Methods & Ses.2 & Ses.3 & Ses.4 & Ses.5 & Ses.6 & Ses.7 & Ses.8 & Ses.9 & Ses.10 \\
\hline
\multicolumn{9}{l}{\textit{CKAA with frozen pre-trained task-shared network.}}\\
\hline
pre-trained backbone 
&67.05\tsb{\(\pm\)3.05}
&67.15\tsb{\(\pm\)0.96}
&69.63\tsb{\(\pm\)1.68}	
&71.60\tsb{\(\pm\)2.15}	
&72.61\tsb{\(\pm\)3.42}	
&73.43\tsb{\(\pm\)2.46}	
&74.12\tsb{\(\pm\)1.52}	
&72.56\tsb{\(\pm\)1.93}	
&72.57\tsb{\(\pm\)1.84} \\
+ TC-MoA 
&79.50\tsb{\(\pm\)3.98}
&82.11\tsb{\(\pm\)3.68}
&82.65\tsb{\(\pm\)1.49}
&83.54\tsb{\(\pm\)1.81}
&82.73\tsb{\(\pm\)0.99}
&82.48\tsb{\(\pm\)0.69}
&82.91\tsb{\(\pm\)2.30}
&82.53\tsb{\(\pm\)1.06}	
&82.41\tsb{\(\pm\)1.36} \\
+ TC-MoA + CA 
&81.92\tsb{\(\pm\)2.65}
&85.46\tsb{\(\pm\)1.61}
&86.08\tsb{\(\pm\)1.03}	
&87.17\tsb{\(\pm\)1.66}	
&87.25\tsb{\(\pm\)1.13}	
&86.24\tsb{\(\pm\)0.63}	
&86.09\tsb{\(\pm\)1.34}	
&85.92\tsb{\(\pm\)1.70}	
&86.36\tsb{\(\pm\)1.68} \\
+ TC-MoA + FA 
&81.07\tsb{\(\pm\)3.75}
&82.75\tsb{\(\pm\)3.30}	
&82.68\tsb{\(\pm\)1.20}
&84.89\tsb{\(\pm\)0.40}
&83.85\tsb{\(\pm\)1.63}	
&84.49\tsb{\(\pm\)1.52}	
&84.97\tsb{\(\pm\)0.86}	
&84.76\tsb{\(\pm\)2.38}	
&84.35\tsb{\(\pm\)2.39} \\
+ TC-MoA + DKA 
&\textbf{83.12}\tsb{\(\pm\)1.31}
&\textbf{86.40}\tsb{\(\pm\)1.86}	
&\textbf{86.19}\tsb{\(\pm\)0.99}	
&\textbf{88.81}\tsb{\(\pm\)1.34}	
&\textbf{87.79}\tsb{\(\pm\)1.34}	
&\textbf{86.78}\tsb{\(\pm\)0.75}	
&\textbf{86.89}\tsb{\(\pm\)2.12}	
&\textbf{86.87}\tsb{\(\pm\)2.89}	
&\textbf{86.83}\tsb{\(\pm\)3.30} \\
\hline
\multicolumn{9}{l}{\textit{CKAA with learnable visual prompt-based task-shared network.}}\\
\hline
$\mathcal{P}_{sh}$ 
&81.92\tsb{\(\pm\)2.49}	
&83.42\tsb{\(\pm\)2.79}	
&85.05\tsb{\(\pm\)3.00}	
&85.80\tsb{\(\pm\)2.99}	
&85.86\tsb{\(\pm\)2.48}	
&85.14\tsb{\(\pm\)1.72}	
&85.79\tsb{\(\pm\)1.49}	
&85.45\tsb{\(\pm\)1.79}	
&85.98\tsb{\(\pm\)1.17} \\
+ TC-MoA 
&79.01\tsb{\(\pm\)3.99}
&80.84\tsb{\(\pm\)2.24}	
&83.49\tsb{\(\pm\)4.47}	
&84.31\tsb{\(\pm\)1.51} 
&84.60\tsb{\(\pm\)1.27}	
&83.62\tsb{\(\pm\)0.48}
&83.07\tsb{\(\pm\)0.64}	
&82.33\tsb{\(\pm\)1.43}	
&82.85\tsb{\(\pm\)1.97} \\
+ TC-MoA + CA 
&80.10\tsb{\(\pm\)9.33}
&84.19\tsb{\(\pm\)3.20}	
&86.26\tsb{\(\pm\)3.78}	
&85.54\tsb{\(\pm\)1.66}	
&86.09\tsb{\(\pm\)1.42}	
&86.64\tsb{\(\pm\)3.07}	
&86.84\tsb{\(\pm\)0.39}	
&85.71\tsb{\(\pm\)0.49}
&85.76\tsb{\(\pm\)0.50}\\
+ TC-MoA + FA 
&80.22\tsb{\(\pm\)1.81}
&82.35\tsb{\(\pm\)2.26}	
&84.26\tsb{\(\pm\)0.55}	
&85.22\tsb{\(\pm\)1.74}	
&85.30\tsb{\(\pm\)1.09}
&85.02\tsb{\(\pm\)2.25}	
&83.91\tsb{\(\pm\)1.05}	
&84.34\tsb{\(\pm\)1.93}	
&83.49\tsb{\(\pm\)2.56} \\
+ TC-MoA + DKA 
&\textbf{82.81}\tsb{\(\pm\)4.11} 	
&\textbf{85.13}\tsb{\(\pm\)1.71} 	
&\textbf{86.95}\tsb{\(\pm\)2.04} 	
&\textbf{87.76}\tsb{\(\pm\)2.66} 	
&\textbf{88.07}\tsb{\(\pm\)1.11} 	
&\textbf{87.39}\tsb{\(\pm\)1.38} 	
&\textbf{87.23}\tsb{\(\pm\)0.85} 	
&\textbf{86.08}\tsb{\(\pm\)0.81} 	
&\textbf{86.51}\tsb{\(\pm\)0.52} \\
\hline
\end{tabular}
\end{adjustbox}
\vspace{-2mm}
\end{table*}

\subsection{Analysis of the Effectiveness of CKAA in handling Misleading Task-ids}

We conduct ablation studies on 10S-ImageNetA to analyze the effectiveness of our proposed CKAA in enhancing the model's robustness against misleading task-ids.
Specifically, we divide the testing samples into two subsets: (1) \emph{\textbf{Challenging samples}}: These are misclassified by the task recognizer and assigned with incorrect task-ids; (2) \emph{\textbf{Easy samples}}: These are classified correct task-ids. We report the task-by-task accuracy of these two types of samples, respectively.
The results are shown in Tab.\ref{tab: incorrect-acc-imageneta} and Tab.\ref{tab: correct-acc-imageneta}.

As shown in Tab.\ref{tab: incorrect-acc-imageneta}, the proposed TC-MoA and DKA effectively improve the prediction accuracy for the challenging samples, delivering more reliable predictions for features under incorrect task-specific sub-modules.
Notably, the accuracy in the half bottom of Tab.\ref{tab: incorrect-acc-imageneta} is lower compared to the top half part.
It is mainly due to our learnable task-shared network (the half bottom) encapsulates sufficient downstream information, the samples it fails to distinguish are inherently more challenging.
Overall, our CKAA achieves significant improvement in the classification accuracy of these challenging samples.

\begin{table}[h]
\centering
\caption{Ablation studies under different task-ID error rate groups on 10S-ImageNetA.}
\label{tab:ablation_error_rate_ina}
\resizebox{\columnwidth}{!}{
\begin{tabular}{c|ccc|ccccc}
\toprule
\multirow{2}{*}{Idx} &
\multirow{2}{*}{TC-MoA} &
\multirow{2}{*}{FA} &
\multirow{2}{*}{CA} &
\multicolumn{5}{c}{Task-ID Error Rate (\%)} \\
\cline{5-9}
& & & & 0--20 & 20--40 & 40--60 & 60--80 & 80--100 \\
\midrule
1 & --         & --         & --         & \textbf{83.87} & 66.96 & 55.18 & 32.18 & 19.25 \\
2 & \checkmark & --         & --         & 79.57 & 68.76 & 55.70 & 40.65 & 19.91 \\
3 & \checkmark & --         & \checkmark & 79.05 & 74.51 & 64.22 & 49.80 & 33.64 \\
4 & \checkmark & \checkmark & --         & 79.72 & 70.99 & 61.72 & 50.63 & 23.11 \\
5 & \checkmark & \checkmark & \checkmark & 79.05 & \textbf{76.02} & \textbf{64.85} & \textbf{58.74} & \textbf{34.50} \\
\bottomrule
\end{tabular}
}
\vspace{-4mm}
\end{table}

\begin{table}[h]
\centering
\caption{Ablation studies under different task-ID error rate groups on 10S-CUB.}
\label{tab:ablation_error_rate_cub}
\resizebox{\columnwidth}{!}{
\begin{tabular}{c|ccc|ccccc}
\toprule
\multirow{2}{*}{Idx} &
\multirow{2}{*}{TC-MoA} &
\multirow{2}{*}{FA} &
\multirow{2}{*}{CA} &
\multicolumn{5}{c}{Task-ID Error Rate (\%)} \\
\cline{5-9}
& & & & 0--20 & 20--40 & 40--60 & 60--80 & 80--100 \\
\midrule
1 & -- & --  & --  & 94.04 & 71.78 & 59.64 & 47.73 & 17.69 \\
2 & \checkmark & --   & --  & 93.54 & 73.18 & 60.30 & 47.97 & 18.82 \\
3 & \checkmark & -- & \checkmark & 94.57 & 78.85 & 69.75 & 54.43 & 45.77 \\
4 & \checkmark & \checkmark & -- & 92.92 & 75.62 & 65.40 & 51.12 & 33.95 \\
5 & \checkmark & \checkmark & \checkmark & \textbf{95.37} & \textbf{82.61} & \textbf{72.43} & \textbf{57.67} & \textbf{56.98} \\
\bottomrule
\end{tabular}
}
\end{table}

We further evaluate CKAA across different task-ID error-rate groups. For each class $c$ from task $t^c$, we define its class-wise task-ID error rate as
$\mathcal{E}_{\mathrm{TID}}(c)=\frac{1}{|\mathcal{D}_c|}\sum_{\mathbf{x}^i\in\mathcal{D}_c}\mathbb{I}[\arg\max_t\alpha_t^i\neq t^c]\times100\%$, where $\mathcal{D}_c$ is the test set of class $c$ and $\alpha_t^i$ is the task-confidence score of sample $i$ for task $t$. 
We divide all classes into five groups according to $\mathcal{E}_{\mathrm{TID}}(c)$, i.e., $0$--$20\%$, $20$--$40\%$, $40$--$60\%$, $60$--$80\%$, and $80$--$100\%$, and report the accuracy of samples belonging to each group in Tabs.~\ref{tab:ablation_error_rate_ina} and~\ref{tab:ablation_error_rate_cub}. The baseline performance drops substantially as the task-ID error rate increases, confirming that misleading task-IDs severely affect prediction reliability. 
In contrast, CKAA consistently improves the accuracy of high-error classes: On 10S-ImageNetA, the accuracy increases from $19.25\%$ to $34.50\%$ in the $80$--$100\%$ group, and On 10S-CUB, the accuracy in the $80$--$100\%$ group increases from $17.69\%$ to $56.98\%$. 
These results indicate that CKAA is particularly effective for samples that are likely to be routed to incorrect task-specific subspaces.

\begin{figure}[!t]
\centering
\includegraphics[width=0.49\textwidth]{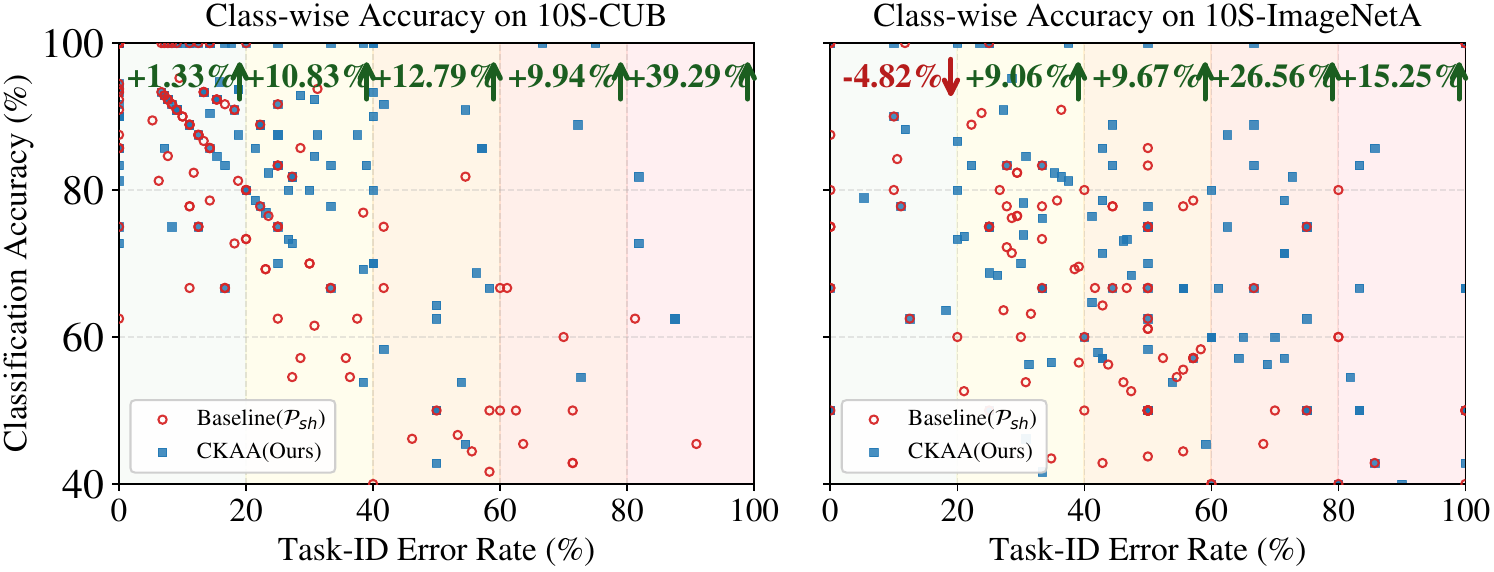}
\caption{Class-wise accuracy versus task-ID error rate on 10S-ImageNetA and 10S-CUB. }
\label{fig: per-class-error}
\end{figure}

Fig.\ref{fig: per-class-error} further visualizes the class-wise relationship between task-ID errors and classification performance on 10S-ImageNetA and 10S-CUB. The visualization compares CKAA and the shared-prompt-based baseline ($\mathcal{P}_{sh}$). The horizontal axis represents class-wise task-ID error rate $\mathcal{E}_{\mathrm{TID}}(c)$, while the vertical axis represents the classification accuracy of that class. These visualizations further confirm that CKAA effectively improves robustness for classes affected by incorrect task-ID estimation.

\begin{figure*}[t!]
\centering
\includegraphics[width=1.0\textwidth]{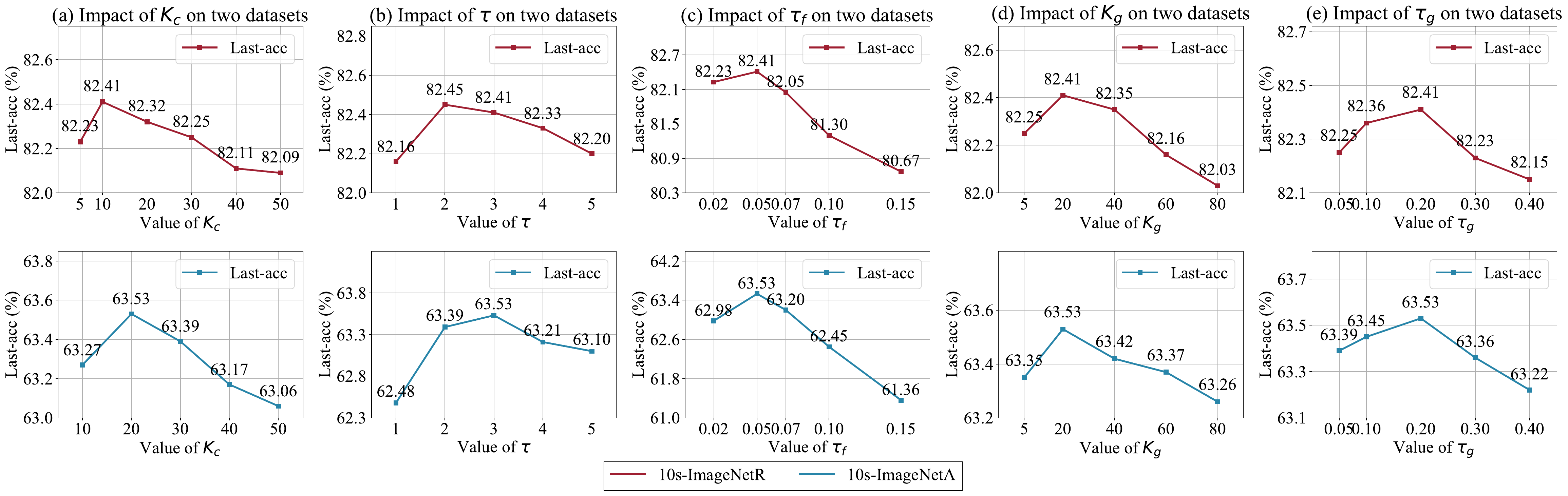}
\vspace{0mm}
\caption{Parameter analysis of $K_c$ and $\tau$ in TC-MoA, and $\tau_f$, $K_g$ and $\tau_g$ in DKA on 10S-ImageNetR and 10S-ImageNetA.
}\label{fig: parameter-analysis}
\end{figure*}

\subsection{Hyper-Parameter Analysis}

\textbf{Key hyper-parameters.} We conduct experiments to analyze the impact of the hyper-parameters in our propsoed method on 10S-ImageNetR and 10S-ImageNetA benchmarks, including $\tau_f$ in Eq.\textcolor{iccvblue}{4}, $K_g$ and $\tau_g$ in Eq.\textcolor{iccvblue}{5}, $K_c$ in Eq.\textcolor{iccvblue}{10} and $\tau$ in Eq.\textcolor{iccvblue}{11}. We first analyze the hyper-parameters in TC-MoA, including $K_c$ and $\tau$. The results are shown in Fig.\ref{fig: parameter-analysis}(a) and Fig.\ref{fig: parameter-analysis}(b). $K_c$ is designed to select the Top-$K_c$ classes with the highest logits to compute the task-confidence scores. A too small $K_c$ makes task-confidence estimation vulnerable to a few spuriously activated classes, whereas a too large $K_c$ includes responses from irrelevant tasks and weakens the dominance of the correct adapter. A small $\tau$ makes the aggregation weights overly sharp, while an excessively large $\tau$ over-smooths the weights and introduces interference from irrelevant adapters. We set $K_c=10$ for ImageNetR and $K_c=20$ for ImageNetA, $\tau=2.0$ for all datasets based on experimental results. Then we analyze the hyper-parameters in DKA, including $\tau_f$, $K_g$, and $\tau_g$.
The results are shown in Fig.\ref{fig: parameter-analysis}(c), Fig.\ref{fig: parameter-analysis}(d), and Fig.\ref{fig: parameter-analysis}(e).
$\tau_f$ is a temperature factor of the Cross Subspace Feature Alignment (CSFA) loss in Eq.\ref{eq: fu-cscl-loss}.
$K_g$ and $\tau_g$ are two parameters for modeling the affinities between current and previous data.
Based on the experimental results, we set $\tau_f=0.05$, $K_g=20$ and $\tau_g=0.2$ for all datasets.

\begin{figure}[!t]
\centering
\begin{subfigure}{\columnwidth}
    \centering
    \includegraphics[width=\columnwidth]{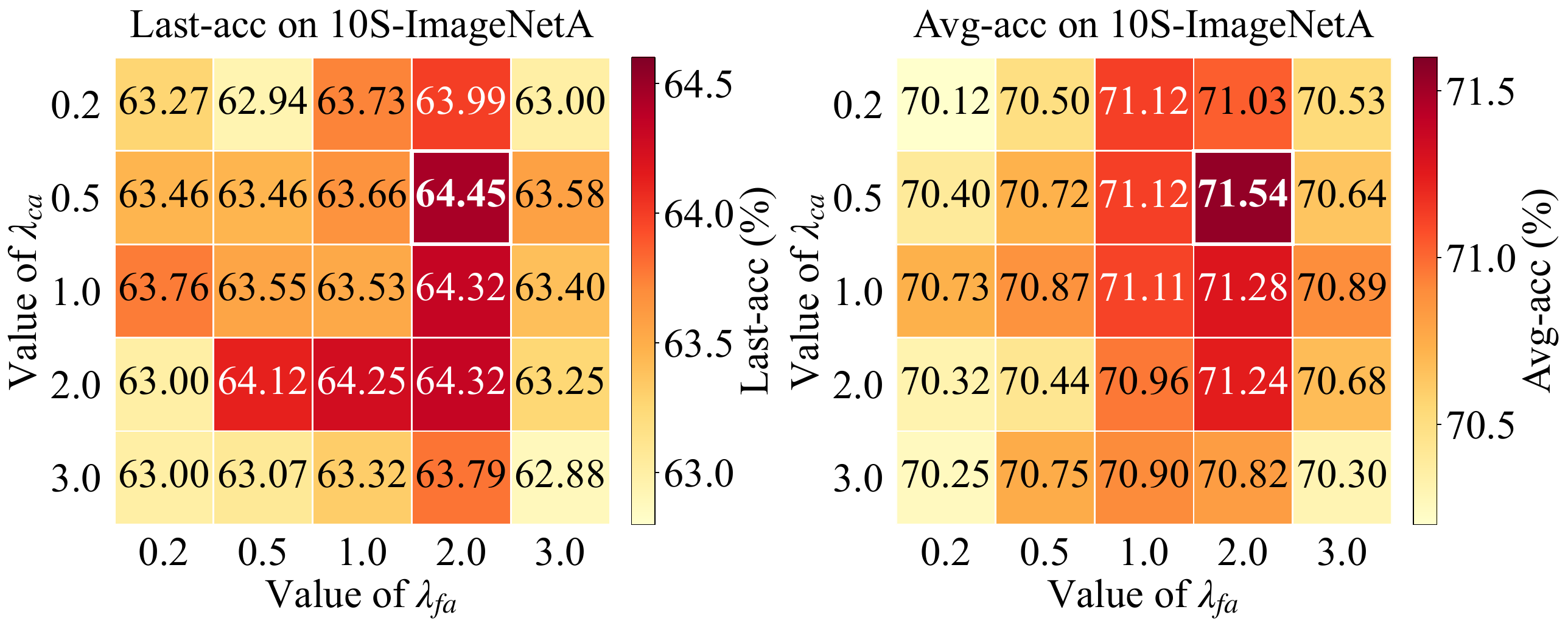}
    \label{fig: loss-term-ina}
\end{subfigure}
\vspace{2mm}
\begin{subfigure}{\columnwidth}
    \centering
    \includegraphics[width=\columnwidth]{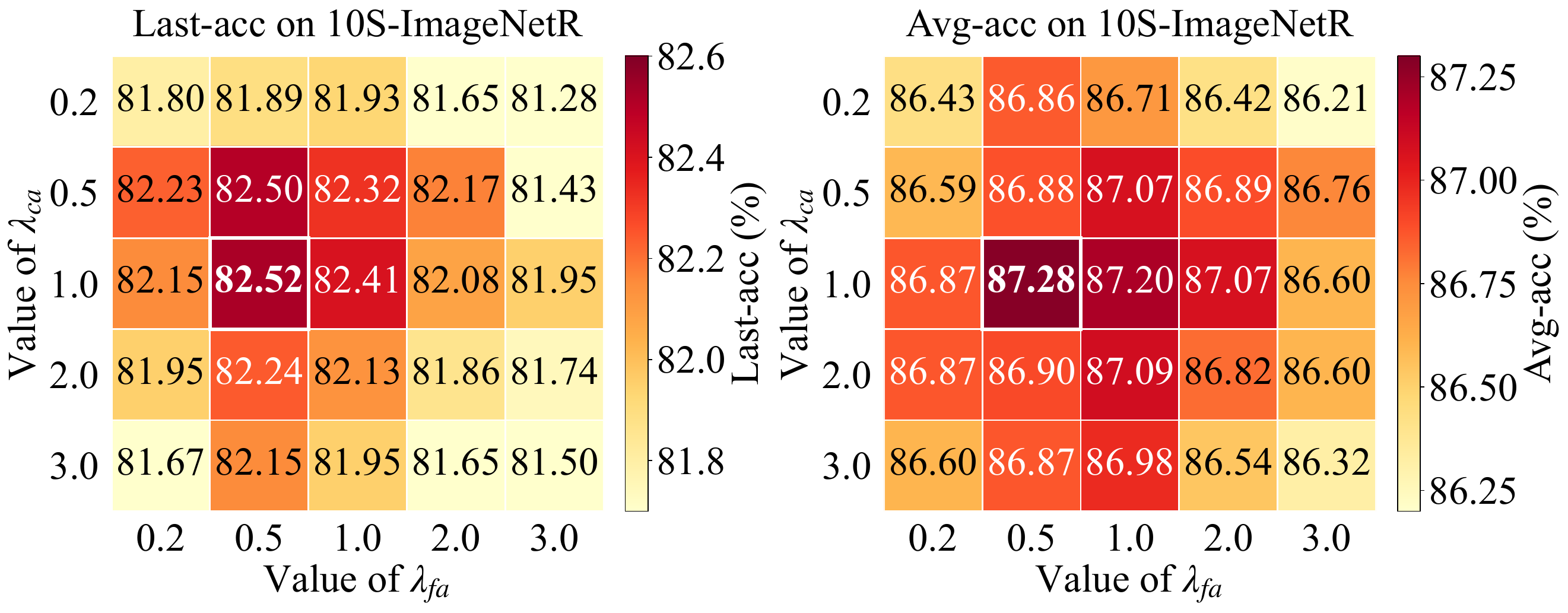}
    \label{fig: loss-term-inr}
\end{subfigure}
\caption{Model performances under varying values of $\lambda_{ca}$ and $\lambda_{fa}$ on 10S-ImageNetA and 10S-ImageNetR.}
\label{fig: loss-term-sensitivity}
\end{figure}

\textbf{The loss weights $\lambda_{fa}$ and $\lambda_{ca}$ in Eq.~\ref{eq: total_loss}.}We vary $\lambda_{fa},\lambda_{ca}\in\{0.2,0.5,1.0,2.0,3.0\}$ while fixing the coefficient of $\mathcal{L}_{ce}$ as $1.0$. As shown in Fig.~\ref{fig: loss-term-sensitivity}, CKAA remains stable across a broad range of weight combinations: the default setting $(\lambda_{fa},\lambda_{ca})=(1.0,1.0)$ achieves results close to the best ones on both 10S-ImageNetA and 10S-ImageNetR. Since the optimal trade-off slightly varies across datasets, we adopt $\lambda_{fa}=\lambda_{ca}=1.0$ by default to avoid dataset-specific tuning.

\begin{figure}[h]
\centering
\includegraphics[width=\columnwidth]{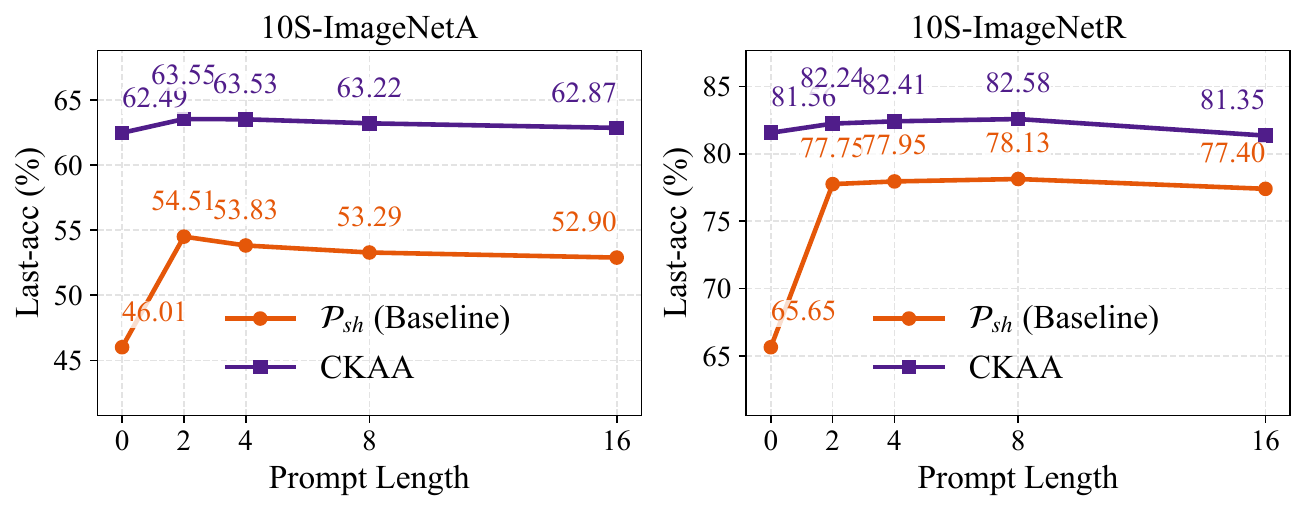}
\caption{Performances under different prompt lengths.}
\label{fig: prompt_length}
\end{figure}


\begin{table}[t]
\centering
\caption{Performances under different adapter widths.}
\label{tab:varying_width}
\resizebox{\columnwidth}{!}{
\begin{tabular}{c|c|c|cc}
\hline
Width & Params & FLOPs 
& 10S-ImageNetA & 10S-ImageNetR \\
\hline
2  & 0.42M (0.48\%)  & 17.09 GMac 
& 63.40\tsb{\(\pm\)0.66} & 81.63\tsb{\(\pm\)0.70} \\

4  & 0.83M (0.96\%)  & 17.16 GMac 
& 63.84\tsb{\(\pm\)0.47} & 82.04\tsb{\(\pm\)0.20} \\

8  & 1.66M (1.93\%)  & 17.31 GMac 
& \textbf{64.03}\tsb{\(\pm\)1.06} & 81.77\tsb{\(\pm\)0.75} \\

16 & 3.32M (3.86\%)  & 17.60 GMac 
& 63.56\tsb{\(\pm\)0.32} & 82.06\tsb{\(\pm\)1.09} \\

32 & 6.64M (7.72\%)  & 18.18 GMac 
& 63.27\tsb{\(\pm\)0.74} & 82.15\tsb{\(\pm\)0.47} \\

64 & 13.28M (15.44\%) & 19.34 GMac 
& 63.53\tsb{\(\pm\)0.66} & \textbf{82.41}\tsb{\(\pm\)0.13} \\
\hline
\end{tabular}
}
\end{table}

\textbf{Prompt length and adapter width.}We added a sensitivity analysis by varying the visual shared prompt length $N_p$ over $\{0, 2, 4, 8, 16\}$ on 10S-ImageNetA and 10S-ImageNetR. The results are shown in Fig.\ref{fig: prompt_length}. The results demonstrate the effectiveness of the shared prompt for task-ID recognition. Compared with the setting without the shared prompt, $i.e.$, $N_p=0$, introducing a short shared prompt with $N_p=2$ clearly improves the performance of our CKAA. The results also indicate that excessively long prompts may introduce redundant task-specific information and lead to overfitting. Based on the results, we set $N_p=4$. We also conducted experiments to analyze the impact of adapter widths on 10S-ImageNetA and 10S-ImageNetR datasets, the results are shown in Tab.\ref{tab: varying_width}. Our method performs well even with a width of 2, achieving a favorable trade-off between efficiency and performances, referred to as `CKAA-'.

\begin{table}[h]
\centering
\caption{Sensitivity analysis of the sample size in DKA on 10S-ImageNetA.}
\label{tab: r2_gaussian_sample_size}
\begin{tabular}{c|c|cc}
\toprule
Sampling strategy & Sample size & Last-acc $\uparrow$ & Avg-acc $\uparrow$ \\
\midrule
\multirow{4}{*}{Class-balanced}
& $1$ per class  & $63.16 \pm 0.47$ & $70.57 \pm 1.10$ \\
& $2$ per class  & $63.16 \pm 0.27$ & $70.83 \pm 1.16$ \\
& $5$ per class  & $\mathbf{63.53 \pm 0.66}$ & $71.11 \pm 1.03$ \\
& $10$ per class & $63.39 \pm 0.24$ & $\mathbf{71.12 \pm 0.98}$ \\
\midrule
\multirow{2}{*}{Fixed-budget}
& $128$ in total & $62.88 \pm 0.76$ & $70.32 \pm 1.08$ \\
& $256$ in total & $63.26 \pm 0.53$ & $70.95 \pm 1.23$ \\
\bottomrule
\end{tabular}
\end{table}

\textbf{The number of sampled pseudo features $B_s$.} We consider two sampling strategies: (1) the class-balanced strategy, where ``$M$ per class'' means that $M$ Gaussian features are generated from the estimated distribution of every class, and (2) the fixed-budget strategy, where the total number of sampled features is limited to a predefined budget, i.e., $128$ or $256$. The results are in Tab.\ref{tab: r2_gaussian_sample_size}. The performance remains relatively stable across different sample sizes and strategies, while the class-balanced strategy achieves slightly better performances. Based on the experimental results, we adopt $5$ sampled features per class in all experiments. The additional cost of feature simulation mainly comes from affinity construction and subspace-shift transfer, resulting in $\Delta C_{\rm sim}^{\rm train}=\mathcal{O}(B_sd(B+K_g))$. Even with a large $B_s$ when $B_s=1000$, the cost introduced by feature simulation is $\Delta C_{\mathrm{sim}}^{\mathrm{train}}=21{,}626{,}880+3{,}932{,}160=
25{,}559{,}040
\approx
0.01$GMac per batch ($\approx 0.0009$GMac per sample). This overhead is negligible compared with the
backbone computation ($\approx 17.23$GMac per sample).

\subsection{Analysis of additional computational costs}

\begin{table}[h] \centering \caption{Analysis of the complexity of components.} \label{tab:theoretical_cost} \vspace{1mm} \resizebox{\columnwidth}{!}{ \begin{tabular}{l|c|c} \toprule Component & Training Complexity & Inference Complexity \\ 
\midrule 
Task-specific adapters & $\mathcal{O}(BLNdw)$ & $\mathcal{O}(TLNdw)$ \\ 
FA ($\mathcal{L}_{csfa}$) & $\mathcal{O}(B^2d)$ & -- \\
CA (feature simulation) & $\mathcal{O}(B_sd(B+K_g))$ & --\\ 
CA (global classifier $g^u$) & $\mathcal{O}(dK_T)$ & $\mathcal{O}(dK_T)$ \\ 
TC-MoA & -- & $\mathcal{O}(TLNd)$ \\ 
\bottomrule 
\end{tabular} } 
\vspace{-1mm} 
\end{table}

\begin{table}[h] \centering \caption{Practical additional computational costs under ViT-B/16, where $L=12$, $N=197$, $d=768$.} \label{tab:practical_cost} \vspace{1mm} \resizebox{\columnwidth}{!}{ \begin{tabular}{l|c|c|c} \toprule Setting & Extra Params & Train Cost/Sample & Inference Cost/Image \\ \midrule CKAA$^{-}$, $T=10,w=2$ & $+0.52$M & $+0.01$ GMac & $+0.091$ GMac \\ CKAA, $T=10,w=64$ & $+11.95$M & $+0.23$ GMac & $+2.342$ GMac \\ CKAA, $T=100,w=8$ & $+14.90$M & $+0.03$ GMac & $+3.086$ GMac \\ \bottomrule \end{tabular} } \vspace{-1mm} \end{table}

\begin{table*}[h]
\centering
\caption{Comparison with representative existing approaches on 10S-ImageNetR in terms of computational overhead, inference time per image, and CL performance. 
The latency is measured on a single NVIDIA RTX 4090 GPU. 
``--'' indicates that the corresponding task recognition process is not involved in the framework}
\label{tab:r2_latency_comparison_imagenetr}
\vspace{2mm}
\resizebox{0.80\textwidth}{!}{
\begin{tabular}{l|c|cc|cc|cc}
\toprule
\multirow{2}{*}{Method} &
\multirow{2}{*}{Venue} &
\multicolumn{2}{c|}{Task Recognition} &
\multicolumn{2}{c|}{Inference} &
\multicolumn{2}{c}{10S-ImageNetR} \\
\cline{3-8}
&
& Overhead
& Time/Image
& Overhead
& Time/Image
& Last-acc $\uparrow$
& Avg-acc $\uparrow$ \\
\midrule
L2P~\cite{l2pCVPR22}
& CVPR-22
& 17.0GMac & 0.515ms
& 19.9GMac & 0.658ms
& $72.34_{\pm 0.17}$
& $77.36_{\pm 0.64}$ \\
DualPrompt~\cite{dualpromptECCV22}
& ECCV-22
& 17.0GMac & 0.515ms
& 17.6GMac & 0.607ms
& $69.10_{\pm 0.62}$
& $74.28_{\pm 0.66}$ \\
CODAPrompt~\cite{codaCVPR23}
& CVPR-23
& 17.0GMac & 0.515ms
& 18.2GMac & 0.720ms
& $73.31_{\pm 0.50}$
& $78.47_{\pm 0.53}$ \\
Adam-adapter~\cite{adamIJCV2024}
& IJCV-24
& -- & --
& 17.8GMac & 0.599ms
& $65.79_{\pm 0.98}$
& $72.42_{\pm 1.41}$ \\
SSIAT~\cite{ssaitCVPR24}
& CVPR-24
& -- & --
& 17.8GMac & 0.599ms
& $79.38_{\pm 0.59}$
& $83.63_{\pm 0.43}$ \\
InfLoRA~\cite{infloraCVPR24} & CVPR-24 & -- & --
& 20.5GMac & 0.650ms 
& $75.65_{\pm 0.14}$
& $80.82_{\pm 0.24}$ \\
VQ-Prompt~\cite{vqprompt-nips24}
& NeurIPS-24
& 17.0GMac & 0.515ms
& 18.2GMac & 0.622ms
& $78.71_{\pm 0.22}$
& $83.24_{\pm 0.68}$ \\
VPT-NSP~\cite{vptnspNIPS-24}
& NeurIPS-24
& -- & --
& 17.0GMac & 0.536ms
& $77.95_{\pm 0.22}$
& $83.44_{\pm 0.40}$ \\
LoRA P\&M~\cite{qiu2026train}
& NeurIPS-25
& -- & --
& 20.5GMac & 0.650ms
& $79.95_{\pm 0.18}$
& $85.29_{\pm 0.93}$ \\
\midrule
\textbf{CKAA- (Ours)}
& --
& \textbf{17.0GMac} & \textbf{0.536ms}
& \textbf{17.1GMac} & \textbf{0.566ms}
& $\mathbf{81.63}_{\pm 0.70}$
& $\mathbf{86.93}_{\pm 0.04}$\\
\textbf{CKAA (Ours)}
& --
& \textbf{17.0GMac} & \textbf{0.536ms}
& \textbf{19.3GMac} & \textbf{0.629ms}
& $\mathbf{82.41}_{\pm 0.13}$
& $\mathbf{87.20}_{\pm 0.35}$ \\
\bottomrule
\end{tabular}
}
\end{table*}

We analyze the additional computational costs introduced by CKAA. Let $T$ denote the number of learned tasks, $L$ the number of Transformer blocks, $N$ the number of tokens, $d$ the feature dimension, $w$ the adapter width. For CA, let $B$ denote the mini-batch size, $B_s$ the number of sampled pseudo features, $K_g$ the number of nearest neighbors, and $K_T$ the total number of seen classes. As shown in Tab.~\ref{tab:theoretical_cost}, the additional training cost mainly comes from the task-specific adapter and the feature simulation in CA, while the inference cost comes mainly from the aggregation of task-specific adapters in TC-MoA. The cost of the classifier increases with $K_T$, but remains negligible compared to the ViT backbone.

Under the standard 10-task setting with $K_T=200$, our standard version CKAA with $w=64$ introduces $11.95$M extra parameters, while our efficient version CKAA- with $w=2$ only introduces $0.52$M. For computation, the additional adapter cost is $2LNdw$ per sample during training and $LTNd(2w+1)$ per image during inference with TC-MoA. The CA feature simulation cost is $\mathcal{O}(B_sd(B+K_g))$ per batch. With $B=110$, $d=768$, and $K_g=20$, even $B_s=1000$ only introduces about $0.10$GMac per batch ($0.0009$ GMac per sample), which is negligible compared to the ViT-B/16 backbone computation ($\approx 17.00$GMac).

We provide a comparison of computational overhead and inference latency with existing PEFT-based CL methods. For fair, comparisons, all latency results are measured using a single NVIDIA RTX 4090 GPU under our standard setting with a batch size of 110, and the inference time per sample is computed by averaging the total inference time over all batch samples. As shown in Tab.\ref{tab:r2_latency_comparison_imagenetr}, CKAA- exhibits an inference latency comparable to SSIAT and VPT-NSP, while achiving better performances.

\section{Conclusion}

In this paper, we propose the Cross-subspace Knowledge Alignment and Aggregation (CKAA) framework to resolve feature subspace misalignment and improve model robustness against misleading task-ids.
During training, CKAA introduces Dual-level Knowledge Alignment (DKA), which aligns feature semantics and unifies decision boundaries across task-specific subspaces, addressing the misalignment issue caused by independent sub-module training.
During inference, CKAA designs Task-Confidence-guided Mixture of Adapters (TC-MoA), a dynamic aggregation scheme that softly combines task-specific knowledge based on confidence scores, mitigating overconfidence in erroneous task-id predictions.

\noindent\textbf{Limitations.} CKAA retains task-specific adapters for inference. Although the adapter width can be reduced to $w=2$ for efficiency, the storage and inference costs still grow with the number of tasks. Future work will investigate more compact adapter compression strategies for long task sequences.

\bibliographystyle{IEEEtran}
\bibliography{main}

\begin{IEEEbiography}[{\includegraphics[width=1in,height=1.23in,clip,keepaspectratio]{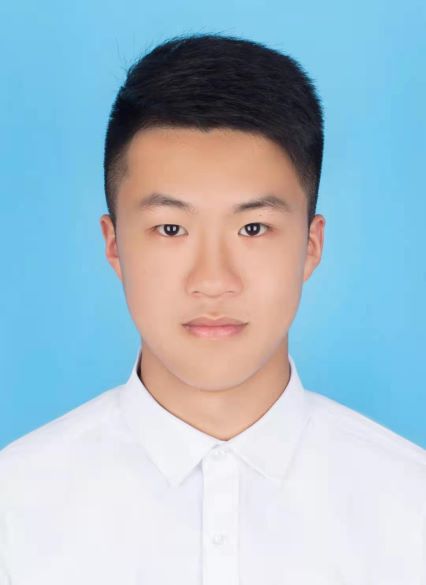}}]{Lingfeng He}
received the B.Sc. degree from Xidian
University, Xi'an, China, in 2023. He is currently
pursuing his M.S. degree in Information and Communication Engineering in Xidian University. His
research interests include person re-identification, unsupervised learning and continual learning.
\end{IEEEbiography}
\vspace{-5.5mm}

\begin{IEEEbiography}[{\includegraphics[width=1in,height=1.25in,clip,keepaspectratio]{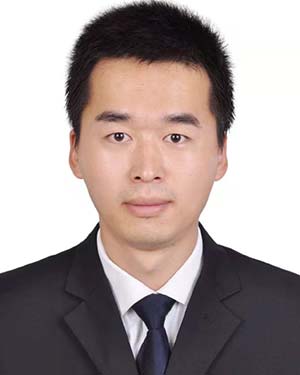}}]{De Cheng}
received the B.S. and Ph.D. degrees from Xi'an Jiaotong University, Xi'an, China, in 2011 and 2017, respectively. He is currently an Associate Professor with the School of Telecommunications Engineering, Xidian University, Xi'an. From 2015 to 2017, he was a Visiting Scholar with Carnegie Mellon University, Pittsburgh, PA, USA. His research interests include pattern recognition, machine learning, and multimedia analysis.
\end{IEEEbiography}

\begin{IEEEbiography}[{\includegraphics[width=1in,height=1.25in,clip,keepaspectratio]{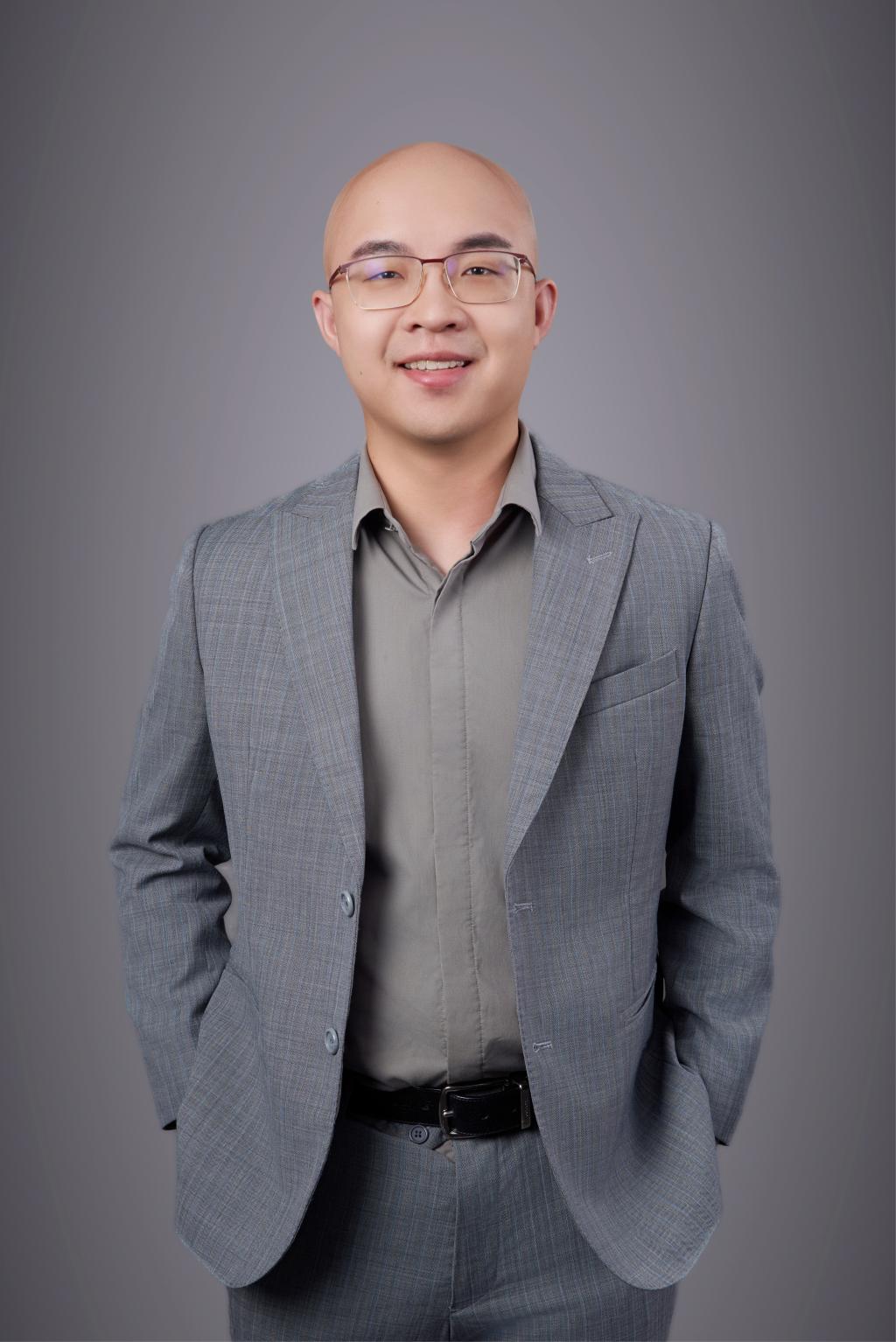}}]{Zhiheng Ma}
received the PhD degree from Xi'an Jiaotong University, in 2021. He is a
research assistant professor with the Shenzhen Institute of Advanced Technology, Chinese Academy of
Sciences (SIAT). He has authored articles in journals
and conferences, such asIEEE Transactions on Image
Processing, CVPR, ICCV, and AAAI. His current research interests include incremental learning, crowd
counting, and novelty detection.
\end{IEEEbiography}
\vspace{-5.5mm}

\begin{IEEEbiography}[{\includegraphics[width=1in,height=1.25in,clip,keepaspectratio]{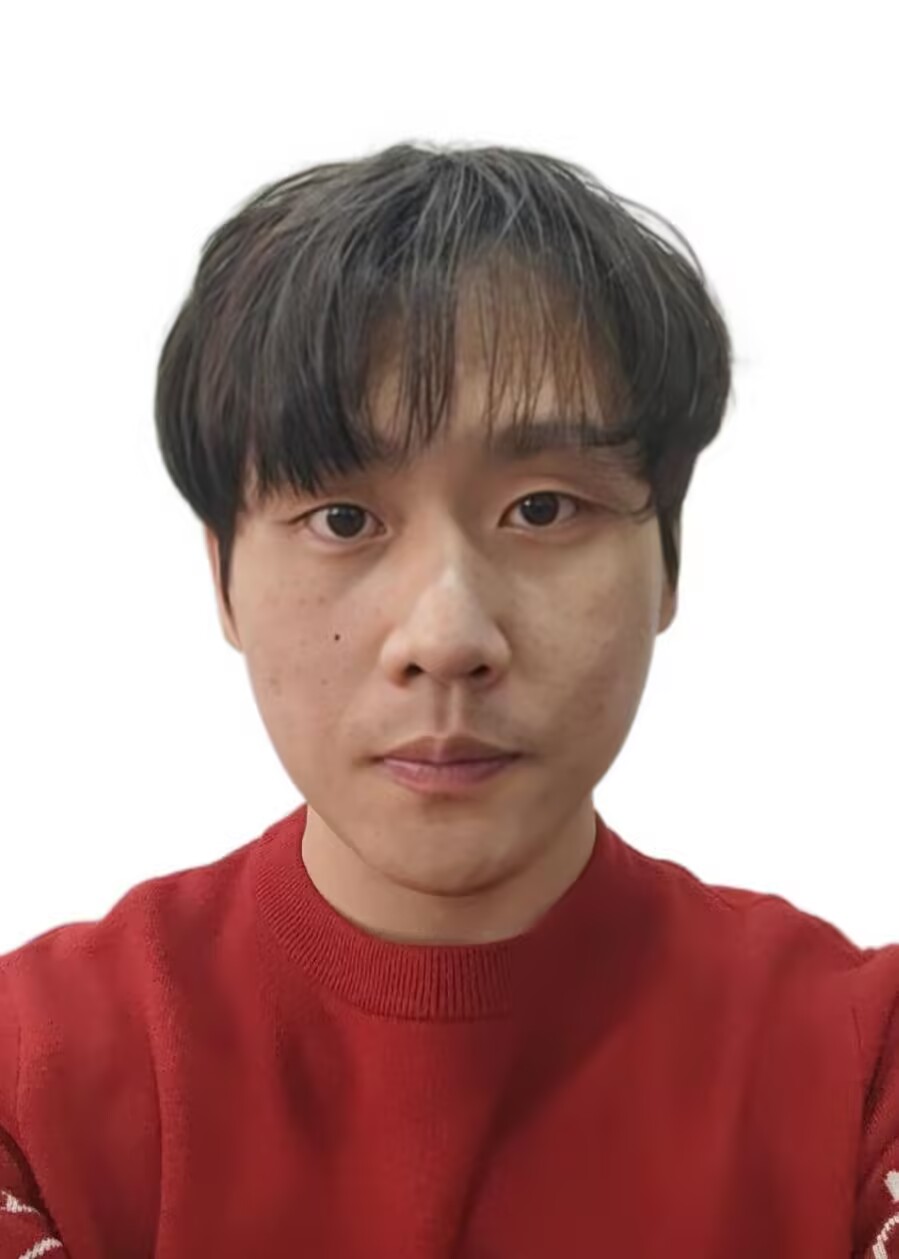}}]{Huaijie Wang}
received the B.Sc. degree from Xidian
University, Xi'an, China, in 2024. He is currently
pursuing his Ph.D. degree in School of Electronic Engineering in Xidian University. His research interest is continual learning.
\end{IEEEbiography}
\vspace{-5.5mm}

\begin{IEEEbiography}[{\includegraphics[width=1in,height=1.25in,clip,keepaspectratio]{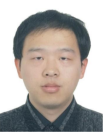}}]{Dingwen Zhang}
is a professor with School of
Automation, Northwestern Polytechnical University,
Xi'an, China. He received his Ph.D. degree from
NPU in 2018. From 2015 to 2017. he was a visiting
scholar at the Robotic Institute, Carnegie Mellon
University, Pittsburgh, United States. His research
interests include computer vision and multimedia
processing, especially on saliency detection and
weakly supervised learning.
\end{IEEEbiography}
\vspace{-5.5mm}

\begin{IEEEbiography}[{\includegraphics[width=1in,height=1.25in,clip,keepaspectratio]{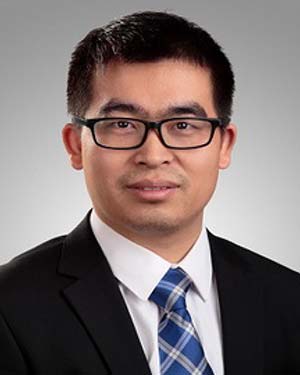}}]{Nannan Wang} received the B.S. degree in information and computation science from the Xi'an University of Posts and Telecommunications in 2009 and the Ph.D. degree in information and telecommunications engineering from Xidian University in 2015. From September 2011 to September 2013, he was a Visiting Ph.D. Student with the University of Technology Sydney, Australia. He is currently a Professor with the State Key Laboratory of Integrated Services Networks, Xidian University. 
\end{IEEEbiography}
\vspace{-5.5mm}

\begin{IEEEbiography}[{\includegraphics[width=1in,height=1.25in,clip,keepaspectratio]{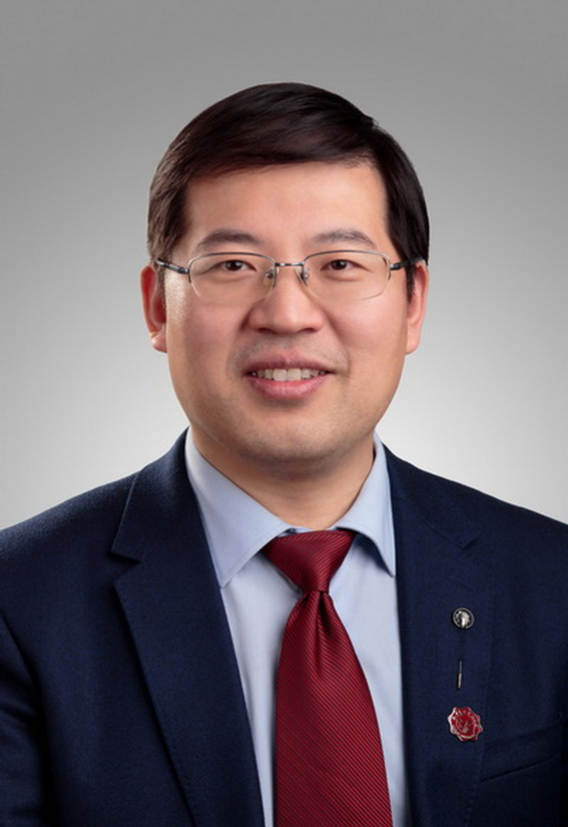}}]{Xinbo Gao}(M'02-SM'07) received the B.Eng., M.Sc. and Ph.D. degrees in electronic engineering, signal and information processing from Xidian University, Xi¡¯an, China, in 1994, 1997, and 1999, respectively. From 1997 to 1998, he was a Research Fellow with the Department of Computer Science, Shizuoka University, Shizuoka, Japan. From 2000 to 2001, he was a Post-Doctoral Research Fellow with the Department of Information Engineering, the Chinese University of Hong Kong, Hong Kong. Since 2001, he has been with the School of Electronic Engineering, Xidian University. He is also a Cheung Kong Professor of the Ministry of Education of China, a Professor of Pattern Recognition and Intelligent System with Xidian University, and a Professor of Computer Science and Technology with the Chongqing University of Posts and Telecommunications, Chongqing, China. He has published 6 books and around 300 technical articles in refereed journals and proceedings. His research interests include image processing, computer vision, multimedia analysis, machine learning, and pattern recognition. Prof. Gao is also a Fellow of the Institute of Engineering and Technology and the Chinese Institute of Electronics. He has served as the general chair/cochair, the program committee chair/co-chair, or a PC member for around 30 major international conferences. He is also on the Editorial Boards of several journals, including Signal Processing (Elsevier) and Neurocomputing (Elsevier).
\end{IEEEbiography}

\clearpage

\appendices
\section{Theoretical Analysis}\label{appx: theory}

In this section, we would like to provide detailed theoretical analysis of the working mechanism of the proposed Dual-level Knowledge Alignment (DKA) and Task-Confidence-guided Mixture-of-Adapters (TC-MoA) modules.

\textbf{Optimization Objective.} Consider a sample $(\mathbf{x},y)\sim\mathcal{D}_r$ from task $r$. Let the feature generated by the correct task-specific module be $\mathbf{z}_r(\mathbf{x})=F(\mathbf{x};\mathcal{P}_{ sh},\mathbf{A}_{sp}^{r})$, and let the feature generated by an incorrectly selected historical module $s<r$ be $\mathbf{z}_s(x)=\mathcal{F}(\mathbf{x};\mathcal{P}_{sh},\mathcal{A}_{sp}^{s})$. Let $g(\cdot)$ denote the classifier and $h_g(\mathbf{z})=\arg\max_k g_k(\mathbf{z})$ denote its prediction function. The risks under correct and incorrect routing are respectively defined as $R_{r\rightarrow r}(g)=\Pr_{(\mathbf{x},y)\sim\mathcal{D}r}[h_g(\mathbf{z}_r(\mathbf{x}))\neq y]$ and $R_{r\rightarrow s}(g)=\Pr_{(\mathbf{x},y)\sim\mathcal{D}_r}[h_g(\mathbf{z}_s(\mathbf{x}))\neq y]$. Our robustness objective is to minimize the classification risk over both the correct subspace and possible incorrect historical subspaces:
\begin{equation}
\begin{aligned}
& \min \mathcal{L}_{r}^{\rm robust}(g)=\\
& R_{r\rightarrow r}(g)+\frac{1}{r-1}\sum_{s=1}^{r-1}\left[\lambda_1 R_{r\rightarrow s}(g) + \lambda_2 R_{s\rightarrow r}(g)\right],
\end{aligned}
\end{equation}
\noindent where $\lambda_1$ and $\lambda_2$ are trade-off hyper-parameters. The first term can be optimized by the standard cross-entropy loss. Therefore, this work mainly focuses on reducing the two cross-subspace misrouting terms. Specifically, the proposed CSFA loss targets the incorrect new-to-old routing risk $R_{r\rightarrow s}(g)$, and the proposed CA targets the incorrect old-to-new routing risk $R_{s\rightarrow r}(g)$. In this way, CSFA and CA complementarily improve model robustness against bidirectional misleading task-ID selection.

\subsection{Theoretical analysis of FA}

In this section, we would like to explain how $\mathcal{L}_{csfa}$ minimizes $R_{r\rightarrow s}(g)$. For $(\mathbf{x},y)$, we define its classification margin in the correct subspace as $m_r(\mathbf{x},y)=g_y(\mathbf{z}_r(\mathbf{x}))-\max_{k\neq y}g_k(\mathbf{z}_r(\mathbf{x}))$. If $m_r(\mathbf{x},y)>0$, the sample is correctly classified when routed to its correct task-specific subspace. We further define its classification margin in the incorrectly  subspace of the $s$-th task as $m_s(\mathbf{x},y)=g_y(\mathbf{z}_s(\mathbf{x}))-\max_{k\neq y}g_k(\mathbf{z}_s(\mathbf{x}))$. We assume that the logit difference between any two classes is $L_g$-Lipschitz with respect to the input feature representation. Specifically, for any classes $y$ and $k$, and any feature representations $\mathbf{z}_r$ and $\mathbf{z}_s$, we have:
\begin{equation}
\begin{aligned}
& \left|
\left(g_y(\mathbf{z}_r(\mathbf{x}))-g_k(\mathbf{z}_r(\mathbf{x}))\right)
-
\left(g_y(\mathbf{z}_s(\mathbf{x}))-g_k(\mathbf{z}_s(\mathbf{x}))\right)
\right| 
\\& \leq
L_g\left\|\mathbf{z}_r(\mathbf{x})-\mathbf{z}_s(\mathbf{x})\right\|_2.
\end{aligned}
\end{equation}
\noindent Therefore, the margin in the incorrect subspace $s$ is lower bounded by:
\begin{equation}
\begin{aligned}
& g_y\bigl(\mathbf{z}_s(\mathbf{x})\bigr)
- g_k\bigl(\mathbf{z}_s(\mathbf{x})\bigr)
\geq\; 
\\&g_y\bigl(\mathbf{z}_r(\mathbf{x})\bigr)
- g_k\bigl(\mathbf{z}_r(\mathbf{x})\bigr) -
L_g \left\| \mathbf{z}_r(\mathbf{x}) -
\mathbf{z}_s(\mathbf{x}) \right\|_2 .
\end{aligned}
\end{equation}
\noindent Since the above inequality holds for every competing class $k\neq y$ for an input $\mathbf{x}$, we take the minimum over all $k\neq y$ on both sides. By the definitions of $m_r(\mathbf{x},y)$ and $m_s(\mathbf{x},y)$, we obtain the following inequality:
\begin{equation}
m_s(\mathbf{x},y)\geq
m_r(\mathbf{x},y)-
L_g\left\|\mathbf{z}_r(\mathbf{x})-
\mathbf{z}_s(\mathbf{x})\right\|_2 .
\end{equation}
\noindent Consider an arbitrary margin threshold $\gamma>0$, we can derive an upper bound on this risk:
\begin{equation}\label{eq: r2_fa_bound}
\begin{aligned} 
R_{r\rightarrow s}(g) & \leq\; \Pr_{(\mathbf{x},y)\sim\mathcal{D}_r} \left[ m_r(\mathbf{x},y)<\gamma \right] 
\\& + \Pr_{(\mathbf{x},y)\sim\mathcal{D}_r} \left[ \left\| \mathbf{z}_r(\mathbf{x}) - \mathbf{z}_s(\mathbf{x}) \right\|_2 \geq \frac{\gamma}{L_g} \right]. 
\end{aligned} 
\end{equation}

\noindent The above upper bound indicates that this risk $R_{r\rightarrow s}(g)$ is associated with two factors: (1) the probability that a sample has an insufficient classification margin in its correct subspace, and (2) the probability that its features induce a large discrepancy when projected into an incorrect historical subspace. 

Then we would like to explain why the proposed CSFA loss minimize this upper bound. Specifically, our CSFA loss can be rewritten as follows:
\begin{equation} 
\begin{aligned} 
\mathcal{L}_{\rm csfa} & = \frac{1}{B} \sum_{i=1}^{B} \log \left[ \sum_{\mathbf{z}_j \in \mathcal{P}(\mathbf{z}_i)\cup \mathcal{N}(\mathbf{z}_i)} \exp \left( \frac{\operatorname{sim}(\mathbf{z}_i,\mathbf{z}_j)} {\tau_f} \right) \right] 
\\& - \frac{1}{B} \sum_{i=1}^{B} \log \left[ \sum_{\mathbf{z}^{+} \in \mathcal{P}(\mathbf{z}_i)} \exp \left( \frac{\operatorname{sim}(\mathbf{z}_i,\mathbf{z}^{+})} {\tau_f} \right) \right].
\end{aligned} 
\end{equation}

The first term in $\mathcal{L}_{\rm csfa}$ contains features with different labels $\mathbf{z}_j \in \mathcal{N}(\mathbf{z}_i)$. Minimizing $\mathcal{L}_{\rm csfa}$ suppresses their similarity to $\mathbf{z}_i$, thereby enlarging the separation between different classes in the current subspace. Such enhanced inter-class discriminability encourages larger classification margins $m_r(\mathbf{x},y)$ and thus helps reduce the first term $\Pr_{(\mathbf{x},y)\sim\mathcal{D}_r} \left[ m_r(\mathbf{x},y)<\gamma \right]$ in Eq.\ref{eq: r2_fa_bound}. 

The second term maximizes the similarity between each feature $\mathbf{z}_i$ and its same-class positive features $\mathbf{z}^{+}\in \mathcal{P}(\mathbf{z}_i)$ extracted from incorrect subspaces. This cross-subspace feature alignment reduces the $L_2$-distance between representations across subspaces, thereby helping reduce the second term in $\Pr_{(\mathbf{x},y)\sim\mathcal{D}_r} \left[ \left\| \mathbf{z}_r(\mathbf{x}) - \mathbf{z}_s(\mathbf{x}) \right\|_2 \geq \frac{\gamma}{L_g} \right]$ in Eq.\ref{eq: r2_fa_bound}.

\subsection{Theoretical analysis of CA}

In this section, we would like to analyze why our CA minimizes the old-to-new misrouting term $R_{s\rightarrow r}(g)$. For a historical sample $(\mathbf{x},y)\sim\mathcal{D}_s$ with $s<r$, it feature from the task-shared network is denoted as $\mathbf{f} = F(\mathbf{x};\mathcal{P}_{sh})$, and its feature generated after being incorrectly routed into the current task-specific subspace $r$ is denoted as $\mathbf{z}_r(\mathbf{x})=F(\mathbf{x};\mathcal{P}_{sh},\mathbf{A}_{sp}^{r})$. Accordingly, the incorrect old-to-new routing risk is defined as \begin{equation} R_{s\rightarrow r}(g^r) = \Pr_{(\mathbf{x},y)\sim\mathcal{D}_s} \left[ h_{g^r}\bigl(\mathbf{z}_r(\mathbf{x})\bigr)\neq y \right]. \end{equation} However, historical training samples are unavailable during the learning of task $r$. Therefore, CA constructs a simulated feature $\tilde{\mathbf{z}}$ to approximate the representation that a historical sample would produce in the current subspace. We define the classification margin of the real incorrectly routed feature as $m_{s\rightarrow r}(\mathbf{z}_r(\mathbf{x}),y) = g_y^r\bigl(\mathbf{z}_r(\mathbf{x})\bigr) - \max_{k\neq y} g_k^r\bigl(\mathbf{z}_r(\mathbf{x})\bigr)$, and define the classification margin of its simulated feature as $\tilde{m}_{s\rightarrow r}(\tilde{\mathbf{z}},y) = g_y^r\bigl(\tilde{\mathbf{z}}\bigr) - \max_{k\neq y} g_k^r\bigl(\tilde{\mathbf{z}}\bigr)$.

Similar to the analysis of FA, we assume that the logit difference between any two classes under $g^r$ is $L_{g^r}$-Lipschitz with respect to the input feature representation. the logit difference between the ground-truth class and the competing class on the real incorrectly routed feature $\mathbf{z}_r(\mathbf{x})$ is lower bounded by the corresponding logit difference on the simulated feature $\tilde{\mathbf{z}}$, minus their feature discrepancy scaled by $L_{g^r}$. Since this relation holds for every competing class $k\neq y$, taking the minimum over all competing classes yields the following inequality:
\begin{equation} \begin{aligned} m_{s\rightarrow r}(\mathbf{z}_r(\mathbf{x}),y) \geq\; \tilde{m}_{s\rightarrow r}(\tilde{\mathbf{z}},y) - L_{g^r} \left\| \mathbf{z}_r(\mathbf{x}) - \tilde{\mathbf{z}} \right\|_2 . \end{aligned} 
\end{equation}
\noindent Given an arbitrary margin threshold $\gamma>0$, we can derive an upper bound on this risk: 
\begin{equation} 
\begin{aligned} 
R_{s\rightarrow r}(g^r) & \leq\; \Pr \left[ \tilde{m}_{s\rightarrow r} \bigl(\tilde{\mathbf{z}},y\bigr) < \gamma \right] 
\\& + \Pr \left[ \left\| \mathbf{z}_r(\mathbf{x}) - \tilde{\mathbf{z}} \right\|_2 \geq \frac{\gamma}{L_{g^r}} \right]. 
\end{aligned} 
\end{equation}

The above upper bound indicates that the incorrect old-to-new routing risk $R_{s\rightarrow r}(g^r)$ is determined by two factors: (1) the classification capability $\Pr \left[ \tilde{m}_{s\rightarrow r} \bigl(\tilde{\mathbf{z}},y\bigr) < \gamma \right] $ on the simulated historical features, and (2) the approximation error $Pr \left[ \left\| \mathbf{z}_r(\mathbf{x}) - \tilde{\mathbf{z}} \right\|_2 \geq \frac{\gamma}{L_{g^r}} \right]$ between the simulated features and the real incorrectly routed features. The first term is explicitly optimized by training the task-adaptive classifier $g^r$ with the simulated features in $\mathcal{L}_{\rm ca}$. Therefore, we mainly focus on analyzing the second term, which characterizes the quality of feature simulation.

Since $\left\|\mathbf{z}_r(\mathbf{x})-\tilde{\mathbf{z}}\right\|_2$ is non-negative, applying Markov's inequality yields:
\begin{equation} 
\begin{aligned} 
\Pr \left[ \left\| \mathbf{z}_r(\mathbf{x}) - \tilde{\mathbf{z}} \right\|_2 \geq \frac{\gamma}{L_{g^r}} \right] \leq \frac{L_{g^r}}{\gamma} \mathbb{E} \left[ \left\| \mathbf{z}_r(\mathbf{x}) - \tilde{\mathbf{z}} \right\|_2 \right]. \end{aligned} 
\end{equation} 
\noindent CA samples a same-class pseudo feature $\tilde{\mathbf{f}}_i$ and constructs its simulated feature as follows:
\begin{equation} 
\tilde{\mathbf{z}}_i = \tilde{\mathbf{f}}_i + \frac{ \sum_{j=1}^{B} G_{ij} \left( \mathbf{z}_j-\mathbf{f}_j \right) }{ \sum_{j=1}^{B}G_{ij} }. 
\end{equation} 
\noindent Accordingly, the feature simulation error can be decomposed as follows:
\begin{equation} \begin{aligned} & \left\| \mathbf{z}_r(\mathbf{x}) - \tilde{\mathbf{z}}_i \right\|_2 \\& = \left\| \mathbf{f} + \left( \mathbf{z}_r(\mathbf{x})-\mathbf{f} \right) - \tilde{\mathbf{f}}_i - \frac{ \sum_{j=1}^{B} G_{ij} \left( \mathbf{z}_j-\mathbf{f}_j \right) }{ \sum_{j=1}^{B}G_{ij} } \right\|_2 
\\ &
\leq \left\| \mathbf{f} - \tilde{\mathbf{f}}_i \right\|_2 + \left\| \left( \mathbf{z}_r(\mathbf{x})-\mathbf{f} \right) - \frac{ \sum_{j=1}^{B} G_{ij} \left( \mathbf{z}_j-\mathbf{f}_j \right) }{ \sum_{j=1}^{B}G_{ij} } \right\|_2 . \end{aligned} \end{equation} 
\noindent The first term measures the approximation error introduced by Gaussian sampling, and the second term measures the error of transferring the subspace shift from current samples to the sampled historical feature through the affinity matrix. 

To further bound the affinity-guided subspace shift transfer error, we first note that the affinity matrix only transfers shifts from  features in the neighborhood. Specifically, according to Eq.\textcolor{red}{5}, $G_{ij}>0$ only when $\mathbf{f}_j\in\mathcal{K}(\tilde{\mathbf{f}}_i,K_g)$. We assume that the subspace shift from the shared space to the current task-specific subspace is smooth within the local neighborhood selected by $G$, where $\left\| \left( \mathbf{z}_r(\mathbf{x})-\mathbf{f} \right) - \left( \mathbf{z}_j-\mathbf{f}_j \right) \right\|_2 \leq L_{\Delta} \left\| \mathbf{f} - \mathbf{f}_j \right\|_2$ and $L_{\Delta}$ characterizes the smoothness of the subspace shift. Since $G_{ij}\geq 0$ and $\sum_{j=1}^{B}G_{ij}>0$, given the vector triangle inequality to the weighted combination and the local smoothness assumption, we have:
\begin{equation} 
\begin{aligned} 
& \left\| \left( \mathbf{z}_r(\mathbf{x})-\mathbf{f} \right) - \frac{ \sum_{j=1}^{B} G_{ij} \left( \mathbf{z}_j-\mathbf{f}_j \right) }{ \sum_{j=1}^{B}G_{ij} } \right\|_2 
\\ & \leq \frac{ \sum_{j=1}^{B} G_{ij} \left\| \left( \mathbf{z}_r(\mathbf{x})-\mathbf{f} \right) - \left( \mathbf{z}_j-\mathbf{f}_j \right) \right\|_2 }{ \sum_{j=1}^{B}G_{ij} } 
\\ & \leq L_{\Delta} \frac{ \sum_{j=1}^{B} G_{ij} \left\| \mathbf{f} - \mathbf{f}_j \right\|_2 }{ \sum_{j=1}^{B}G_{ij} }. 
\end{aligned} 
\end{equation}

\noindent Furthermore, by the inequality $\left\| \mathbf{f} - \mathbf{f}_j \right\|_2 \leq \left\| \mathbf{f} - \tilde{\mathbf{f}}_i \right\|_2 + \left\| \tilde{\mathbf{f}}_i - \mathbf{f}_j \right\|_2$, the feature simulation error is bounded by:
\begin{equation} 
\begin{aligned} 
& \left\| \mathbf{z}_r(\mathbf{x}) - \tilde{\mathbf{z}}_i \right\|_2 \leq 
\\& \left( 1+L_{\Delta} \right) \left\| \mathbf{f} - \tilde{\mathbf{f}}_i \right\|_2 + L_{\Delta} \frac{ \sum_{j=1}^{B} G_{ij} \left\| \tilde{\mathbf{f}}_i - \mathbf{f}_j \right\|_2 }{ \sum_{j=1}^{B}G_{ij} }. 
\end{aligned} 
\end{equation} 
Then we obtain the upper bound of the risk as follows: 
\begin{equation} 
\begin{aligned} 
R_{s\rightarrow r}(g^r) 
& \leq \Pr \left[ \tilde{m}_{s\rightarrow r} \bigl(\tilde{\mathbf{z}},y\bigr) < \gamma \right] \\ & + \frac{L_{g^r}}{\gamma} \Bigg\{ \left( 1+L_{\Delta} \right) \mathbb{E} \left[ \left\| \mathbf{f} - \tilde{\mathbf{f}}_i \right\|_2 \right] 
\\ & + 
L_{\Delta} \mathbb{E} \left[ \frac{ \sum_{j=1}^{B} G_{ij} \left\| \tilde{\mathbf{f}}_i - \mathbf{f}_j \right\|_2 }{ \sum_{j=1}^{B}G_{ij} } \right] \Bigg\}. \end{aligned} 
\end{equation} 

Specifically, the first term is explicitly optimized through $\mathcal{L}_{\rm ca}$. The term $\mathbb{E}\left[ \left\|\mathbf{f}-\tilde{\mathbf{f}}_i\right\|_2 \right]$ corresponds to the approximation error from Gaussian sampling. The term $\mathbb{E} \left[ \frac{ \sum_{j=1}^{B} G_{ij} \left\| \tilde{\mathbf{f}}_i - \mathbf{f}_j \right\|_2 }{ \sum_{j=1}^{B}G_{ij} } \right]$ measures the affinity-weighted neighborhood radius between the sampled historical feature and the current-task features in the shared feature space. The second term becomes small when historical and current tasks exhibit locally related semantics.

\begin{figure}[h]
\centering
\includegraphics[width=0.48\textwidth]{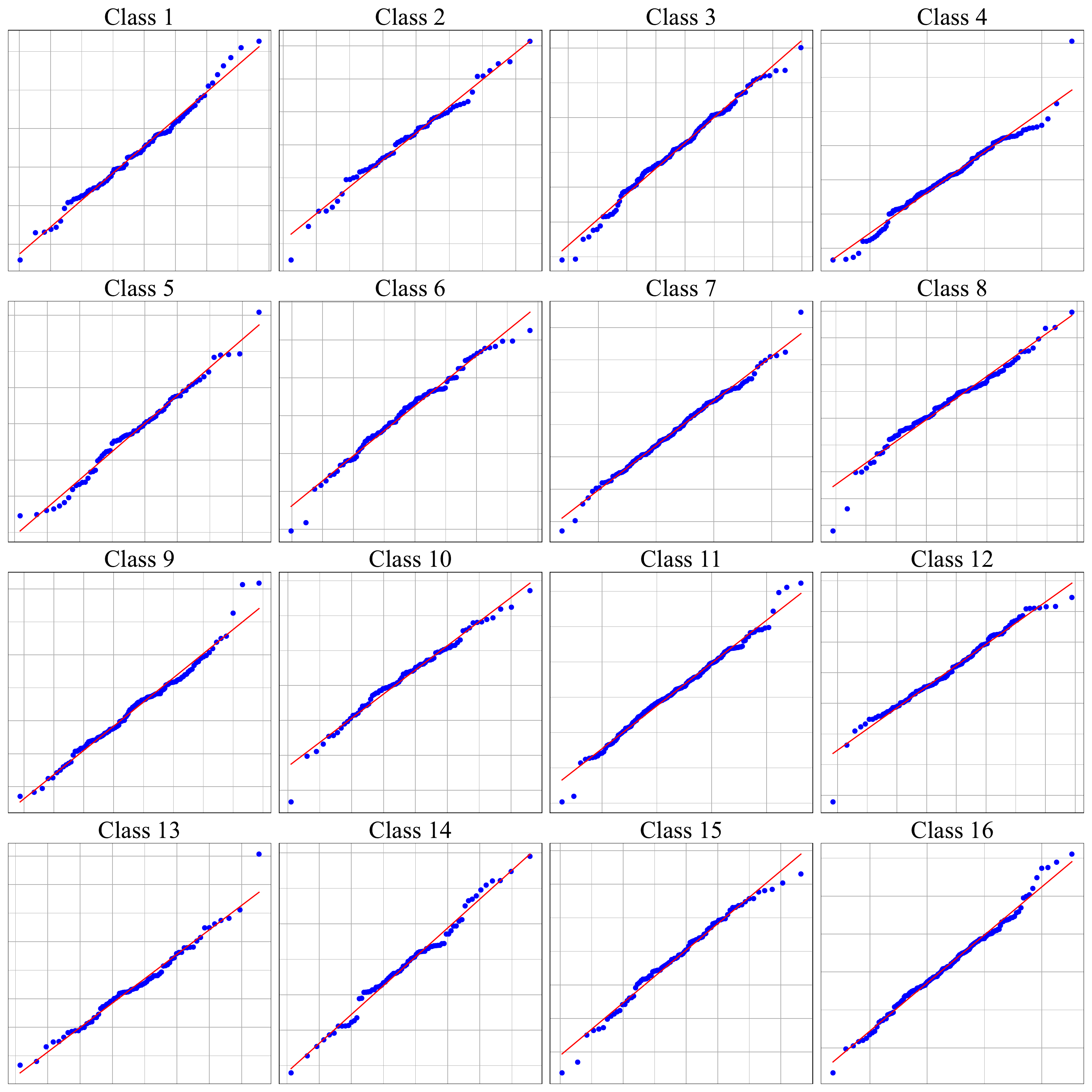}
\caption{Chi-square Q-Q plots of the squared Mahalanobis distances for 16 randomly selected classes on 10S-ImageNetR. }
\label{fig: qq-plot}
\end{figure}

The Gaussian sampling error term $\mathbb{E}\left[ \left\|\mathbf{f}-\tilde{\mathbf{f}}_i\right\|_2 \right]$ is small when the shared features of historical classes can be well approximated by Gaussian distributions. Although this assumption is supported by existing works~\cite{hide-prompt, zhu2021prototype}, we would like to provide further evidence to analyze why it is reasonable. For the $k$-th class in task $t$, let its shared feature distribution be approximated by
$G_{k,t}=\mathcal{N}(\boldsymbol{\mu}_{k,t},\boldsymbol{\Sigma}_{k,t})$.
A fundamental property of the multivariate Gaussian distribution is that, if
$\mathbf{f}\sim\mathcal{N}(\boldsymbol{\mu}_{k,t},\boldsymbol{\Sigma}_{k,t})$, then its squared Mahalanobis distance to the class center follows a chi-square distribution:
\begin{equation}
\begin{aligned}
m(\mathbf{f}, \boldsymbol{\mu}_{k,t}))=
(\mathbf{f}-\boldsymbol{\mu}_{k,t})^{\top}
\boldsymbol{\Sigma}_{k,t}^{-1}
(\mathbf{f}-\boldsymbol{\mu}_{k,t}) \sim
\chi^2_d,
\end{aligned}
\end{equation}
where $d$ is the feature dimension.This property provides a direct diagnostic for examining whether class-wise shared features follow the assumed Gaussian distribution. Specifically, We randomly select 16 historical classes from 10S-ImageNetR and construct chi-square Q-Q plots for their squared Mahalanobis distances. It compares the ordered empirical values of $m(\mathbf{f})$ with the corresponding theoretical quantiles of $\chi^2_r$. As shown in Fig.\ref{fig: qq-plot}, the points for the randomly selected classes are well aligned with the diagonal line, supporting that the Gaussian sampling error term is small.

\begin{figure}[h]
\centering
\includegraphics[width=0.48\textwidth]{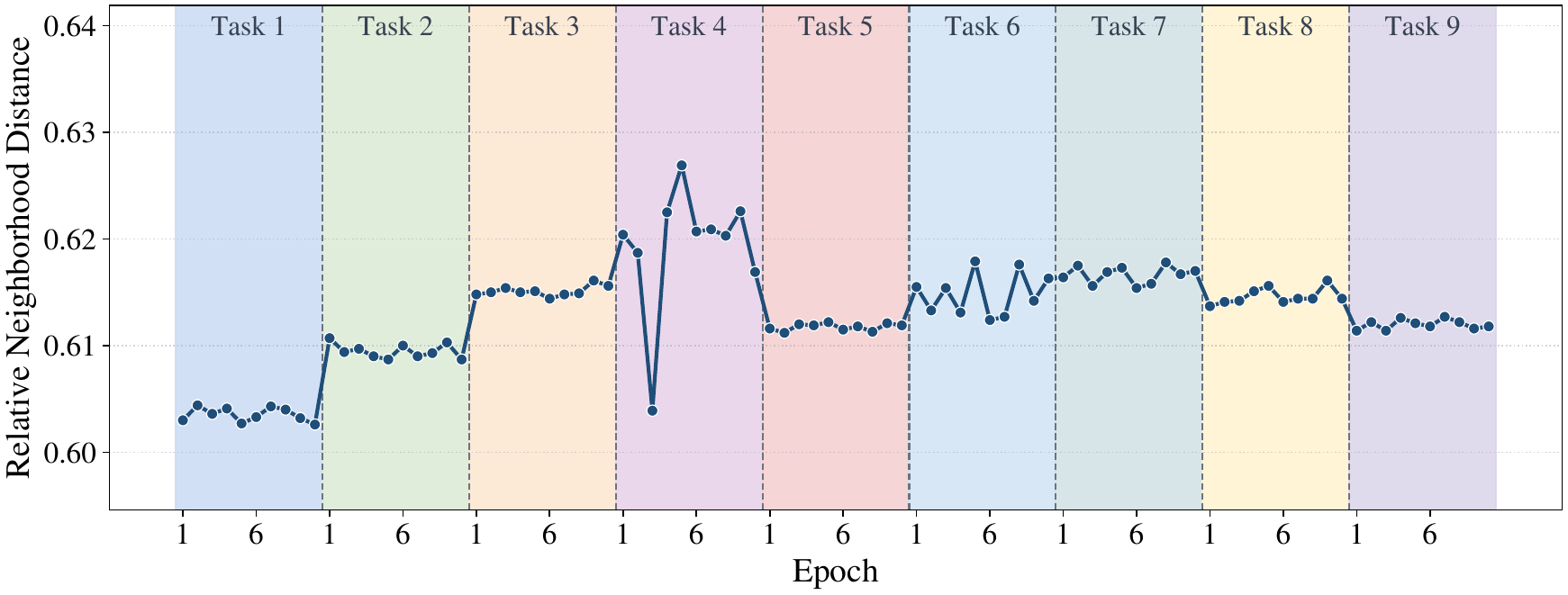}
\caption{Training dynamics of the relative neighborhood distance $\mathcal{R}_{\mathrm{rel}}$.}
\label{fig: relative_neighborhood_dist}
\end{figure}

Then, we further investigate the weighted neighborhood distance term $\mathbb{E} \left[
\frac{
\sum_{j=1}^{B} G_{ij}
\left\| \tilde{\mathbf{f}}_i-\mathbf{f}_j \right\|_2
}{\sum_{j=1}^{B}G_{ij}}
\right]$.
Since the absolute magnitude of this term depends on the scale of features, its value alone cannot directly indicate whether affinity-guided transfer is effective. Therefore, we evaluate its relative magnitude with respect to the average distance obtained without affinity-guided weighting. Specifically, we define the relative neighborhood distance as follows:
\begin{equation}
\mathcal{R}_{\mathrm{rel}}
=
\mathbb{E} \left[
\frac{
\sum_{j=1}^{B} G_{ij}
\left\| \tilde{\mathbf{f}}_i-\mathbf{f}_j \right\|_2
}{
\sum_{j=1}^{B}G_{ij}
}
\right] / 
\mathbb{E} \left[
\frac{1}{B}
\sum_{j=1}^{B}
\left\| \tilde{\mathbf{f}}_i-\mathbf{f}_j \right\|_2
\right].
\end{equation}
\noindent $\mathcal{R}_{\mathrm{rel}}<1$ indicates that, compared with feature simulation without affinity guidance, the proposed affinity-guided transfer reduces this term. We visualize the training dynamics of $\mathcal{R}_{\mathrm{rel}}$ in Fig.\ref{fig: relative_neighborhood_dist} on 10S-ImageNetA. The figure shows that $\mathcal{R}_{\mathrm{rel}}$ remains consistently within $0.60$--$0.63$ throughout continual training across different tasks. This result indicates that affinity-guided transfer reduces the neighborhood distance term to about $60\%$ of that obtained without affinity guidance. 

Therefore, global classifier training explicitly reduces the classification-related term, while Gaussian sampling and affinity-guided subspace shift transfer reduce the simulation-error term, thereby tightening the upper bound of $R_{s\rightarrow r}(g^r)$.

\subsection{Theoretical analysis of TC-MoA} 

\begin{figure}[h]
\centering
\includegraphics[width=0.45\textwidth]{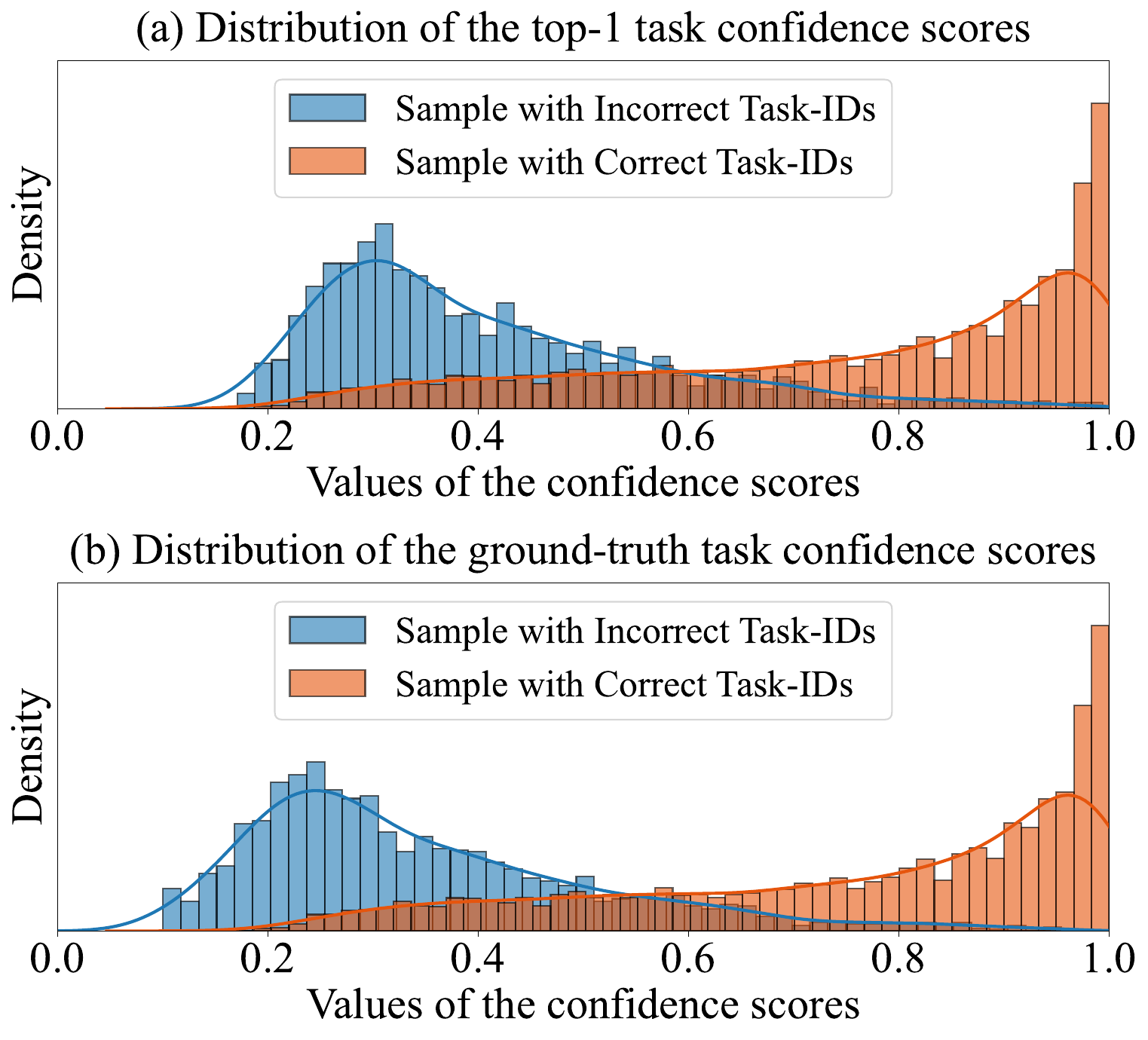}
\caption{Visualization of the distributions of the top-1 task-confidence scores (the upper figure) and the task-confidence scores of the ground-truth task (the lower figure).}
\label{fig: r2-tc-moa-2}
\end{figure}

We further provide an theoretical explanation for why our proposed TC-MoA is
more robust than deterministic maximum selection under confusing task boundaries. For a test sample $\mathbf{x}_i^t$ from task $t$, let $\hat{t}=\arg\max_{t'}\alpha_i^{t'}$ denote the predicted top-1
task-id. Given the test dataset, we define $\mathbb{P}(\hat{t}=t)=p$ and
$\mathbb{P}(\hat{t}\neq t)=1-p$. For easy samples with correct task-IDs, we define the expected confidence score assigned to the ground-truth task as $\mathbb{E}\!\left[\alpha_i^t\mid\hat{t}=t\right]=\tau_1$. For challenging samples with misleading task-ids, although the predicted top-1 task is incorrect, the ground-truth task may still receive a relative high confidence. We define
$\mathbb{E}\!\left[\alpha_i^t\mid\hat{t}\neq t\right]=\tau_2$. This formulation is consistent with the observation in Fig.\ref{fig: r2-tc-moa-2} on 10S-ImageNetA. 
For easy samples with correct task-IDs, the top-1 task-confidence scores are highly concentrated near $1$. This observation directly indicates that
$\tau_1=\mathbb{E}[\alpha_i^t\mid \hat{t}=t]$ is close to $1$. In contrast, for challenging samples with misleading task-IDs, the top-1 task-confidence scores are substantially lower. This indicates that the incorrectly selected adapter is generally not assigned large confidence. 

To analyze how the perturbation introduced by adapter routing propagates to the final prediction, we make a standard Lipschitz smoothness assumption. Specifically, for the $l$-th Transformer block, we assume that the subsequent mapping from its adapter residual output to the final classification scores defined in Eq.\textcolor{red}{13} is $\rho_l$-Lipschitz:
\begin{equation}
\vspace{-3mm}
\begin{aligned}
\mathcal{E}(\mathbf{x}_i^t)
&= \left\|
\frac{t_c}{t_c+1}g^u(\hat{\mathbf{z}}_i)
+
\frac{1}{t_c+1}g(\hat{\mathbf{z}}_i)
\right. \\
&\quad \left.
-
\frac{t_c}{t_c+1}
g^u\!\left(
\mathcal{F}(\mathbf{x}_i^t;\mathcal{P}_{sh},
\mathcal{A}_{sp}^t)
\right)
\right. \\
&\quad \left.
-
\frac{1}{t_c+1}
g\!\left(
\mathcal{F}(\mathbf{x}_i^t;\mathcal{P}_{sh},
\mathcal{A}_{sp}^t)
\right)
\right\|_2 \\
&\leq
\left(1-\alpha_i^t\right)
\max_{\mathbf{x}_i^t,\;t'\neq t}
\sum_{l=1}^{L}
\rho_l
\left\|
A_l^{t'}(\hat{\mathbf{X}}_{i,l})
-
A_l^t(\hat{\mathbf{X}}_{i,l})
\right\|_2 .
\end{aligned}
\end{equation}

\noindent Therefore, the expected error upper bound of TC-MoA is derived as follows:
\begin{equation}
\begin{aligned}
&\mathbb{E}\!\left[
\mathcal{E}_{\rm TC\text{-}MoA}(\mathbf{x}_i^t)
\right] \\
&= p\,\mathbb{E}\!\left[
\mathcal{E}_{\rm TC\text{-}MoA}(\mathbf{x}_i^t)
\mid \hat t=t
\right]
+(1-p)\,\mathbb{E}\!\left[
\mathcal{E}_{\rm TC\text{-}MoA}(\mathbf{x}_i^t)
\mid \hat t\neq t
\right] \\
&\leq
\left[
p\,\mathbb{E}\!\left[
1-\alpha_i^t \mid \hat t=t
\right]
+
(1-p)\,\mathbb{E}\!\left[
1-\alpha_i^t \mid \hat t\neq t
\right]
\right] \\
&\quad \cdot
\max_{\mathbf{x}_i^t,\;t'\neq t}
\sum_{l=1}^{L}
\rho_l
\left\|
A_l^{t'}(\hat{\mathbf{X}}_{i,l})
-
A_l^t(\hat{\mathbf{X}}_{i,l})
\right\|_2 \\
&=
\left[
p(1-\tau_1)
+
(1-p)(1-\tau_2)
\right] \\
&\quad \cdot
\max_{\mathbf{x}_i^t,\;t'\neq t}
\sum_{l=1}^{L}
\rho_l
\left\|
A_l^{t'}(\hat{\mathbf{X}}_{i,l})
-
A_l^t(\hat{\mathbf{X}}_{i,l})
\right\|_2 .
\end{aligned}
\end{equation}

\noindent In contrast, deterministic maximum selection introduces no routing-induced perturbation when $\hat{t}=t$, but fully commits to an incorrect adapter when $\hat{t}\neq t$. The the expected error upper bound is formulated as follows:
\begin{equation}
\begin{aligned}
\mathbb{E}\!\left[
\mathcal{E}_{\mathrm{Max}}
\right] \leq 
(1-p)
\max_{\mathbf{x}_i^t,\;t'\neq t}
\sum_{l=1}^{L}
\rho_l
\left\|
\mathcal{A}_l^{t'}(\hat{\mathbf{X}}_{i,l})
-
\mathcal{A}_l^t(\hat{\mathbf{X}}_{i,l})
\right\|_2.
\end{aligned}
\end{equation}
Therefore, under the following condition, TC-MoA achieves a tighter expected routing-induced error upper bound
than deterministic maximum selection, $i.e.$, $\mathbb{E}\!\left[\mathcal{E}_{\rm TC\text{-}MoA}(\mathbf{x}_i^t)\right]
\leq \mathbb{E}\!\left[\mathcal{E}_{\rm Max}(\mathbf{x}_i^t)\right]$:
\begin{equation}
\boxed{
p<
\frac{\tau_2}{1-\tau_1+\tau_2}.
}
\end{equation}

This condition is especially likely to hold when task boundaries are ambiguous. As shown in Fig.~\ref{fig: r2-tc-moa-2}, on 10S-ImageNetA with ambiguous task boundaries, $1-\tau_1$ is small, indicating that TC-MoA introduces negligible additional perturbation on samples with correctly predicted task-IDs. Moreover, the task-ID accuracy is relatively low ($p<0.5$ on 10S-ImageNetA) but $\tau_2$ remains at a moderate level, indicating that TC-MoA still preserves a relatively large contribution from the ground-truth adapter for these challenging samples. In this case, the condition is easily to be satisfied. 

\begin{table}[t]
\centering
\caption{{Validation of the theoretical condition for TC-MoA.}}
\label{tab: empirical_moa}
\begin{adjustbox}{max width=0.49\textwidth}
\begin{tabular}{l|c|ccccc}
\hline
Benchmark & Filtered & $p$ & $\tau_1$ & $\tau_2$ &
$\frac{\tau_2}{1-\tau_1+\tau_2}$ &  Satisfied \\
\hline
10S-ImageNetA & \ding{55} & 0.541 & 0.843 & 0.247 & 0.611 &
\textcolor{green!50!black}{\ding{51}~(0.611 $>$ 0.541)} \\
10S-ImageNetA & \ding{51} & 0.096 & 0.546 & 0.253 & 0.358 &
\textcolor{green!50!black}{\ding{51}~(0.358 $>$ 0.096)} \\
20S-ImageNetA & \ding{55} & 0.484 & 0.815 & 0.224 & 0.548 &
\textcolor{green!50!black}{\ding{51}~(0.548 $>$ 0.484)} \\
20S-ImageNetA & \ding{51} & 0.042 & 0.476 & 0.208 & 0.284 &
\textcolor{green!50!black}{\ding{51}~(0.284 $>$ 0.042)} \\
10S-ImageNetR & \ding{55} & 0.847 & 0.953 & 0.313 & 0.869 &
\textcolor{green!50!black}{\ding{51}~(0.869 $>$ 0.847)} \\
10S-ImageNetR & \ding{51} & 0.030 & 0.541 & 0.310 & 0.403 &
\textcolor{green!50!black}{\ding{51}~(0.403 $>$ 0.030)} \\
20S-ImageNetR & \ding{55} & 0.830 & 0.916 & 0.292 & 0.777 &
\textcolor{red}{\ding{55}~(0.777 $<$ 0.830)} \\
20S-ImageNetR & \ding{51} & 0.067 & 0.680 & 0.275 & 0.462 &
\textcolor{green!50!black}{\ding{51}~(0.462 $>$ 0.067)} \\
\hline
\end{tabular}
\end{adjustbox}
\vspace{-4mm}
\end{table}

\textbf{Empirical Validation.} We empirically verify whether the theoretical condition $p<\tau_2/(1-\tau_1+\tau_2)$ holds across five representative benchmarks, including 10S-ImageNetA, 20S-ImageNetA, 10S-ImageNetR, 20S-ImageNetR, and 10S-CIFAR100. We report the results in Tab.\ref{tab: empirical_moa}. ``Filtered'' in Tab.\ref{tab: empirical_moa} denotes \emph{\textbf{excluding simple samples that can already be correctly classified using only the task-shared network}}. 

As shown in Tab.\ref{tab: empirical_moa}, the condition is easily to be satisfied on ImageNetA, where its ambiguous task boundaries lead to low task-ID accuracy. In contrast, task-ID prediction on ImageNet-R is more accurate ($p>0.8$), making the condition more stringent and causing it to fail on 20S-ImageNetR. However, our theoretical error quantifies score discrepancy rather than actual misclassification. For easy samples already correctly classified by the task-shared network, this discrepancy rarely changes the predicted class because they rely little on task-specific information and generally have sufficient classification margins. 

Therefore, to remove this confounding effect and provide a cleaner comparison, we exclude these routing-insensitive samples, denoted as ``Filtered". After filtering, the condition is satisfied by a substantial margin across all benchmarks, further demonstrating that TC-MoA provides a clear advantage over deterministic selection for ambiguous samples whose predictions genuinely depend on task-specific information.


\begin{table*}[!t]
\centering
\caption{Experiments with ViT models with different sizes.}\label{tab: r1_varying_models}
\begin{adjustbox}{max width=\textwidth}
\begin{tabular}{c|c|c|cc|cc}
\hline
\multirow{2}*{Models} &
\multirow{2}*{Params} & 
\multirow{2}*{Methods} & 
\multicolumn{2}{c|}{10s-ImageNetA} &
\multicolumn{2}{c}{10s-ImageNetR} \\
\cline{4-7}
& & & Last-acc $\uparrow$ & Avg-acc $\uparrow$ & Last-acc $\uparrow$ & Avg-acc $\uparrow$\\
\hline
\multirow{2}*{ViT-Tiny} 
& \multirow{2}*{5.5M} 
& $\mathcal{P}_{sh}$ (Baseline) 
& 21.02\tsb{\(\pm\)1.27} 
& 31.20\tsb{\(\pm\)1.66}
& 47.82\tsb{\(\pm\)0.68} 
& 56.76\tsb{\(\pm\)0.77}\\
& & CKAA (Ours)
& \textbf{27.48}\tsb{\(\pm\)0.42} 
& \textbf{39.54}\tsb{\(\pm\)0.54}
& \textbf{58.49}\tsb{\(\pm\)0.14} 
& \textbf{67.65}\tsb{\(\pm\)1.08}\\
\hline
\multirow{2}*{ViT-Small} 
& \multirow{2}*{22M} 
& $\mathcal{P}_{sh}$ (Baseline) 
& 37.85\tsb{\(\pm\)0.60} 
& 48.37\tsb{\(\pm\)0.99} 
& 63.61\tsb{\(\pm\)0.34} 
& 71.08\tsb{\(\pm\)0.20}\\
& & CKAA (Ours)
& \textbf{47.07}\tsb{\(\pm\)0.40} 
& \textbf{57.60}\tsb{\(\pm\)1.21} 
& \textbf{73.77}\tsb{\(\pm\)0.15} 
& \textbf{81.05}\tsb{\(\pm\)1.09}\\
\hline
\multirow{2}*{ViT-Base} & \multirow{2}*{86M} & $\mathcal{P}_{sh}$ (Baseline) & 
53.83\tsb{\(\pm\)0.66} & 63.93\tsb{\(\pm\)0.66} & 77.95\tsb{\(\pm\)0.66} & 83.44\tsb{\(\pm\)0.66}\\
& & CKAA (Ours)&
\textbf{63.53}\tsb{\(\pm\)0.66} & 
\textbf{71.11}\tsb{\(\pm\)1.03} & 
\textbf{82.15}\tsb{\(\pm\)0.47} & 
\textbf{87.21}\tsb{\(\pm\)0.13}\\
\hline
\multirow{2}*{ViT-Base (CLIP)} 
& \multirow{2}*{86M} 
& $\mathcal{P}_{sh}$ (Baseline) 
& 51.07\tsb{\(\pm\)0.69} 
& 63.08\tsb{\(\pm\)0.88} 
& 80.55\tsb{\(\pm\)0.62} 
& 86.78\tsb{\(\pm\)0.32}\\
& & CKAA (Ours)
& \textbf{59.25}\tsb{\(\pm\)0.33} 
& \textbf{69.63}\tsb{\(\pm\)1.36} 
& \textbf{83.94}\tsb{\(\pm\)0.46} 
& \textbf{89.42}\tsb{\(\pm\)0.40}\\
\hline
\end{tabular}
\end{adjustbox}
\end{table*}

\section{Additional Experimental Results}

\subsection{Comparison with Traditional CL Methods}

\begin{table}[h]
\centering
\caption{Comparison with representative traditional CL methods on 10S-CIFAR100 and 10S-ImageNetR. For fair comparisons, we report baseline results under ImageNet-21K supervised pre-trained ViT-B/16 backbone.}
\label{tab: traditional_cl_comparison}
\begin{adjustbox}{max width=\columnwidth}
\begin{tabular}{l|c|cc|cc}
\hline
\multirow{2}{*}{Method} &
\multirow{2}{*}{Buffer} &
\multicolumn{2}{c|}{10S-CIFAR100} &
\multicolumn{2}{c}{10S-ImageNetR} \\
\cline{3-6}
& &
Last-acc $\uparrow$ & Avg-acc $\uparrow$ &
Last-acc $\uparrow$ & Avg-acc $\uparrow$ \\
\hline
GDumb\cite{prabhu2020gdumb} & 1,000 &
81.92\tsb{\(\pm\)0.15} &
89.46\tsb{\(\pm\)0.94} &
24.23\tsb{\(\pm\)0.35} &
43.48\tsb{\(\pm\)0.49} \\
DER++\cite{buzzega2020DER} & 1,000 &
84.50\tsb{\(\pm\)1.67} &
91.49\tsb{\(\pm\)0.61} &
67.75\tsb{\(\pm\)0.93} &
78.13\tsb{\(\pm\)1.14} \\
iCARL\cite{rebuffi2017icarl} & 1,000 & 
85.68\tsb{\(\pm\)1.33} &
92.45\tsb{\(\pm\)0.98} &
63.75\tsb{\(\pm\)1.58} &
71.93\tsb{\(\pm\)0.45} \\
BiC\cite{wu2019BiC} & 1,000 &
88.45\tsb{\(\pm\)0.57} &
93.37\tsb{\(\pm\)0.32} &
64.89\tsb{\(\pm\)0.80} &
73.66\tsb{\(\pm\)1.61} \\
EWC\cite{EwC} & 0 &
89.30\tsb{\(\pm\)0.23} &
92.31\tsb{\(\pm\)1.66} &
70.27\tsb{\(\pm\)1.99} &
76.27\tsb{\(\pm\)2.13} \\
LwF\cite{regularization4} & 0 &
87.99\tsb{\(\pm\)0.05} &
92.13\tsb{\(\pm\)1.16} &
67.29\tsb{\(\pm\)1.67} &
74.47\tsb{\(\pm\)1.48} \\
\hline
CKAA & 0 &
\textbf{92.26}\tsb{\(\pm\)0.39} &
\textbf{95.18}\tsb{\(\pm\)0.86} &
\textbf{82.41}\tsb{\(\pm\)0.13} &
\textbf{87.20}\tsb{\(\pm\)0.35} \\
\hline
\end{tabular}
\end{adjustbox}
\end{table}

We compare CKAA with representative traditional CL methods under the ImageNet-21K supervised pre-trained ViT-B/16 backbone, including rehearsal-free regularization-based methods, $i.e.$, EWC~\cite{EwC} and LwF~\cite{regularization4}, and replay-based methods, $i.e.$, GDumb~\cite{prabhu2020gdumb}, DER++~\cite{buzzega2020DER}, iCaRL~\cite{rebuffi2017icarl}, and BiC~\cite{wu2019BiC}. 
For replay-based methods, we use a memory buffer of 1,000 images, while the other methods do not store historical images. As shown in Tab.~\ref{tab: traditional_cl_comparison}, CKAA consistently outperforms these non-PEFT CL baselines without exemplar replay. 
Compared with the strongest traditional baseline EWC in terms of Last-acc, CKAA improves the performance by $2.96\%$ on 10S-CIFAR100 and $12.14\%$ on 10S-ImageNetR. 
These results show that CKAA benefits from preserving the generalization ability of the pre-trained backbone via lightweight task-specific adapters.

\subsection{Generality under Various Models and Settings}

\begin{table}[H]
\centering
\caption{Results of existing methods under various pre-trained models. Here, DINO-1k and iBOT-1k denote that the frozen backbone is pre-trained on ImageNet-1k through DINO and iBOT pre-training algorithms, respectively.}\label{tab: dino-ibot}
\begin{adjustbox}{max width=0.50\textwidth}
\begin{tabular}{l|l|cc}
\hline
PTM & Method & Last-acc $\uparrow$ &Avg-acc $\uparrow$\\
\hline
\multirow{9}*{DINO-1k}  &L2P \cite{l2pCVPR22} & 56.71\tsb{\(\pm\)0.12} & 63.59\tsb{\(\pm\)0.21}\\
{} & DualPrompt \cite{dualpromptECCV22} & 60.23\tsb{\(\pm\)0.42} & 66.57\tsb{\(\pm\)0.25} \\
{} & CODAPrompt \cite{codaCVPR23} & 64.02\tsb{\(\pm\)0.68} &71.50\tsb{\(\pm\)0.42} \\
{} & C-LoRA \cite{clora} & 63.07\tsb{\(\pm\)0.36} & 68.09\tsb{\(\pm\)0.41} \\
{} & LAE \cite{laeICCV23} & 61.03\tsb{\(\pm\)0.27} & 69.89\tsb{\(\pm\)0.15} \\
{} & HiDe-Prompt \cite{hide-prompt} & 68.11\tsb{\(\pm\)0.18} & 71.70\tsb{\(\pm\)0.01} \\
{} & InfLoRA \cite{infloraCVPR24} & 68.31\tsb{\(\pm\)0.28} & 76.15\tsb{\(\pm\)0.05} \\
{} & VPS-NSP$^{\dag}$ \cite{vptnspNIPS-24} & \underline{68.96}\tsb{\(\pm\)0.94} & \underline{76.22}\tsb{\(\pm\)0.56}\\
\rowcolor{lightblue}
{} & \textbf{CKAA (Ours)} &\textbf{75.19}\tsb{\(\pm\)0.29} & \textbf{81.17}\tsb{\(\pm\)0.21}\\
\hline
\multirow{9}*{iBOT-1k}  &L2P \cite{l2pCVPR22} & 60.80\tsb{\(\pm\)0.35} & 66.58\tsb{\(\pm\)0.28}\\
{} & DualPrompt \cite{dualpromptECCV22} & 63.78\tsb{\(\pm\)0.38} & 68.88\tsb{\(\pm\)0.16} \\
{} & CODAPrompt \cite{codaCVPR23} & 68.02\tsb{\(\pm\)0.48} & 74.28\tsb{\(\pm\)0.47}\\
{} & C-LoRA \cite{clora} & 68.60\tsb{\(\pm\)0.07} & 73.47\tsb{\(\pm\)0.28}\\
{} & LAE \cite{laeICCV23} & 64.14\tsb{\(\pm\)0.29} & 72.59\tsb{\(\pm\)0.22}\\
{} & HiDe-Prompt \cite{hide-prompt} & 71.33\tsb{\(\pm\)0.21} & 73.62\tsb{\(\pm\)0.13} \\
{} & InfLoRA \cite{infloraCVPR24} & 71.84\tsb{\(\pm\)0.09} & 78.29\tsb{\(\pm\)0.09}\\
{} & VPS-NSP$^{\dag}$ \cite{vptnspNIPS-24} & \underline{73.25}\tsb{\(\pm\)0.78} & \underline{79.65}\tsb{\(\pm\)0.63} \\
\rowcolor{lightblue}
{} & \textbf{CKAA (Ours)} & \textbf{77.96}\tsb{\(\pm\)0.22} & \textbf{83.26}\tsb{\(\pm\)0.15} \\
\hline
\end{tabular}
\end{adjustbox}
\end{table}

\begin{table}[h]
\centering
\caption{Results under the class-imbalanced setting.}\label{tab: r1_class_imbalance}
\begin{adjustbox}{max width=0.48\textwidth}
\begin{tabular}{l|cc|cc}
\hline
\multirow{2}{*}{Methods} & \multicolumn{2}{c|}{10S-ImageNetA-TI} & \multicolumn{2}{c}{10S-ImageNetR-TI}\\
\cline{2-5}
& Last-acc $\uparrow$ &Avg-acc $\uparrow$
& Last-acc $\uparrow$ &Avg-acc $\uparrow$ \\
\hline
Only $\mathcal{P}_{sh}$ (Baseline) & 46.57\tsb{\(\pm\)0.74} & 56.99\tsb{\(\pm\)0.51} & 
70.54\tsb{\(\pm\)0.12} & 
76.03\tsb{\(\pm\)0.74} \\ 
$\mathcal{P}_{sh}$ + TC-MoA & 48.45\tsb{\(\pm\)0.99} & 58.34\tsb{\(\pm\)0.05} &
72.17\tsb{\(\pm\)0.29} &
77.50\tsb{\(\pm\)0.95} \\
$\mathcal{P}_{sh}$ + TC-MoA+FA & 52.60\tsb{\(\pm\)0.99} & 60.44\tsb{\(\pm\)0.23} &
73.22\tsb{\(\pm\)0.52} &
78.03\tsb{\(\pm\)0.73} \\
CKAA & 
\textbf{58.18}\tsb{\(\pm\)0.38} & \textbf{64.93}\tsb{\(\pm\)0.15} 
& \textbf{75.24}\tsb{\(\pm\)0.21} 
& \textbf{79.60}\tsb{\(\pm\)0.80} \\
\hline
\end{tabular}
\end{adjustbox}
\end{table}

\begin{table}[h]
\centering
\caption{Results under the task-imbalance setting.}\label{tab: r1_class_imbalance}
\begin{adjustbox}{max width=0.48\textwidth}
\begin{tabular}{l|cc|cc}
\hline
\multirow{2}{*}{Methods} & \multicolumn{2}{c|}{10S-ImageNetA-TI} & \multicolumn{2}{c}{10S-ImageNetR-TI}\\
\cline{2-5}
& Last-acc $\uparrow$ &Avg-acc $\uparrow$
& Last-acc $\uparrow$ &Avg-acc $\uparrow$ \\
\hline
Only $\mathcal{P}_{sh}$ (Baseline)& 52.86\tsb{\(\pm\)0.92} & 
62.91\tsb{\(\pm\)1.07} &
77.16\tsb{\(\pm\)0.24} & 
81.99\tsb{\(\pm\)0.44} \\ 
$\mathcal{P}_{sh}$ + TC-MoA &
55.69\tsb{\(\pm\)0.69} & 
64.39\tsb{\(\pm\)0.45} &
79.37\tsb{\(\pm\)0.21} & 
83.97\tsb{\(\pm\)0.24} \\
$\mathcal{P}_{sh}$ + TC-MoA+FA &
57.71\tsb{\(\pm\)0.83} & 
66.29\tsb{\(\pm\)0.58} &
80.39\tsb{\(\pm\)0.37} & 
84.39\tsb{\(\pm\)0.46} \\
CKAA &
\textbf{62.58}\tsb{\(\pm\)0.42} &
\textbf{70.77}\tsb{\(\pm\)0.79}
& 
\textbf{82.15}\tsb{\(\pm\)0.17}
&
\textbf{86.34}\tsb{\(\pm\)0.31}
\\
\hline
\end{tabular}
\end{adjustbox}
\end{table}

\begin{table}[h]
\centering
\caption{Comparison with existing Class-Incremental Semantic Segmentation (CISS) methods on Pascal VOC under the disjoint setting of 15-5 and 15-1 scenarios.}
\label{tab:voc_disjoint_15_5_15_1}
\resizebox{0.48\textwidth}{!}{
\begin{tabular}{l|ccc|ccc}
\hline
\multirow{2}{*}{Method} 
& \multicolumn{3}{c|}{15-5 (2 steps)} 
& \multicolumn{3}{c}{15-1  (6 steps)} \\
\cline{2-7}
& 1-15 & 16-20 & All 
& 1-15 & 16-20 & All \\
\hline
\multicolumn{7}{c}{\textbf{CNN-based Methods}} \\
\hline
EWC~\cite{EwC}    & 26.7 & 37.7 & 29.4 & 0.3  & 4.3  & 1.3  \\
ILT~\cite{css-ILT}   & 63.2 & 39.5 & 57.3 & 3.7  & 5.7  & 4.2  \\
MiB~\cite{css-MiB}    & 71.8 & 43.3 & 64.7 & 46.2 & 12.9 & 37.9 \\
RECALL~\cite{css-RECALL}          & 66.3 & 49.8 & 63.5 & 66.0 & 44.9 & 62.1 \\
SPPA~\cite{css-SPPA}            & 75.3 & 48.7 & 69.0 & 59.6 & 15.6 & 49.1 \\
RBC~\cite{css-RBC}    & 75.1 & 49.7 & 69.9 & 61.7 & 19.5 & 51.6 \\
\rowcolor{lightgray}
Upper Bound & 79.1 & 72.6 & 77.4 & 79.1 & 72.6 & 77.4 \\
\hline
\multicolumn{7}{c}{\textbf{Transformer-based Methods}} \\
\hline
MiB~\cite{css-MiB}    & 75.0 & 59.9 & 72.3 & 66.7 & 26.3 & 58.3 \\
RBC~\cite{css-RBC}    & 77.7 & 59.1 & 74.0 & 69.0 & 28.4 & 60.5 \\
INC~\cite{css-INC}    & 81.6 & 62.2 & 77.6 & 81.4 & 57.1 & 76.2 \\
MBS~\cite{park2024mitigating} & 81.9 & 67.2 & \underline{79.0} & \textbf{81.5} & \underline{64.7} & \underline{78.1} \\
\rowcolor{lightblue}
Task-shared Net & 79.9 & 67.5 & 77.4 & 76.5 & 61.7 & 72.8 \\
\rowcolor{lightblue}
TC-MoA (Ours) & 81.5 & \underline{68.5} & 78.9 & 78.2 & 60.9 & 73.9 \\
\rowcolor{lightblue}
CKAA (Ours) & \underline{81.7} & \textbf{71.9} & \textbf{79.9} & \underline{81.3} & \textbf{68.5} & \textbf{78.7} \\
\rowcolor{lightgray}
Upper Bound   & 84.8 & 80.7 & 83.8 & 84.8 & 80.7 & 83.8 \\
\hline
\end{tabular}
}
\vspace{-4mm}
\end{table}

\textbf{Performances under various pre-trained models.} We conduct experiments on 10s-ImageNetR with ViT-B/16 under various self-supervised pre-training methods, including DINO \cite{DINO} and iBOT \cite{iBOT}, following \cite{infloraCVPR24, hide-prompt}. The results are shown in Tab.\ref{tab: dino-ibot}, which exhibit a performance degradation with a self-supervised pre-trained backbone compared to utilizing a backbone pre-trained in a supervised manner. However, our CKAA still outperforms existing PEFT-based methods. The results demonstrate that CKAA can be adapted to models with various pre-training schemes, highlighting its generalizability.

\textbf{Performances under various model sizes.} We evaluate CKAA with ViT backbones of different scales, including ViT-Tiny (5.5M), ViT-Small (22M), ViT-Base (86M), and the CLIP-pretrained ViT-Base backbone~\cite{2021clip}. For each backbone, we compare CKAA with our $\mathcal{P}_{sh}$ baseline. As shown in Tab.\ref{tab: r1_varying_models}, CKAA consistently outperforms the baseline, demonstrating the effectiveness of CKAA across different model capacities.

\textbf{Performances under imbalanced settings.} We verify the effectiveness of CKAA under class-imbalanced setting by building two long-tailed variants of 10S-ImageNetA and 10S-ImageNetR, denoted as 10S-ImageNetA-LT and 10S-ImageNetR-LT. Specifically, we resample the training set of each class according to a long-tailed distribution. For the $i$-th class among $C$ classes, the number of retained training samples is defined as $n_i = \max \left( 1, \min \left( 
\left\lfloor N_{\max} \cdot \rho^{-\frac{i}{C-1}} \right\rfloor, 
N_i^{\mathrm{ori}} 
\right) \right).$ In our experiments, we set $N_{\max}=200$ and $\rho=20$, resulting in a head-to-tail imbalance ratio of $20:1$. The results are shown in Tab.\ref{tab: r1_class_imbalance}. We further examine the effectiveness of our method under more realistic non-uniform task distributions by building two task-imbalanced variants of 10S-ImageNetA and 10S-ImageNetR, denoted as 10S-ImageNetA-TI and 10S-ImageNetR-TI. we randomly split all classes in the original datasets into 10 non-overlapping tasks with varying class numbers. The results are shown in Tab.\ref{tab: r1_class_imbalance}. These results indicate that CKAA remains effective under the challenging imbalanced scenarios.

\textbf{Performances on the Class-Incremental Semantic Segmentation (CISS) task.} To further verify the generality of CKAA, we conduct experiments on Pascal VOC~\cite{everingham2015pascal} under the disjoint 15-5 and 15-1 settings~\cite{park2024mitigating, css-MiB}, where the model first learns classes 1--15 and then incrementally learns classes 16--20. Following prior works~\cite{park2024mitigating, css-INC, css-MiB}, we report mIoU over old classes, new classes, and all 21 semantic classes including background. Following Transformer-based CISS work~\cite{park2024mitigating}, we adopt an ImageNet-pretrained ViT~\cite{ViT} as the encoder and a lightweight Transformer-based segmentation decoder. We build the task-shared network by fully fine-tuning the backbone with cross-entropy and distillation losses, while training task-specific adapters via the cross-entropy loss. For inference, to make the task confidence score more semantically discriminative, we compute it only over the predicted foreground regions. As shown in Tab.~\ref{tab:voc_disjoint_15_5_15_1}, CKAA achieves the best overall performance among Transformer-based CISS methods. The ablation variants ($i.e.$, Task-shared Net and TC-MoA) also verify the effectiveness of task-confidence-based aggregation and DKA. The improvement is relatively moderate compared to the CIL setting because CISS has strong old-new task ambiguity: new images may contain old objects and old images may contain future objects, which (1) makes task-specific structures less suitable and (2) weakens the benefit of the Gaussian sampling and feature simulation in CA.




\end{document}